\documentclass[10pt,twocolumn,letterpaper]{article}

\usepackage[pagenumbers]{cvpr} 

\definecolor{cvprblue}{rgb}{0.21,0.49,0.74}
\usepackage{natbib}
\usepackage{multibib}
\usepackage[pagebackref,breaklinks,colorlinks,allcolors=cvprblue]{hyperref}
\usepackage{url}
\usepackage{graphicx}
\usepackage{multirow}
\usepackage{xcolor}
\usepackage[verbose]{placeins}
\usepackage[export]{adjustbox}
\usepackage{booktabs}
\usepackage{amsmath,amsfonts,bm}
\usepackage{makecell}
\usepackage[table]{xcolor}
\usepackage{comment}
\usepackage{subcaption} 
\usepackage{capt-of}    
\usepackage[accsupp]{axessibility}  

\definecolor{MoreReadableGreen}{HTML}{00b900}

\title{AlignPose: Generalizable 6D Pose Estimation via \\ Multi-view Feature-metric Alignment}

\author{
\centering
\begin{tabular}{c}
Anna Šárová Mikeštíková$^{1}$ \quad
Médéric Fourmy$^{1}$ \quad
Martin Cífka$^{1,2}$ \quad
Josef Sivic$^{1}$ \quad
Vladimir Petrik$^{1}$\vspace{0.6em} \\
\small{$^{1}$Czech Institute of Informatics, Robotics and Cybernetics, Czech Technical University in Prague}\\
\small{$^{2}$Faculty of Electrical Engineering, Czech Technical University in Prague}\\ 
\small{\url{https://mikestikova.github.io/alignpose/}}
\end{tabular}
}

\begin{document}
\maketitle
\begin{abstract}
Single-view RGB model-based object pose estimation methods achieve strong generalization but are fundamentally limited by depth ambiguity, clutter, and occlusions. Multi-view pose estimation methods have the potential to solve these issues, but existing works rely on precise single-view pose estimates or lack generalization to unseen objects. We address these challenges via the following three contributions. First, we introduce AlignPose, a 6D object pose estimation method that aggregates information from multiple extrinsically calibrated RGB views and does not require any object-specific training or symmetry annotation. Second, the key component of this approach is a new multi-view feature-metric refinement specifically designed for object pose. It optimizes a single, consistent world-frame object pose by minimizing the feature discrepancy between on-the-fly rendered object features and observed image features across all views simultaneously. Third, we report extensive experiments on six datasets (YCB-V, T-LESS, HouseCat6D, ITODD-MV, IPD, XYZ-IBD) using the BOP benchmark evaluation and show that AlignPose outperforms other published methods, especially on challenging industrial datasets where multiple views are readily available in practice.
\end{abstract}    
\section{Introduction}

Model-based 6D object pose estimation is a key component of several real-world applications including augmented reality, robotic manipulation, and automated industrial inspection. Recent RGB-only learning-based methods have demonstrated strong performance on this problem including generalization to new objects unseen during training~\citep{labbe2022megapose,ornek2024foundpose}. However, these approaches are single-view and hence inherently struggle in cluttered scenes due to occlusions, appearance ambiguities (e.g., a cup with a hidden handle~\citep{hodavn2016evaluation}), and the depth ambiguity. On the other hand, depth-based estimators typically achieve superior precision but may fail for certain types of objects (e.g., reflective, transparent~\citep{chen2022clearpose}), and industrial depth cameras are often expensive. Leveraging multiple RGB views thus offers a compelling alternative to address these issues.

However, existing multi-view methods either rely on too restrictive assumptions that discard valid candidate poses~\citep{labbe2020cosypose} or require object-specific training~\citep{shugurov2021dpodv2}, which may require hours or days of training time. Drawing inspiration from feature-metric refinement, which has proven highly effective for camera localization and scene-level bundle adjustment~\citep{lindenberger2021pixelperfect, sarlin2021back}, we adapt and reformulate this principle for the distinct problem of object-centric 6D pose estimation. We formulate the object pose estimation problem as a {\em joint multi-view feature-metric} image alignment using frozen features from a foundation model~\citep{oquab2023dinov2}, which enables us to achieve generalizability to new objects without training. Our contributions are three-fold:
\begin{itemize}
    \item We design an approach for multi-view 6D pose estimation from RGB images that does not require any object-specific training or symmetry annotation. We use per-view pose candidates, obtained by any existing single-view pose estimation method, aggregate them in a common 3D coordinate frame, and refine them (see next) to obtain a new, improved object pose estimate.

    \item We propose a new multi-view feature-metric refinement formulation specifically for object pose. It jointly optimizes a single, consistent world-frame pose ($\bm T_{WO}$) by minimizing the feature discrepancy between on-the-fly rendered object features and observed image features across all views simultaneously.

    \item We evaluate our method following the BOP challenge methodology~\citep{bop23} using six datasets: YCB-V, T-LESS, HouseCat6D, ITODD-MV, IPD, and XYZ-IBD. We demonstrate substantial improvements in performance measured by average recall and average precision, including on challenging texture-less, industrial, reflective, and transparent objects. Our approach consistently outperforms both single-view and published state-of-the-art multi-view RGB-based pose estimates. On industrial datasets, we surpass these baselines by more than 14\%.
\end{itemize}
\begin{figure*}
    \centering
    \includegraphics[width=1\linewidth]{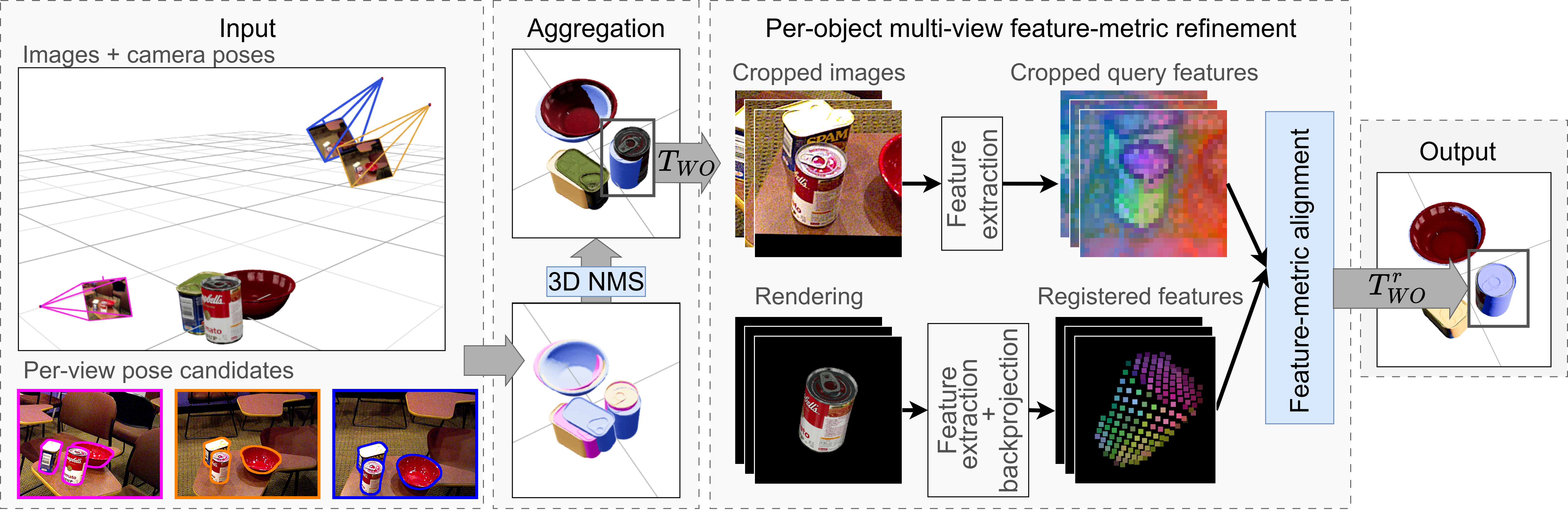}
        \caption{\textbf{Our feature-based multi-view pose estimation pipeline.} Single-view pose candidates are first generated independently for each view using state-of-the-art pose estimation methods (e.g. \citet{labbe2022megapose, ornek2024foundpose}). During aggregation, candidates are transformed into a common coordinate frame, and non-maximum suppression (NMS) is applied to eliminate redundant detections of the same object. These filtered pose candidates $\bm{T}_{WO}$ are then refined using a new multi-view feature-metric refinement to obtain object poses that are consistent from all views.}
    \label{fig:multi_view_pipeline}
    \vspace{-3mm}
\end{figure*}

\section{Related work}
\paragraph{Single-view 6D pose estimation.}
Model-based object detection and 6D pose estimation from \mbox{RGB(-D)} images is a well-established field whose progress is tracked in several benchmarks and datasets \citep{hodan2018bop,van2025bop}. Leading approaches are all deep learning-based, and can be broadly classified into direct regression \citep{xiang2018posecnn}, template-matching \citep{nguyen2024gigapose,ornek2024foundpose,caraffa2024freeze}, 2D-3D correspondence matching \citep{liu2025gdrnpp}, and render-and-compare \citep{li2018deepim,labbe2020cosypose,labbe2022megapose}. A recent focus has been to develop methods able to generalize to objects that are not seen at training time \citep{bop23} by training on large simulated datasets \citep{labbe2022megapose, nguyen2024gigapose, wen2024foundationpose}. Visual \citep{oquab2023dinov2} and point-cloud \citep{poiesi2022learning} foundation models have also been shown to provide useful features for building training-free generalizable pose estimation approaches \citep{ornek2024foundpose, caraffa2024freeze}. We build on this line of work by leveraging a vision foundation model for the task of multi-view object pose estimation.

\paragraph{Multi-view object pose estimation.}
Multiple viewpoints can be used to improve single-view RGB-based pose estimation methods by resolving depth/scale ambiguity and occlusions. Some methods estimate jointly the object pose and temporally linked camera poses, a problem known as object-SLAM \citep{salas2013slam,fu2021multi,zorina2024temporally}, while other assume access to camera poses, e.g. from robot kinematics \citep{wada2020morefusion} or off-line localization \citep{li2018unified}. One approach is to first build a 3D scene representation, using either volumetric representation obtained from RGBD frames \citep{wada2020morefusion,kaskman20206} or implicit representations \citep{taher2024fit} and then register the object model in the scene. Others fuse multiple single-view pose estimates, associated with confidence scores \citep{labbe2022megapose} or even probability distributions \citep{yang20236d}. They are first expressed in a common reference frame using camera poses and can then be aggregated using voting schemes \citep{sock2017multi, li2018unified} or Maximum Likelihood Estimation \citep{erkent2016integration, yang20236d}. CosyPose \citep{labbe2020cosypose} uses pose-level RANSAC \citep{fischler1981random} aggregation to recover consistent camera and object poses, and refines them using bundle adjustment. DPODv2 \citep{shugurov2021multi} proposes a refinement based differentiable rendering by training a model to predict object NOCS representation \citep{wang2019normalized}. We introduce a simple but effective 3D Non Maximum Suppression (NMS) aggregation scheme followed by a multi-view feature-metric refinement which does not require any object specific training and can therefore generalize to any object model.

\paragraph{Feature-metric refinement.}
Direct image alignment \citep{irani1999direct,baker2004lucas} methods localize a camera by minimizing photometric loss and have been used to build highly precise visual SLAM systems \citep{engel2014lsd}. However, their performance drops significantly when the brightness consistency assumption is violated, which often appears in visual localization tasks. To improve robustness, several works propose a \textit{feature-metric} refinement approach in which images are encoded using deep features, which, combined with a non-linear solver, enables camera localization \citep{shu2020feature, von2020lm, sarlin2021back, trivigno2024unreasonable}. In the context of Structure from Motion (SfM), \citep{lindenberger2021pixelperfect} proposes a feature-metric bundle adjustment procedure. Features around previously matched keypoints are compared against a reference appearance to refine all camera and sparse environment estimates. In the context of object pose estimation, \citep{iwase2021repose} uses differentiable rendering and a feature-metric loss to refine object poses, by learning dataset specific deep features.
Inspired by \cite{sarlin2021back},\cite{ornek2024foundpose} minimizes a feature-metric error between pre-rendered templates and query images using an image foundation model \citep{oquab2023dinov2}, which unlocks generalization to new objects. However, these methods target single-view settings, which limits their accuracy. We build on this line of work and introduce an approach to minimize feature-metric loss across multiple camera viewpoints. We do not require matched keypoints from an SfM pipeline and go beyond pre-rendered templates. We jointly optimize a single, consistent world-frame pose by minimizing the feature discrepancy between on-the-fly rendered object features and observed image features simultaneously across all views.

\section{Generalizable multi-view object pose estimation}
\paragraph{Problem formulation.}
The goal of multi-view 6D object pose estimation is to determine the 3D position and orientation of a previously unseen object, given its known 3D mesh and a set of RGB images from a calibrated multi-camera setup.
The object's pose is represented by a rigid transformation $\bm{T}_{WO} \in SE(3)$, which maps the object’s local coordinate frame to a shared world coordinate frame.
The pose observed by any individual camera, $\bm{T}_{CO} \in SE(3)$, is related to this world pose through the known camera extrinsics, $\bm{T}_{CW}$, via the kinematic chain: $\bm{T}_{CO} = \bm{T}_{CW} \bm{T}_{WO}$. Here $\bm{T}_{CW}$ is the transformation from the world frame to the camera's frame, which is assumed to be known from a one-time offline camera calibration procedure.

The primary challenges for multi-view pose estimation are threefold. First, the system must robustly aggregate multiple, often conflicting, pose candidates from different views to resolve per-view ambiguities and establish a single, globally consistent estimate. Second, this estimate must be refined by aligning it with subtle image features for final accuracy. Finally, this aggregation-and-refinement pipeline must achieve zero-shot generalization, operating on novel objects using only their 3D mesh without object-specific training. To address these challenges we propose multi-view pose estimation pipeline, as illustrated in \cref{fig:multi_view_pipeline}. Our approach leverages existing single-view pose estimators to produce per-view {\em candidate poses}, which are then aggregated, filtered, and refined to yield poses that are 3D consistent and well-aligned with the input images. Our method's components are described next.

\subsection{Generation of single-view pose candidates}
As an input to our method, we expect multiple RGB images of a scene, camera intrinsics, along with poses of cameras that captured these images relative to some world frame. For each view we let a single-view pose estimation method predict a set of {\em object pose candidates}. Each candidate is associated with a pose relative to the camera and a confidence score. We do not provide a specific implementation; rather, we assume the use of any off-the-shelf method for single-view pose estimation of unseen objects.

\subsection{Multi-view aggregation}
The aggregation stage aims to consolidate the noisy and redundant pose candidates from all single views into a clean, minimal set of unique 3D object pose candidates. The main challenge is that transforming these single-view estimates into a common world frame creates multiple, overlapping pose candidates for each physical object. To resolve this, we employ 3D Non-Maximum Suppression (NMS) to filter the duplicates. In this process, each candidate is represented by a 3D bounding box derived from its pose and 3D model, using the confidence score from the single-view estimator. Overlapping boxes are then suppressed based on their 3D Intersection over Union (IoU) if they exceed a predefined threshold. This stage concludes with a set of unique pose candidates, each with a coarse but consolidated pose with respect to the world coordinate frame, ready for refinement.

\subsection{Multi-view refinement}
In the refinement stage, each pose candidate is refined independently via {\em multi-view feature-metric alignment}.
The goal is to maximize the alignment between the projection of 3D registered features extracted from the 3D object model and 2D feature maps extracted from input images by optimizing the object pose using a multi-view feature-metric loss.  
In the following, we explain how the 3D registered features and 2D feature maps are obtained, and then describe our loss formulation.

\paragraph{Feature extraction.} Before the multi-view alignment begins, we prepare two fixed representations for each input view: a {\em query 2D feature map} derived from the observed image and {\em 3D registered features} derived from the object's model and coarse pose candidate.
The query is the observed appearance of the scene that acts as the evidence we want to match. For each view, it is derived from the input image by cropping the region that (presumably) contains the object to a standardized size and extracting its feature map with a feature extractor.
The 3D registered features represent the view-dependent appearance of the object based on the coarse pose. This view-dependent appearance is represented by a set of {\em 3D registered features} $\mathcal{F}_{CO} = \{\mathbf{p}_i,\mathbf{x}_i\}$. This is a point cloud where each 3D point $\mathbf{x}_i$ on the object's surface has an associated feature descriptor $\mathbf{p}_i$. We obtain this representation by first rendering color and depth frames for each view using the coarse object pose. We then extract features from the color render and register them in 3D by lifting them to the object's coordinate system using the rendered depth map. Note that while \citep{ornek2024foundpose} pre-computes a large collection of registered features from template views, we compute one set of registered features per view, which may reduce memory usage, depending on the other components of the pose estimation pipeline. Further details on the feature extraction process are provided in Appendix~\ref{sec:appendix_preprocessing}. 

\begin{figure}[t!]
\begin{center}
\includegraphics[width=1.0\linewidth]{figures/figure_templates.png}
\end{center}
\vspace{-5mm}
\caption{\textbf{Multi-view feature-metric refinement.} This figure illustrates a two-view feature-metric refinement. The two cameras show cropped query features (mapping the first three PCA components to RGB values). The partial multi-color point clouds represent the registered features $\mathcal{F}_{CO}$, shifted towards their corresponding camera for visualization purposes. The projection of registered feature 3D coordinates $\mathbf{x}_i$ is represented by a line color coded with a matching color of the feature value $\mathbf{p}_i$.
During refinement, each registered model feature $\mathbf{p}_i$ is compared to the feature value interpolated from query image features $\mathbf{F}_q$ at the location of the corresponding  3D point $\mathbf{x}_i$ projected into the query image at ${\pi}_C \left(  \bm{T}_{CO}  \mathbf{x}_i \right)$, according to \cref{eq:LFE}. For clarity, only a few (3 per view) projected 3D points are shown.
}\label{fig:multiview_refine}
\vspace{-3mm}
\end{figure}

\paragraph{Per-view loss function.}  
For each view, we define the {\em feature-metric loss} of a pose estimate $\bm{T}_{CO}$ as:

\begin{equation}
\mathcal{L}^C_{\text{FE}}(\bm{T}_{CO}) = \sum_{(\mathbf{p}_i, \mathbf{x}_i) \in \mathcal{F}_{CO} }\rho \left( \mathbf{p}_i - \mathbf{F}_q \left( {\pi}_C \left(  \bm{T}_{CO}  \mathbf{x}_i \right)\right) \right) ,
\label{eq:LFE}
\end{equation}
where $\mathcal{F}_{CO}$ is the set of registered features of the view and $\mathbf{F}_q$ are the 2D query features. Each registered object feature $\mathbf{p}_i$ is compared with its corresponding feature from the query feature map $\mathbf{F}_q \left( {\pi}_C \left( \bm{T}_{CO}  \mathbf{x}_i \right)\right)$.
The correspondence is obtained by transforming the 3D object point $\mathbf{x}_i$ from the object frame into the camera frame (by transformation $\bm{T}_{CO}\mathbf{x}_i$), projecting it into the query image (with camera projection ${\pi}_C$), and then sampling the query feature map \( \mathbf{F}_q\) with bilinear interpolation at the given image coordinates. The loss for each registered feature is then calculated using a robust cost function \( \rho(\cdot) \) by \citet{barron2019general}.

\paragraph{Multi-view alignment.}
The goal of multi-view alignment is to find a single consistent object pose in the world frame $\bm{T}^r_{WO}$ that minimizes the feature-metric loss across all camera views. This is achieved by minimizing the sum of per-view feature-metric losses:
\begin{equation}
\bm{T}_{WO}^r = \underset{\bm{T}_{WO}}{\arg\min} \quad
\sum_{C \in \mathcal{C}} {\mathcal{L}^C_{\text{FE}}(\bm{T}_{CW}\bm{T}_{WO})} \, ,
\end{equation}
where $\mathcal{C}$ is the set of all cameras. The optimized object pose $\bm{T}_{CO}$ is expressed in each camera frame following the kinematic chain $\bm{T}_{CO} = \bm{T}_{CW}\bm{T}_{WO}$. 
We visualize the multi-view alignment and the feature-metric loss in \cref{fig:multiview_refine}.
The pose is refined iteratively by the Levenberg-Marquardt~\citep{levenberg1944method, marquardt1963algorithm} optimization algorithm until convergence or a maximum of 30 iterations.

\paragraph{Scoring.} 
\label{sec:scoring}
A confidence score on pose estimates is necessary for evaluation purposes~\citep{bop23} and may be used in practical scenarios such as object grasping. We propose to score the refined poses $\bm{T}^r_{WO}$ based on the average per-view feature-metric loss:
\begin{equation}
s({\bm{T}^r_{WO}}) = 1-\frac{1}{|\mathcal{C}|}\sum_{C \in \mathcal{C}} {\mathcal{L}^{C}_{\text{FE}}(\bm{T}_{CW}\bm{T}^r_{WO})} \, .
\end{equation}
The score ranges from 0 to 1, with higher values indicating better poses, reaching its maximum when the pose best aligns the registered features with the query features.
This score serves as a final confidence measure, quantifying how consistently the refined object pose aligns with the visual evidence across all available camera views.

\section{Experiments}
Implementation details can be found in Appendix~\ref{sec:appendix_implementation_details}, detailed results and ablation studies in Appendix~\ref{sec:appendix_additional_experiments}.

\begin{table}[t]
\renewcommand{\arraystretch}{0.9}
\caption{\textbf{Multi-view pose estimation of unseen objects on YCB-V and T-LESS datasets.} Both our method and CosyPose~\citep{labbe2020cosypose} refine candidates from FoundPose~\citep{ornek2024foundpose}, GigaPose~\citep{nguyen2024gigapose}, MegaPose~\citep{labbe2022megapose}, and Co-Op~\citep{moon2025co} downloaded from the BOP leaderboard. Our approach outperforms all other methods by a large margin. Results with MSPD/MSSD/VSD metrics can be found in Appendix~\ref{sec:appendix_detailed_results}.}
\vspace{-5mm}
\small
\label{table:unseen_results_bop_classic}
\begin{center}
\begin{tabular}{l cc | cc}
\toprule
\multirow{2}{*}{\textbf{Method}} &
\multicolumn{2}{c}{YCB-V} &
\multicolumn{2}{c}{T-LESS} \\
& AR & AP & AR & AP \\
\toprule

FoundPose & 69.0 & 63.0 & 57.0 & 57.0 \\
+ CosyPose MV & 79.2 & 76.1 & 66.4 & 63.0 \\
+ Ours & \textbf{83.9} & \textbf{83.2} & \textbf{84.1} & \textbf{88.6} \\
\midrule

GigaPose & 66.6 & 63.1 & 58.2 & 54.3 \\
+ CosyPose MV & 76.5 & 70.9 & 61.9 & 55.9 \\
+ Ours & \textbf{81.9} & \textbf{79.7} & \textbf{81.9} & \textbf{84.9} \\
\midrule

MegaPose & 62.0 & 56.1 & 50.8 & 50.5 \\
+ CosyPose MV & 71.1 & 64.5 & 57.6 & 56.1 \\
+ Ours & \textbf{80.0} & \textbf{77.3} & \textbf{78.8} & \textbf{81.4} \\
\midrule

Co-op & 69.7 & 69.5 & 68.2 & 68.9 \\
+ CosyPose MV & 81.0 & 79.2 & 78.7 & 78.9 \\
+ Ours & \textbf{83.8} & \textbf{83.3} & \textbf{89.6} & \textbf{92.4} \\
\bottomrule

\end{tabular}
\end{center}
\vspace{-7mm}
\end{table}
 
\paragraph{Evaluation.}
We follow the BOP evaluation protocol~\citep{hodavn2020bop} and use three pose-error functions: Maximum Symmetry-Aware Surface Distance (MSSD), which measures the maximum 3D distance between corresponding object vertices accounting for symmetries; Maximum Symmetry-Aware Projection Distance (MSPD), which computes the maximum 2D distance between the projected vertices of the predicted and ground-truth poses in the image plane; and Visible Surface Discrepancy (VSD), which measures the difference between visible object surfaces by comparing rendered depth maps, considering occlusions.

We evaluate our method on two tasks: 6D object localization, measured by Average Recall (AR) and 6D object detection, measured by Average Precision (AP) (see \citet{hodavn2020bop}[A.1]). The AR metric measures the fraction of object instances for which a correct pose is found, averaged across several error thresholds. The AP metric follows an evaluation methodology similar to the COCO challenge \citep{lin2014microsoft} but it replaces the standard Intersection over Union (IoU) with the MSSD and MSPD pose errors. AP is generally a stricter metric than AR. See \citet{van2025bop} for more details. For each method, we measure the average recall for each error function ($\text{AR}\textsubscript{VSD},\text{AR}\textsubscript{MSSD}, \text{AR}\textsubscript{MSPD}$), their average $ \text{AR}= (\text{AR}\textsubscript{VSD} +  \text{AR}\textsubscript{MSSD} + \text{AR}\textsubscript{MSPD}) / 3$, as well as the average precision for two error functions
($\text{AP}\textsubscript{MSSD}, \text{AP}\textsubscript{MSPD}$), 
and their average $ \text{AP}=( \text{AP}\textsubscript{MSSD}+\text{AP}\textsubscript{MSPD})/2$. Note that for all the quantitative results in the main paper, we report just the average metrics; results for individual metrics can be found in Appendix~\ref{sec:appendix_detailed_results}. Additionally, BOP benchmark does not provide AR evaluation for the BOP-Industrial datasets, thus we report only AP.

\paragraph{Datasets.}\label{section:datasets}
We evaluate our method on six datasets.
From the BOP-Classic datasets, we use YCB-V~\citep{xiang2018posecnn} and T-LESS~\citep{hodan2017tless}. YCB-V contains textured household objects, while T-LESS focuses on industry-relevant, texture-less objects, often arranged in heavily cluttered scenes. Next, we include three BOP-Industrial datasets: ITODD~\citep{itodd}, IPD~\citep{ipd}, and XYZ-IBD~\citep{xyzibd}. These datasets consist of small, metallic, and reflective objects captured in grayscale, reflecting realistic industrial conditions. For ITODD, we use the ITODD-MV variant provided by BOP, which includes multi-view data. In addition, we evaluate our method on HouseCat6D~\cite{jung2024housecat}, which contains household objects including challenging metallic (cutlery) and transparent (glass) objects.
For YCB-V, T-LESS, and HouseCat6D, we sample view groups of 4 views from the test scenes following the same procedure as \citep{labbe2020cosypose} for a fair comparison.
In contrast, BOP-Industrial datasets provide predefined views for each scene, so no sampling is necessary. All datasets include camera calibration.

\begin{table}[t]
\renewcommand{\arraystretch}{0.9} 
\caption{\textbf{Multi-view pose estimation of unseen objects for HouseCat6D dataset and for its metallic and glass subsets}. We generate 2D detections by FOSOD (our implementation of \cite{lu2024adapting}) and predict pose candidates using FoundPose~\citep{ornek2024foundpose}.}
\label{table:unseen_results_housecat}
\vspace{-4mm}
\small
\begin{center}
\begin{tabular}{l c c | c c | c c}
\toprule
\multirow{2}{*}{\textbf{Method}} &
\multicolumn{2}{c}{HouseCat} &
\multicolumn{2}{c}{glass} &
\multicolumn{2}{c}{metallic} \\
& AR & AP & AR & AP & AR & AP \\
\midrule
\hspace{-0.5mm}FoundPose & 80.3 & 72.2 & 75.1 & 71.4 &  19.3 & 19.0  \\
\hspace{-0.5mm}+ CosyPose MV & 86.7 & 86.6 & 84.6 &  86.8 & 25.8 & 27.1 \\
\hspace{-0.5mm}+ Our & \textbf{88.9} & \textbf{89.6} & \textbf{88.9} & \textbf{92.8} &  \textbf{43.5} & \textbf{46.5} \\
\bottomrule
\end{tabular}
\end{center}
\vspace{-5mm}
\end{table}

\begin{table}[t]
\renewcommand{\arraystretch}{0.9} 
\centering
\caption{\textbf{Multi-view pose estimation of unseen objects on BOP-Industrial datasets.} We obtain 2D detection from 3PT~\citep{3pt}, generate 2D segmentations by SAM2~\citep{ravi2024sam} and predict pose candidates using FoundPose~\citep{ornek2024foundpose}.}
\label{table:unseen_results_bop_industrial}
\vspace{-3mm}
\small
\begin{tabular}{lc|c|c}
\toprule
\multirow{2}{*}{\textbf{Method}} & IPD & XYZ-IBD & ITODD-MV \\
& AP & AP & AP\\ 
\midrule
FoundPose & 31.4 & 32.5 & 41.2 \\
+ CosyPose MV & 36.7 & 52.4 & 54.5 \\
+ Ours & \textbf{79.8} & \textbf{66.5} & \textbf{76.8} \\
\bottomrule
\end{tabular}
\end{table}

\setlength{\fboxsep}{0pt}
\begin{figure*}[tbp]
    \centering
    \begin{tabular}{@{}c@{}c@{}c@{}c@{}}
        \small
        \textbf{Input image} & \textbf{Single-view} & \textbf{CosyPose MV} & \textbf{Ours} \\
        \rotatebox{90}{\hspace{10mm} \small{YCB-V}}\hspace{1.0mm}\includegraphics[width=0.24\linewidth, frame]{./imgs/qualitative/ycbv/ycbv_001094.png} & 
        \colorbox{gray!80}{\includegraphics[width=0.24\linewidth, trim=60pt 30pt 60pt 60pt, clip, frame]{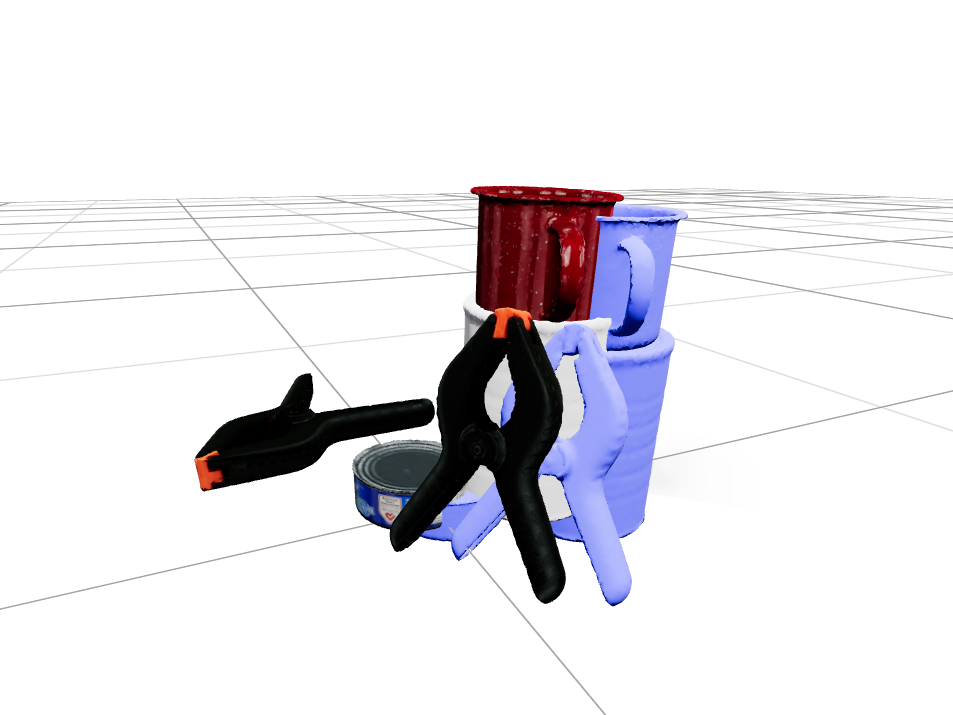}} & 
         \colorbox{gray!80}{\includegraphics[width=0.24\linewidth, trim=60pt 30pt 60pt 60pt, clip,  frame]{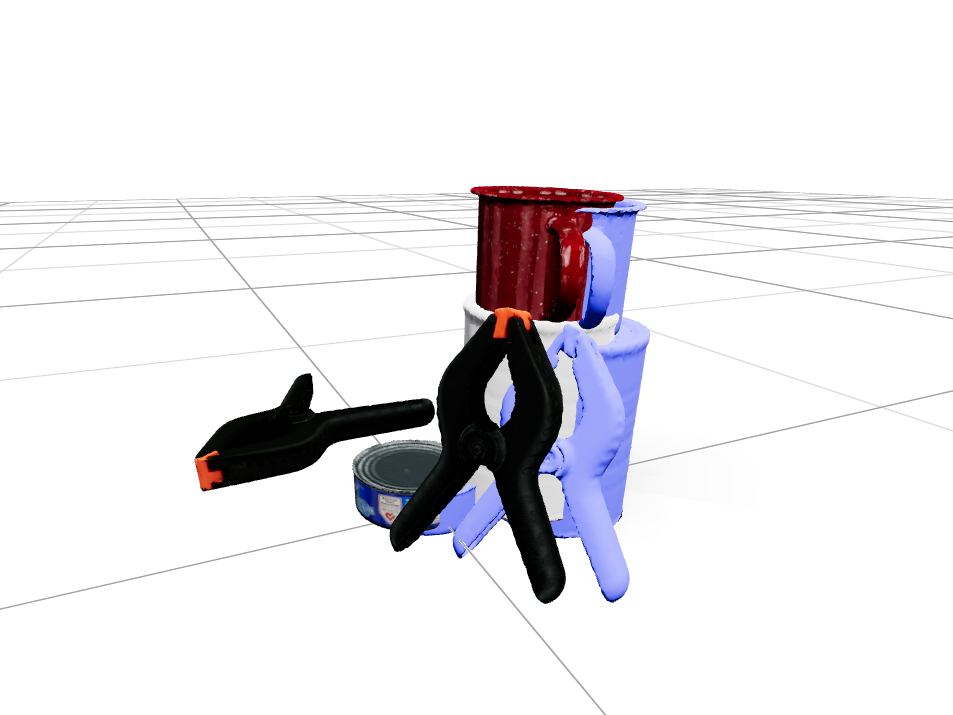}} & 
        \colorbox{gray!80}{\includegraphics[width=0.24\linewidth,trim=60pt 30pt 60pt 60pt, clip,  frame]{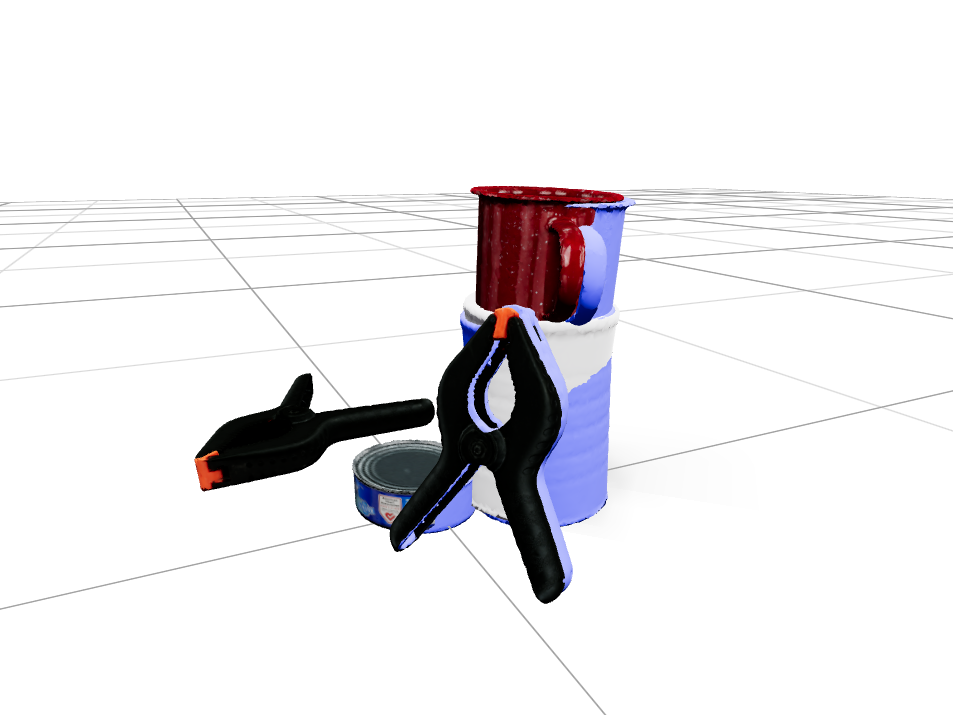}} \\

        \rotatebox{90}{\hspace{10mm} \small{YCB-V}}\hspace{1.0mm}\includegraphics[width=0.24\textwidth, frame]{./imgs/qualitative/ycbv/ycbv_000946.png} & 
        \colorbox{gray!80}{\includegraphics[width=0.24\textwidth, trim=40pt 60pt 80pt 30pt, clip, frame]{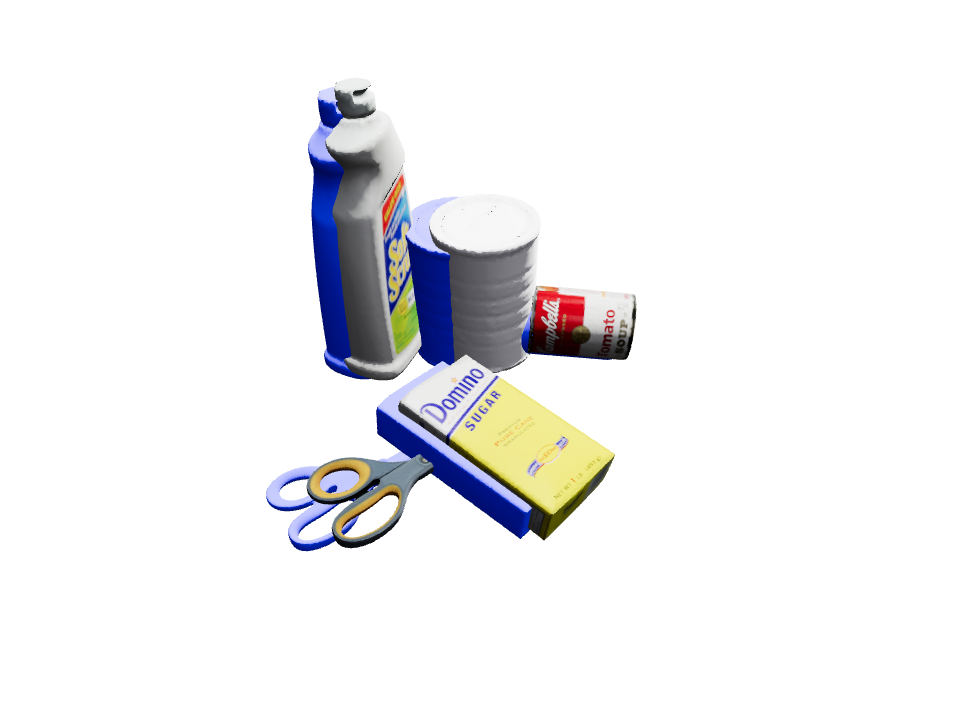}} & 
        \colorbox{gray!80}{\includegraphics[width=0.24\textwidth, trim=40pt 60pt 80pt 30pt, clip, frame]{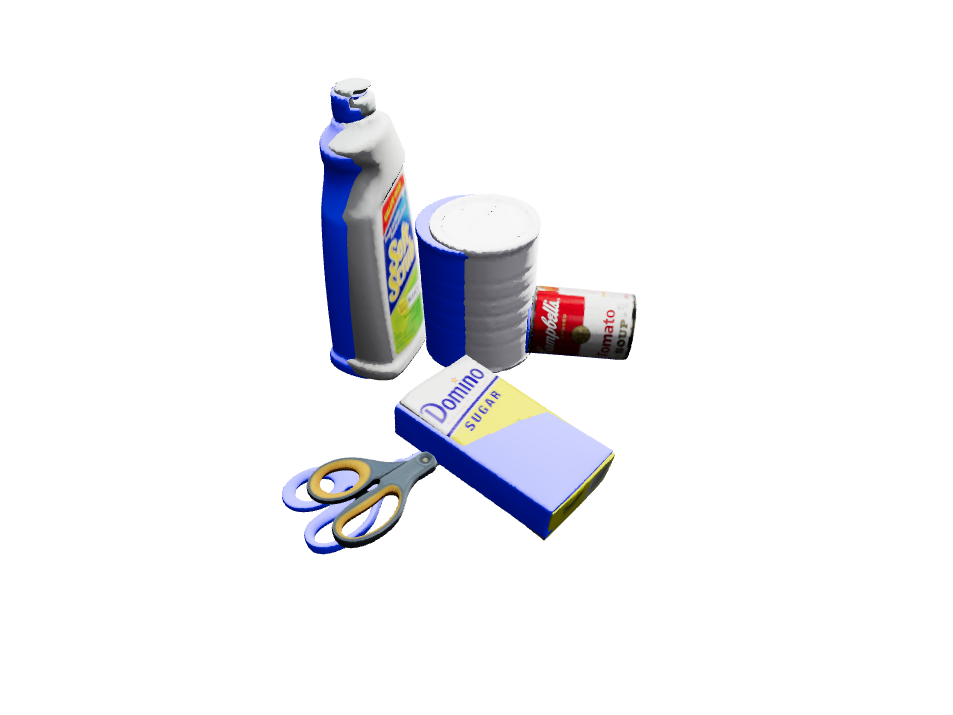}} & 
        \colorbox{gray!80}{\includegraphics[width=0.24\textwidth, trim=40pt 60pt 80pt 30pt, clip, frame]{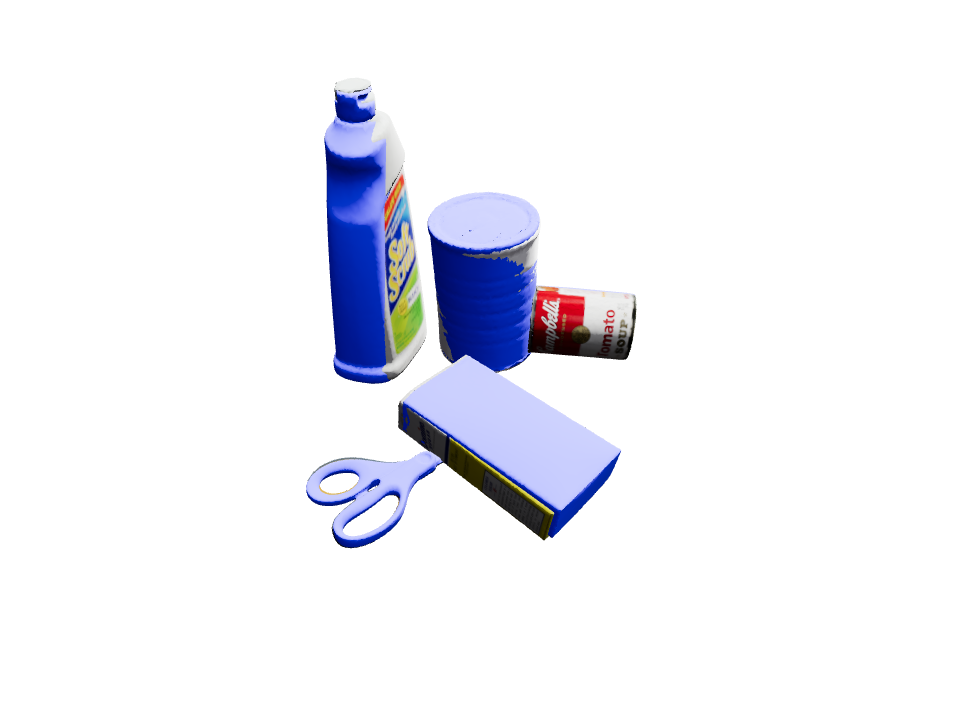}} \\
        
        \rotatebox{90}{\hspace{6mm} \small{HouseCat6D}}\hspace{1.0mm}\includegraphics[width=0.24\linewidth, frame]{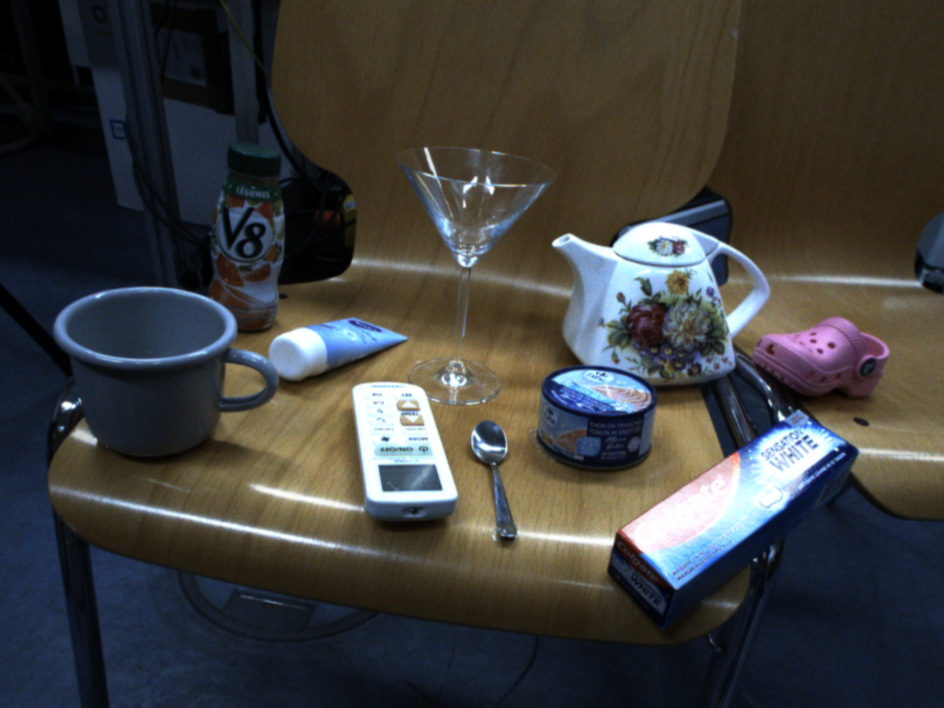} & 
        \colorbox{gray!80}{\includegraphics[width=0.24\linewidth,  frame]{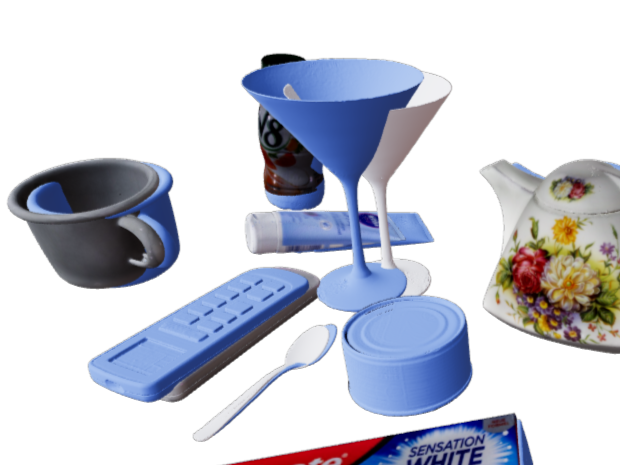}} & 
        \colorbox{gray!80}{\includegraphics[width=0.24\linewidth, frame]{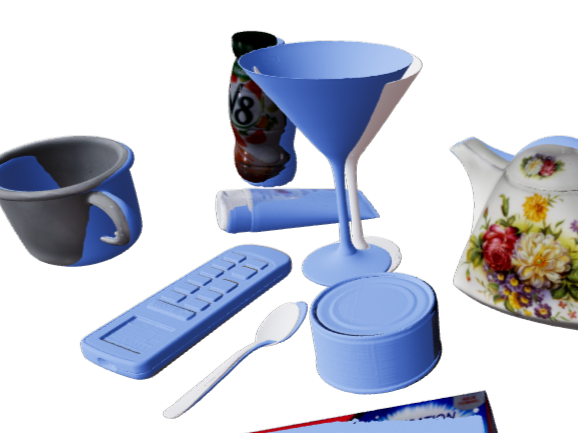}} & 
        \colorbox{gray!80}{\includegraphics[width=0.24\linewidth, frame]{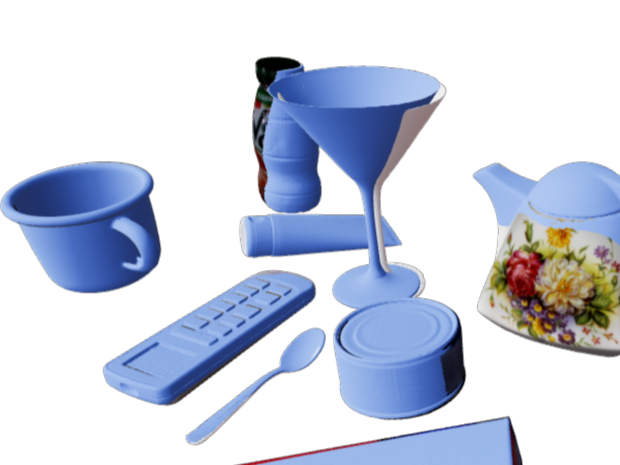}} \\

        \rotatebox{90}{\hspace{6mm} \small{HouseCat6D}}\hspace{1.0mm}\includegraphics[width=0.24\linewidth, frame]{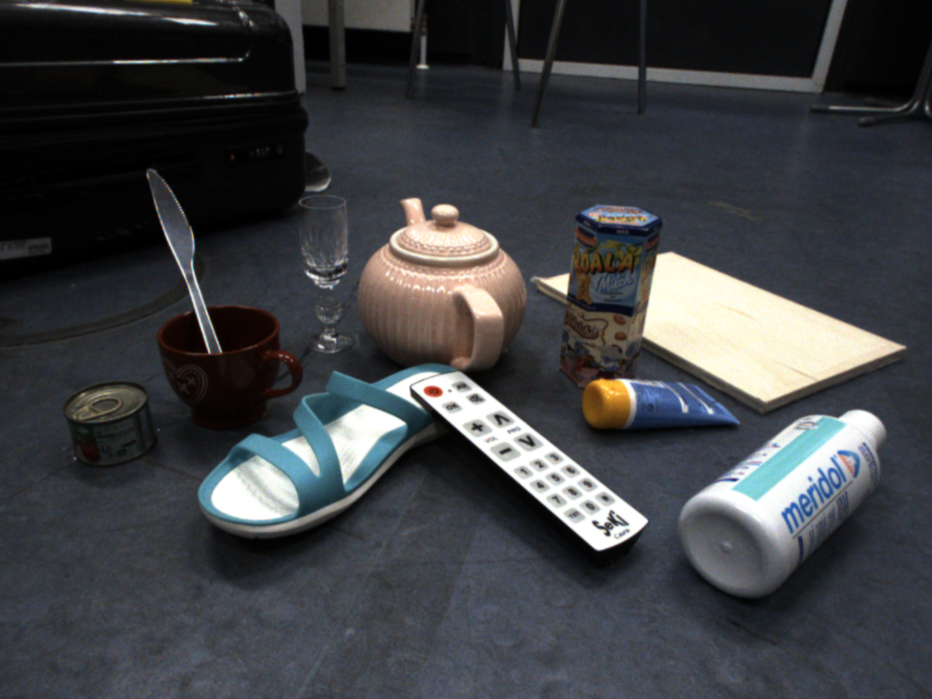} & 
        \colorbox{gray!80}{\includegraphics[width=0.24\linewidth,  frame]{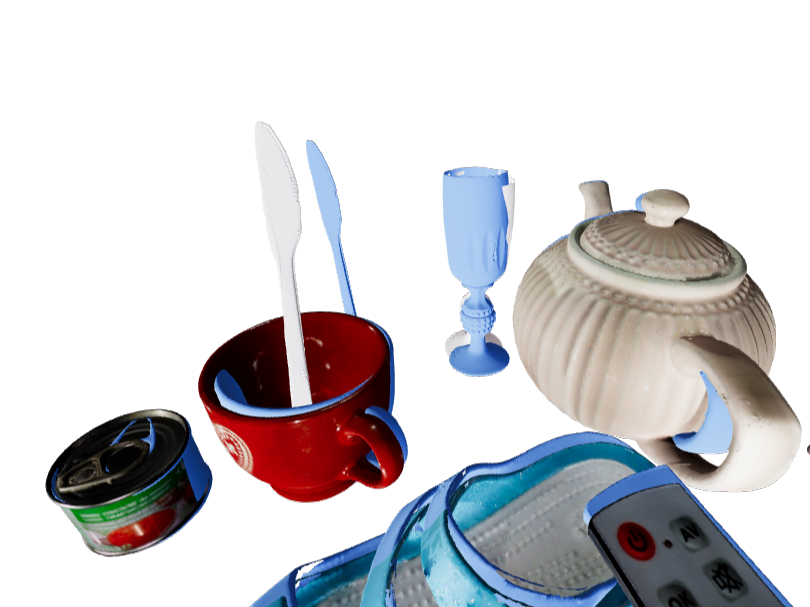}} & 
        \colorbox{gray!80}{\includegraphics[width=0.24\linewidth, frame]{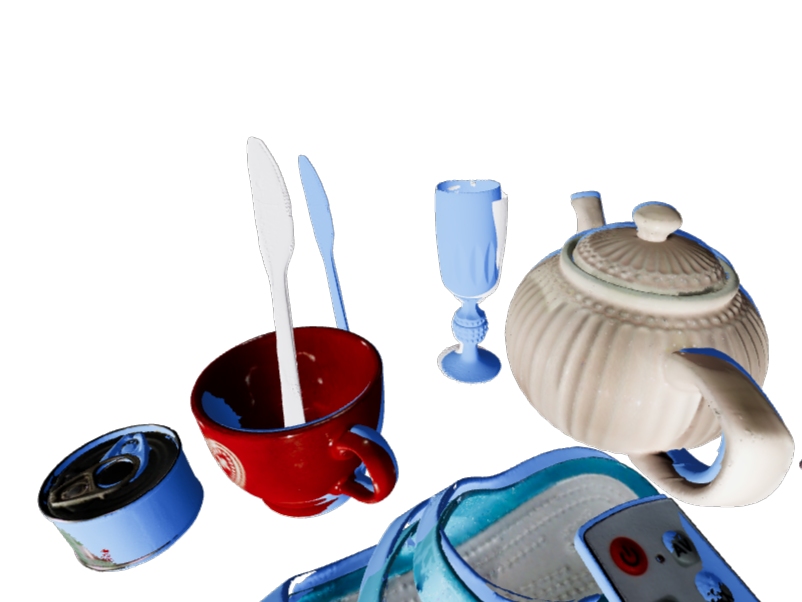}} & 
        \colorbox{gray!80}{\includegraphics[width=0.24\linewidth,  frame]{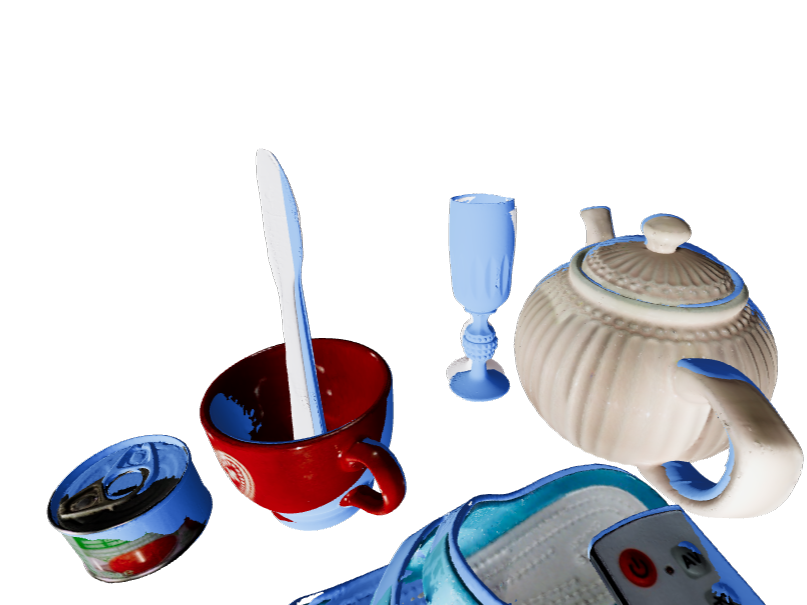}} \\
    \end{tabular}
        \vspace*{-2mm}
    \caption{\textbf{Qualitative results of multi-view refinement on the YCB-V and HouseCat6D datasets.}  1st column: one of the input images, 2nd column: single-view pose estimates in blue, ground-truth shown using textured models. 3rd column: results of CosyPose multi-view baseline. 4th column: results of our multi-view method. CosyPose and our method both use single-view pose candidates from four input views. Please see the major improvements by our method: "clamp" (top row), "scissors" (2nd row), "glass" (3rd row), "knife" (4th row).}
    \label{fig:ycbv_qualitative}
    \vspace{-4mm}
\end{figure*}

\begin{figure*}[tbp]
    \centering
    \begin{tabular}{@{}c@{}c@{}c@{}c@{}}
        \small
        \textbf{Input image} & \textbf{Single-view} & \textbf{CosyPose MV} & \textbf{Ours} \\
        
        \includegraphics[width=0.24\linewidth, frame]{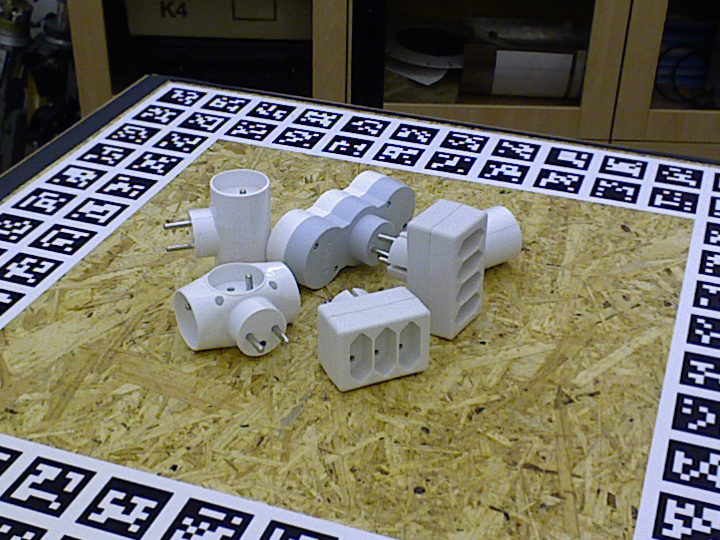} & 
        {\includegraphics[width=0.24\linewidth, frame]{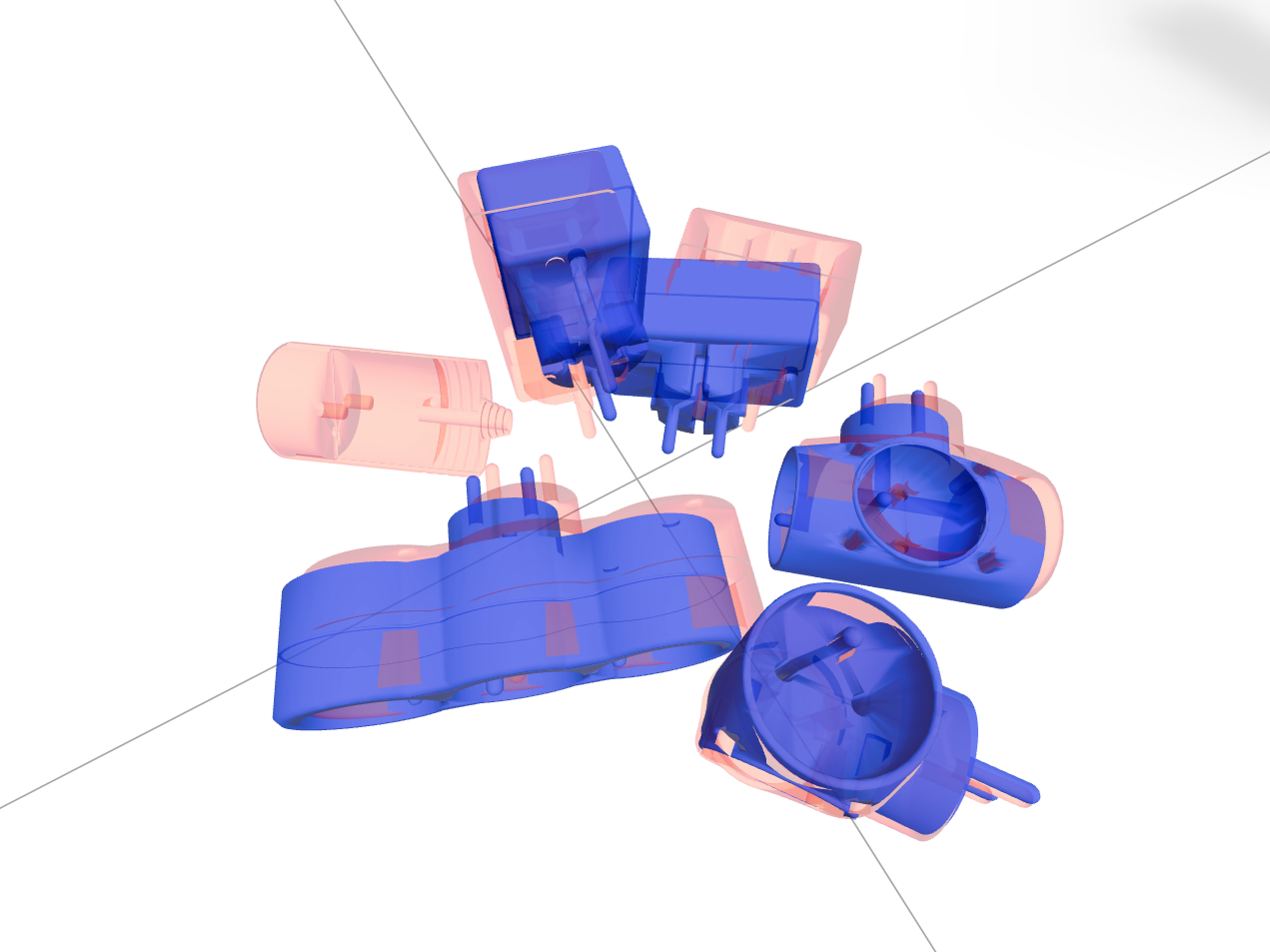}} & 
        {\includegraphics[width=0.24\linewidth, frame]{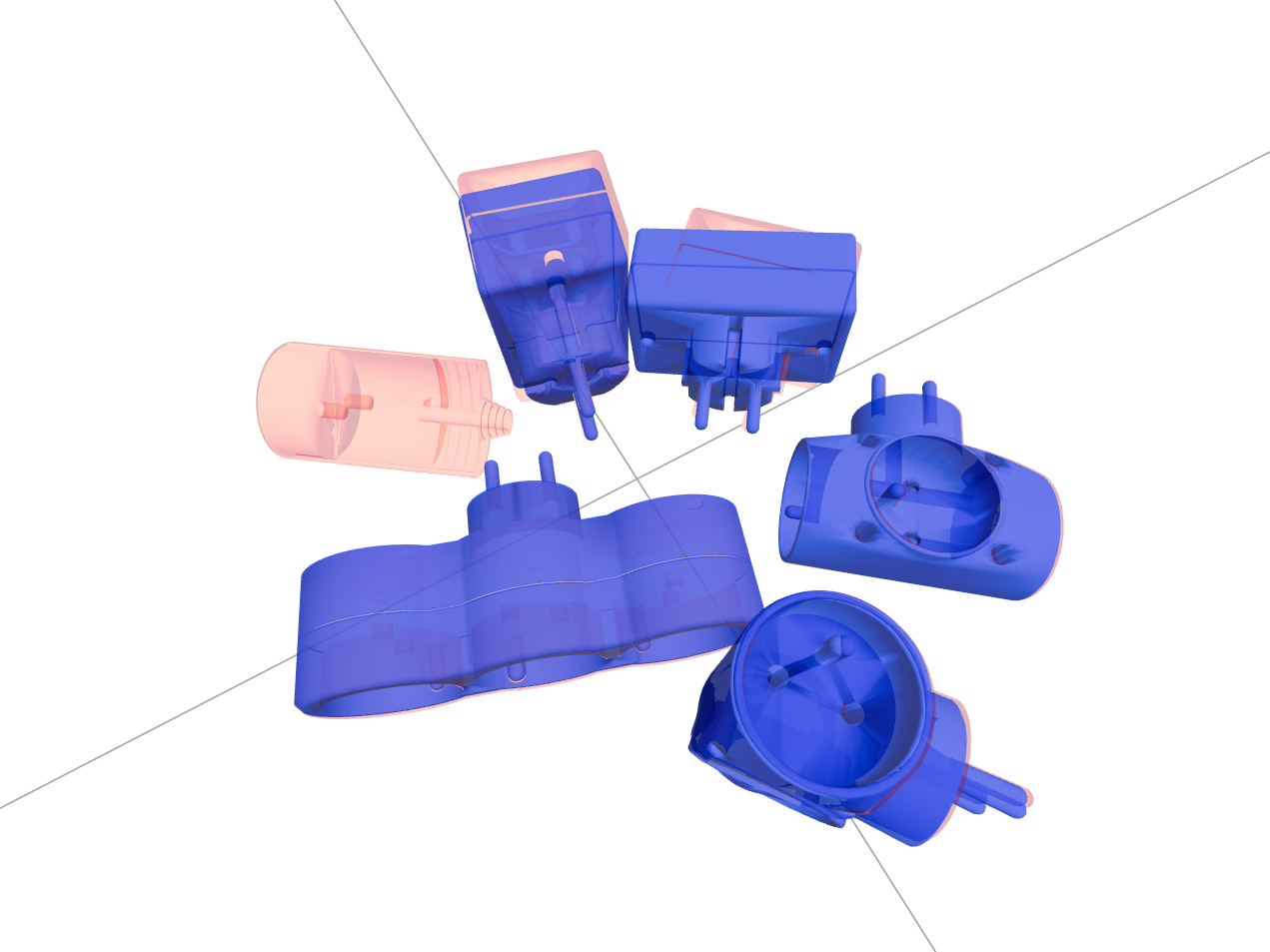}} & 
        {\includegraphics[width=0.24\linewidth, frame]{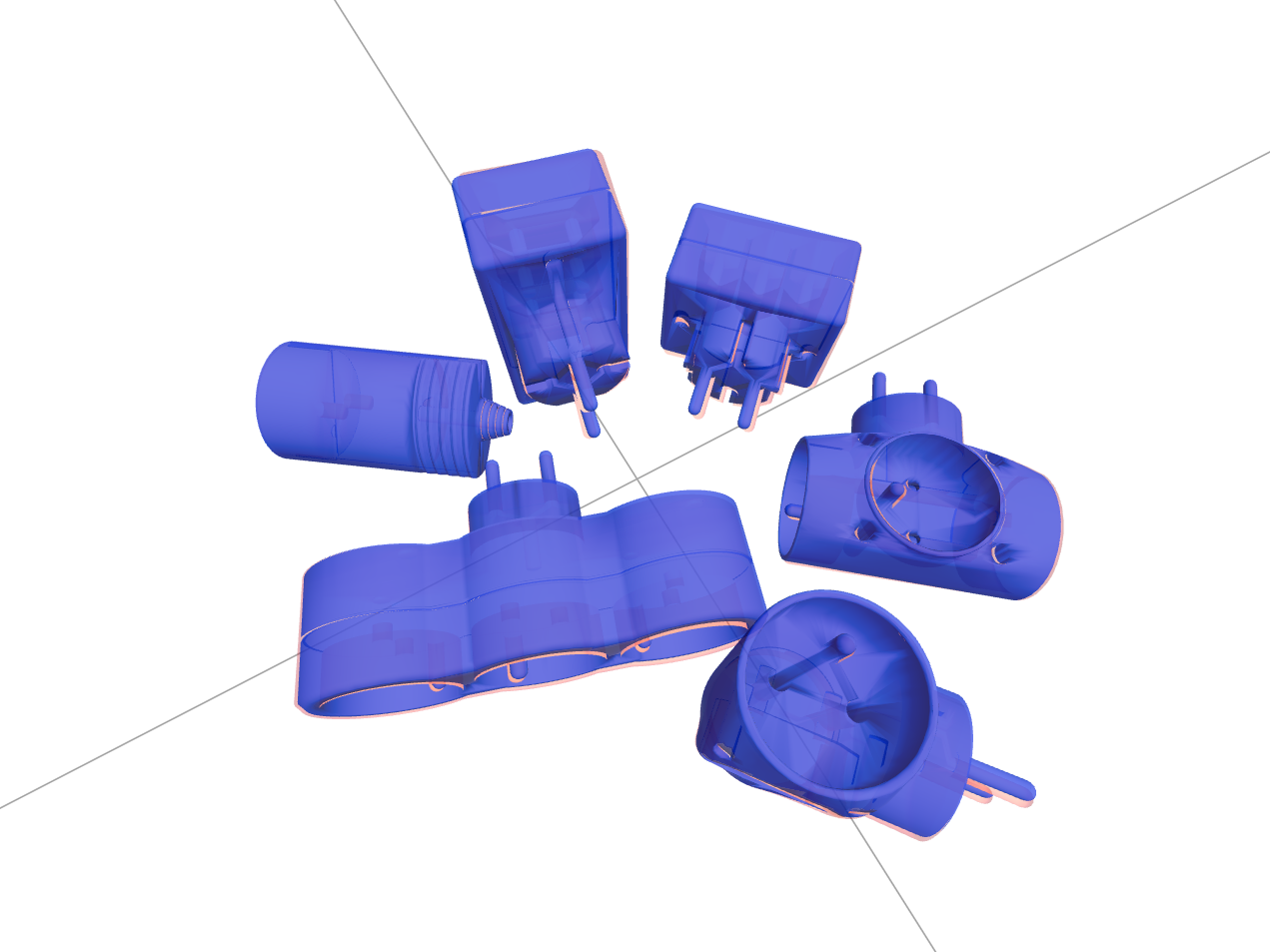}} \\
        
        \includegraphics[width=0.24\linewidth, frame]{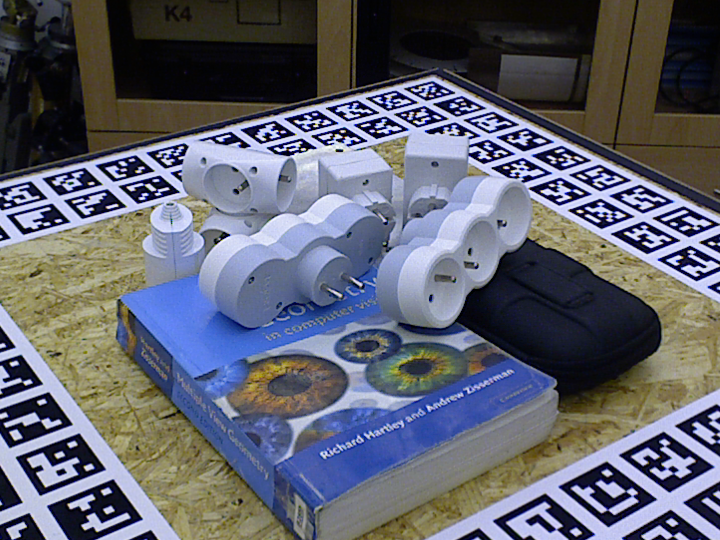} & 
        {\includegraphics[width=0.24\linewidth, frame]{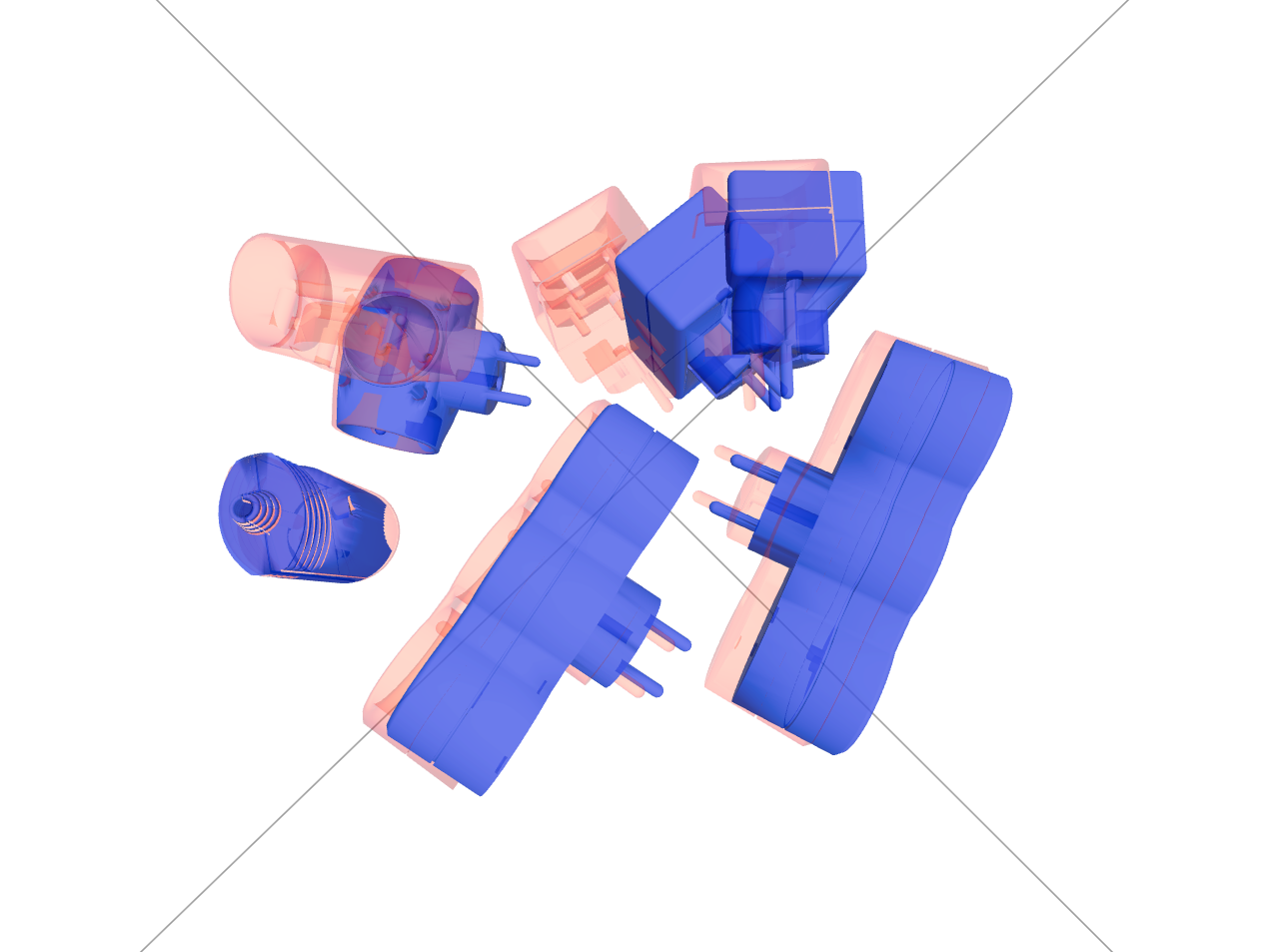}} & 
        {\includegraphics[width=0.24\linewidth, frame]{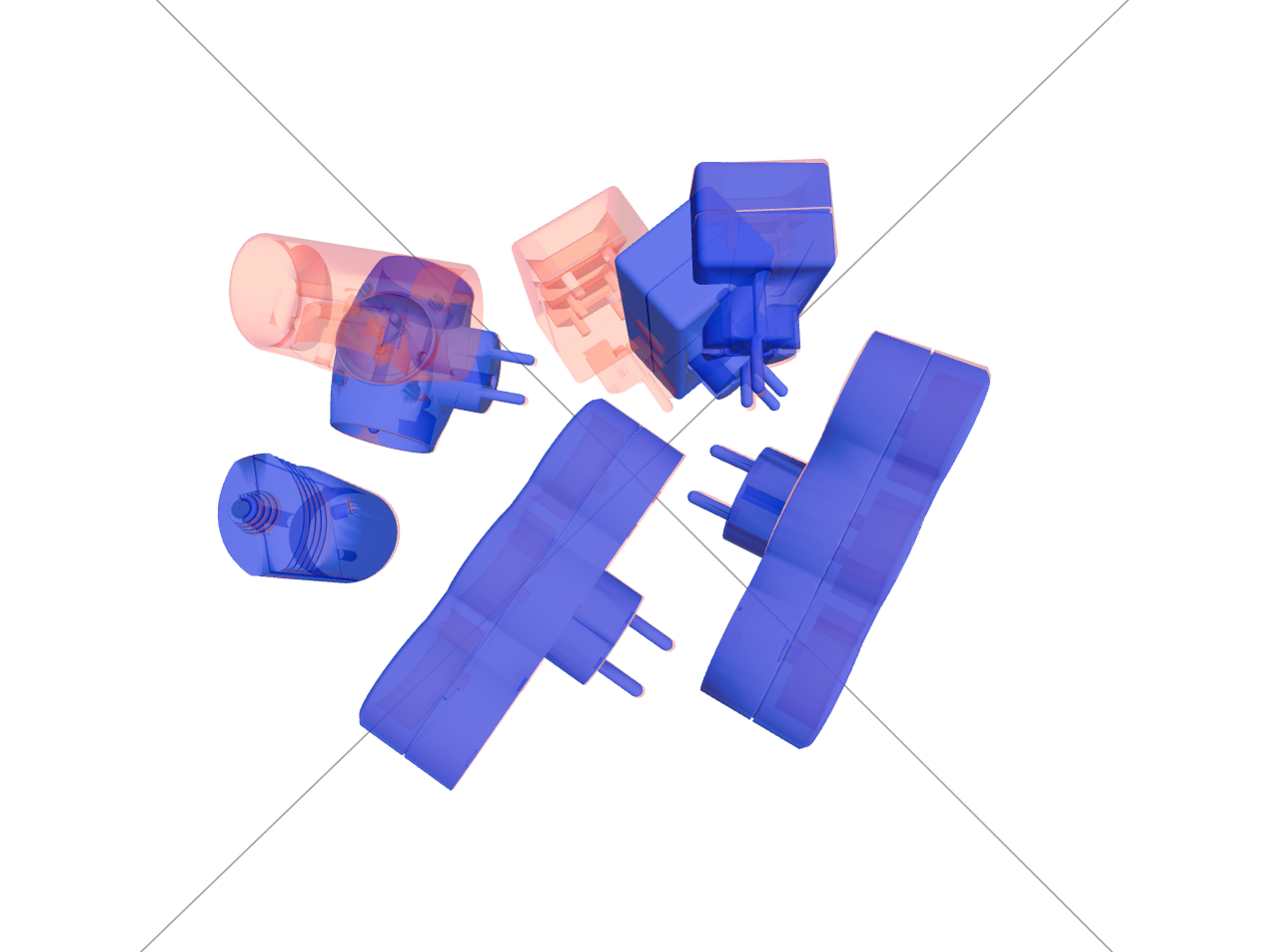}} & 
        {\includegraphics[width=0.24\linewidth, frame]{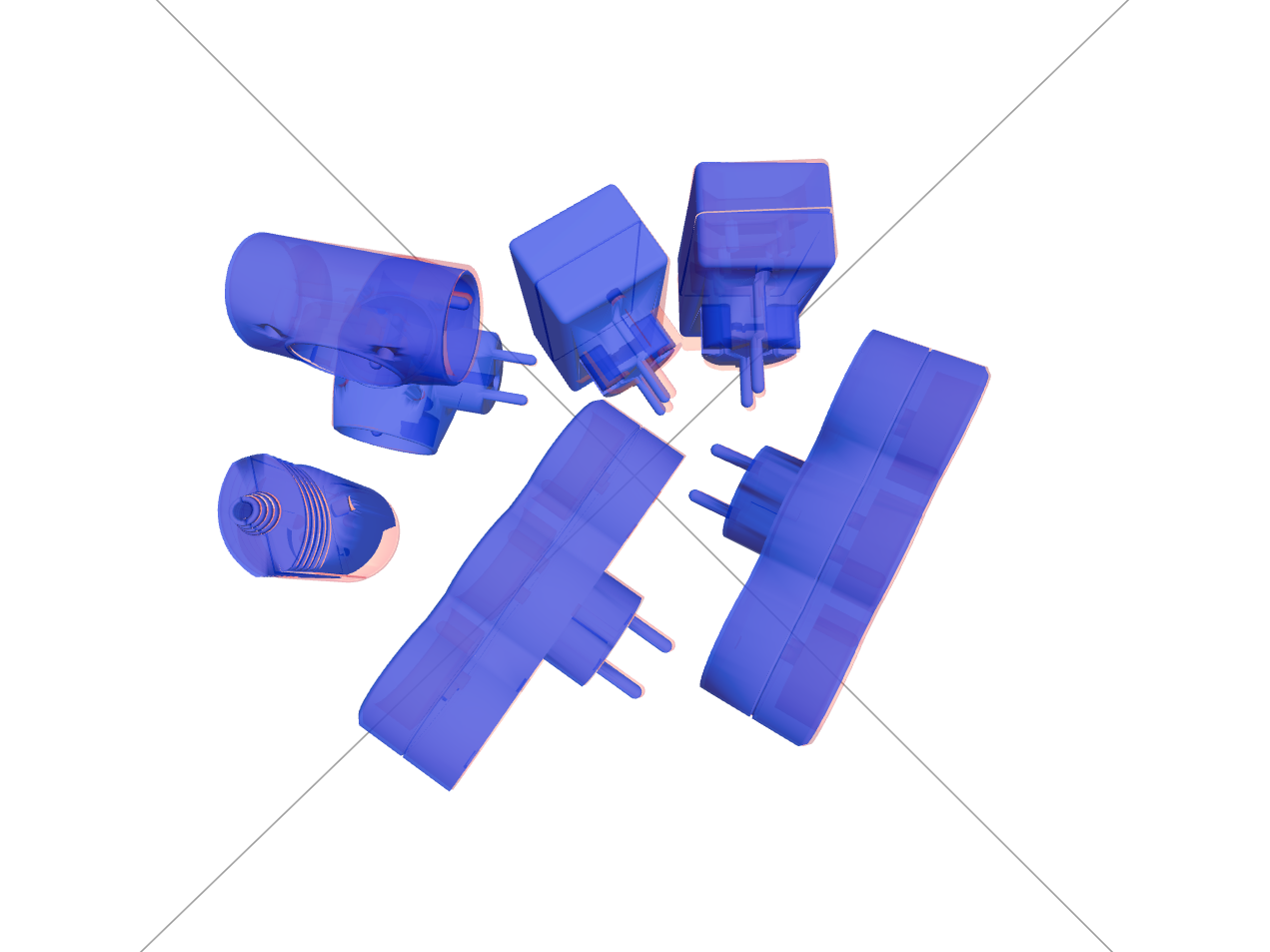}} \\
        
        \includegraphics[width=0.24\linewidth, frame]{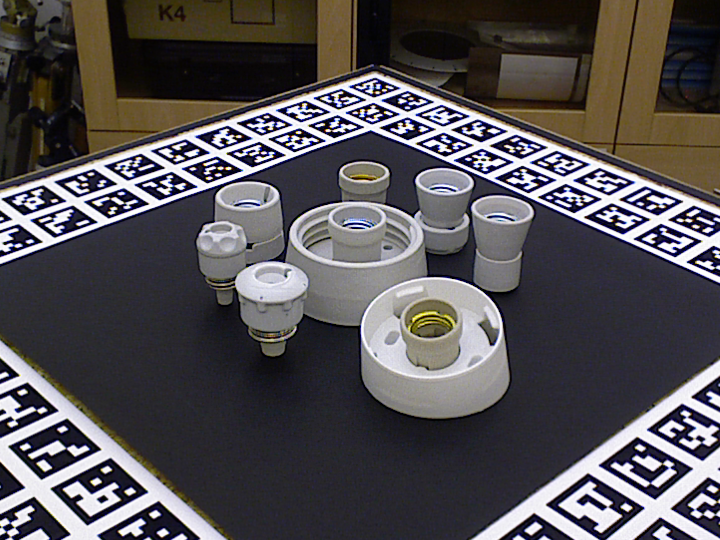} & 
        {\includegraphics[width=0.24\linewidth, frame]{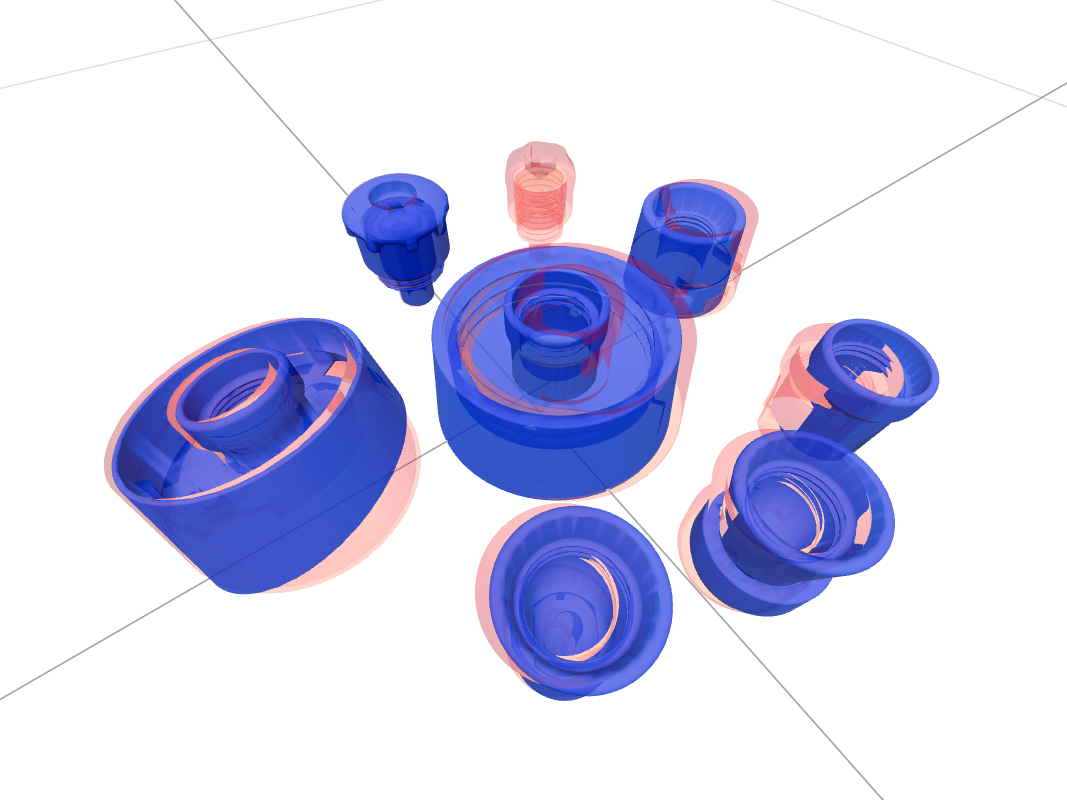}} & 
        {\includegraphics[width=0.24\linewidth, frame]{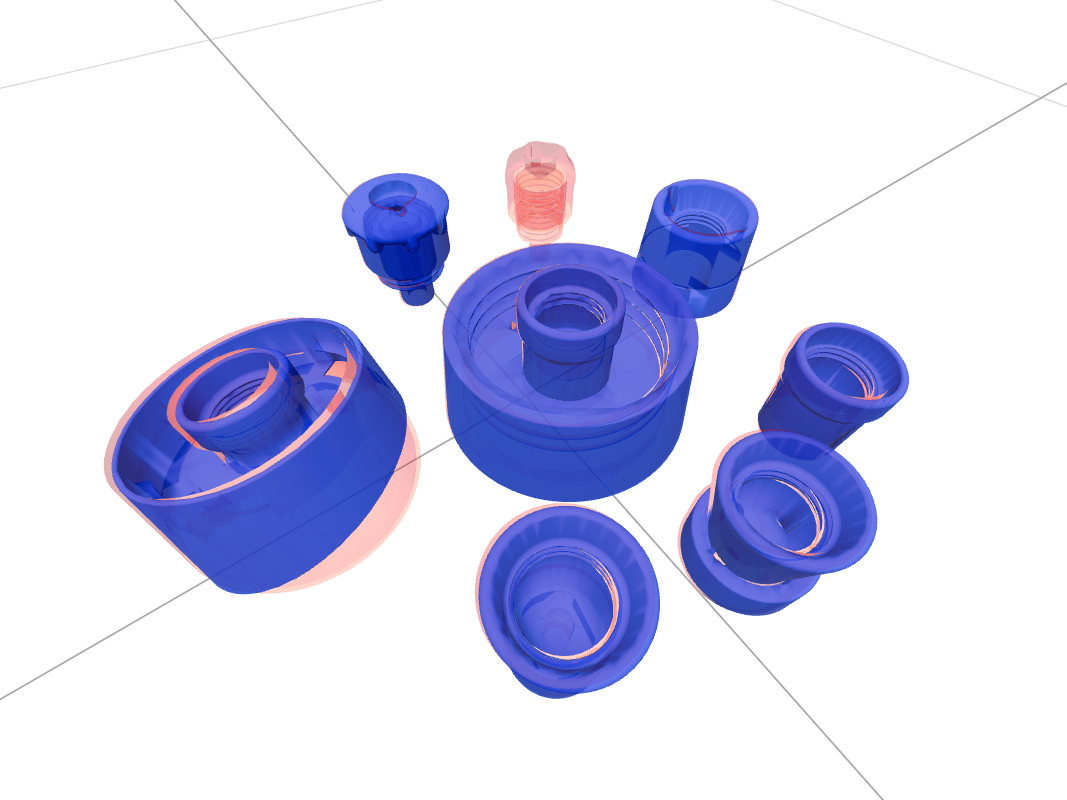}} & 
        {\includegraphics[width=0.24\linewidth, frame]{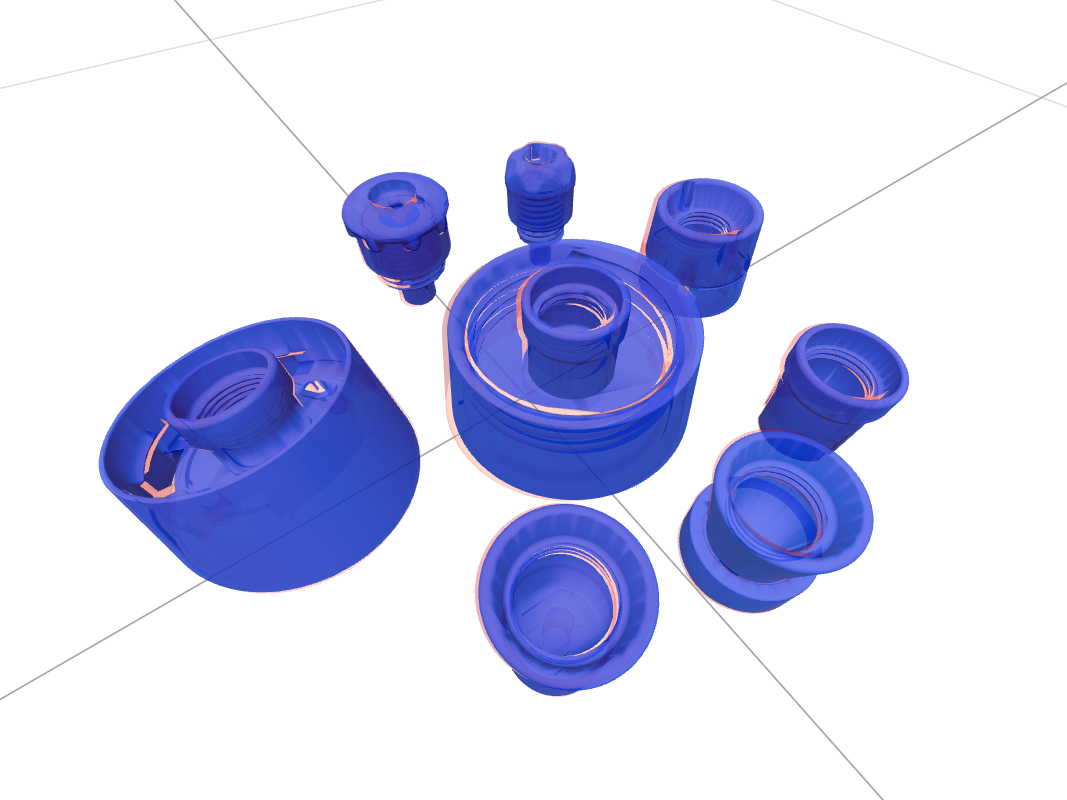}} \\
        
        \includegraphics[width=0.24\linewidth, frame]{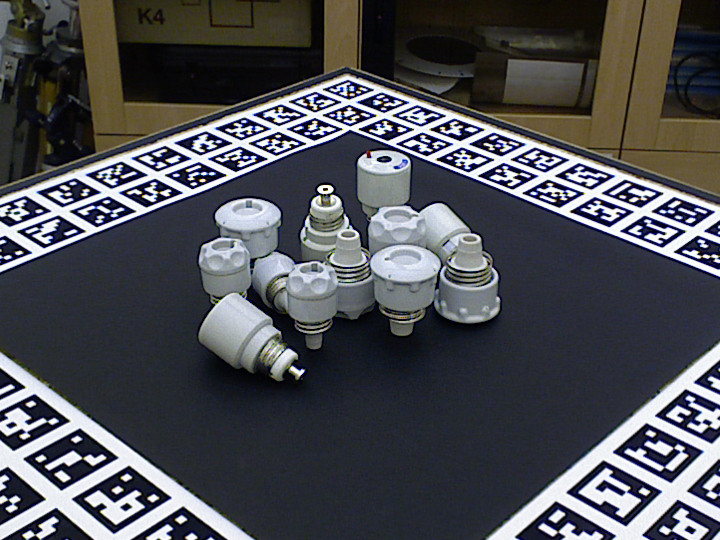} & 
        {\includegraphics[width=0.24\linewidth, trim=40pt 120pt 160pt 30pt, clip, frame]{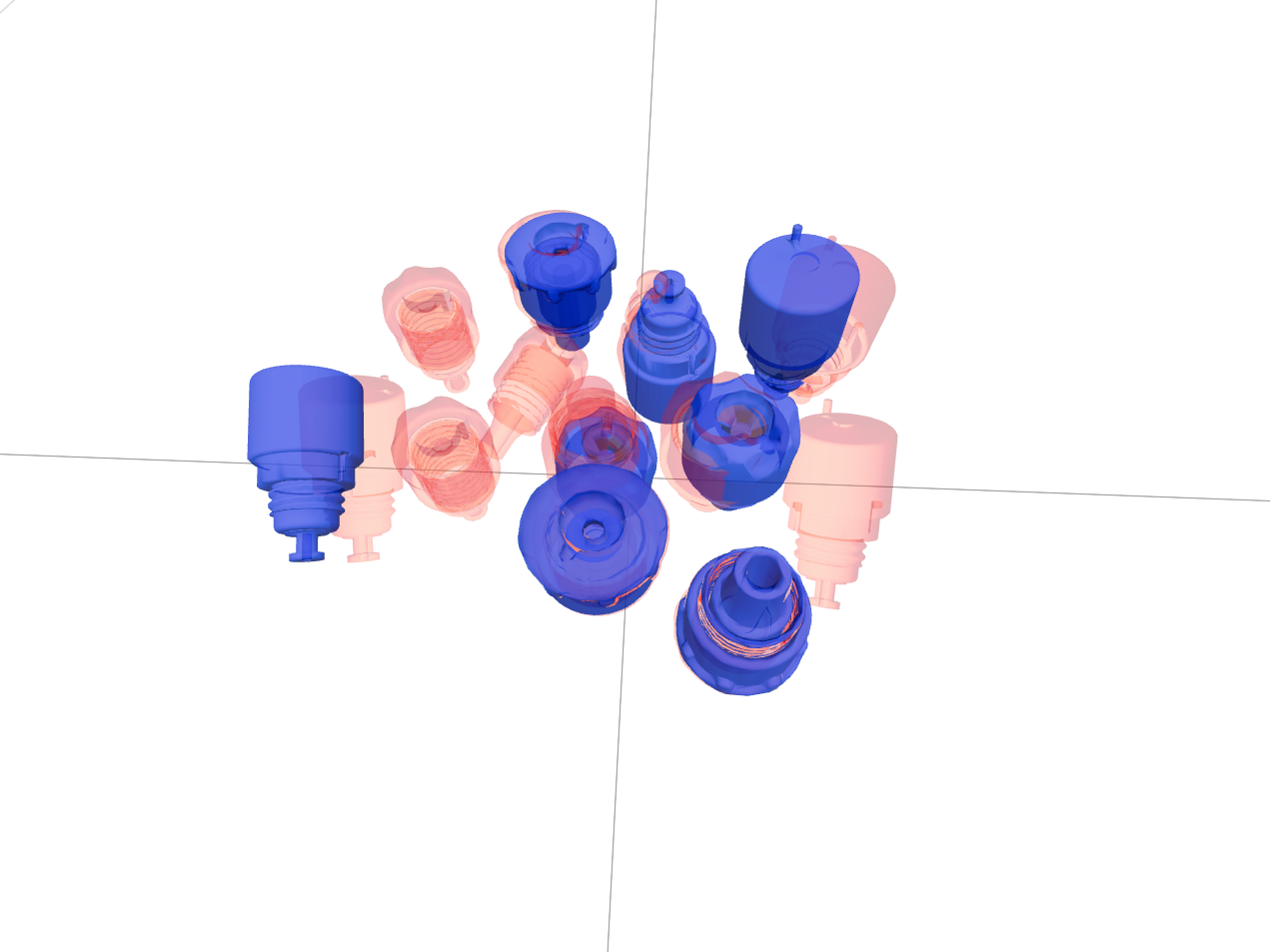}} & 
        {\includegraphics[width=0.24\linewidth, trim=40pt 120pt 160pt 30pt, clip, frame]{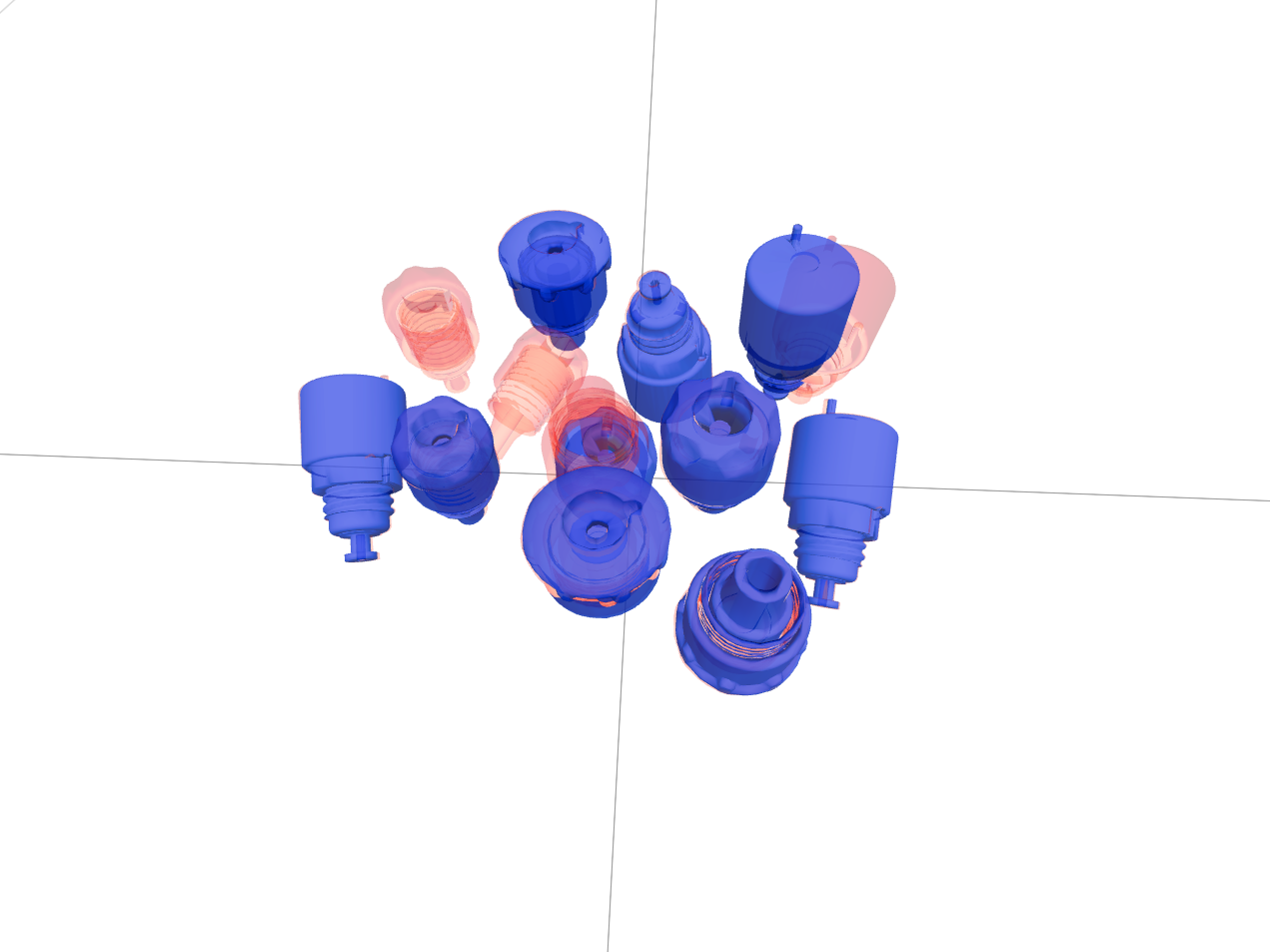}} & 
        {\includegraphics[width=0.24\linewidth, trim=40pt 120pt 160pt 30pt, clip, frame]{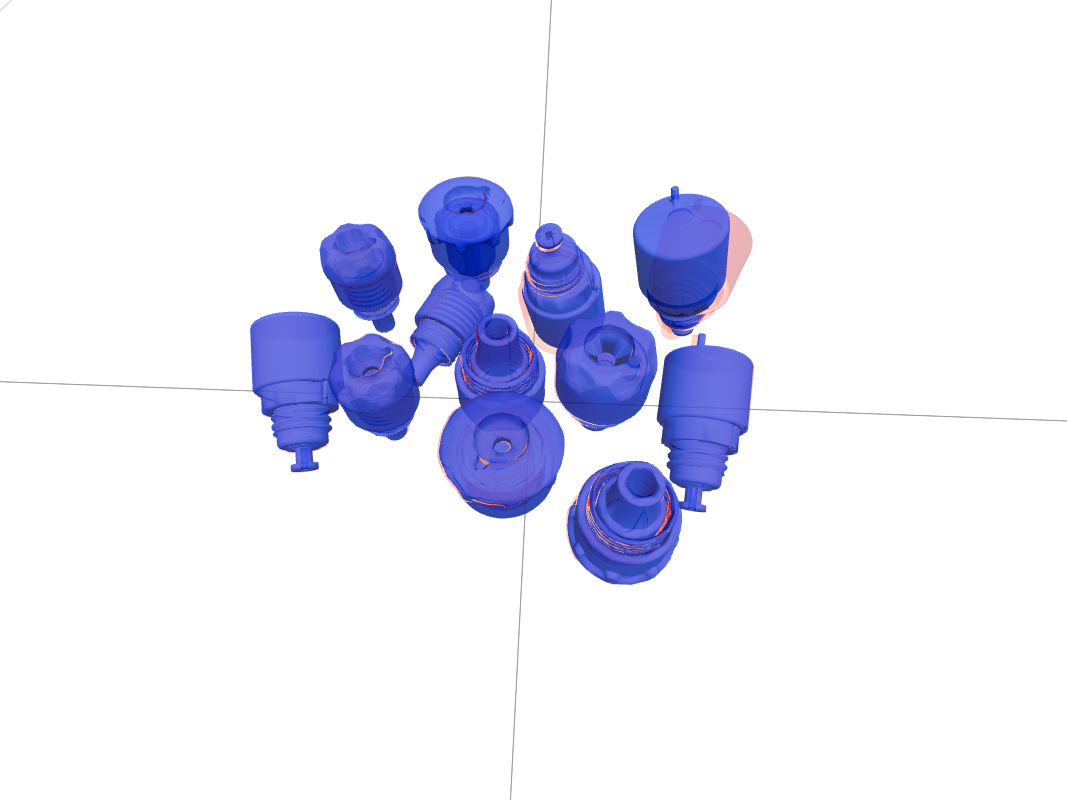}} \\

    \end{tabular}
    \vspace*{-2mm}
    \caption{\textbf{Qualitative results of multi-view refinement on the T-LESS dataset.} 1st column: one of the four input images. 2nd column: single-view pose estimates obtained by MegaPose~\citep{labbe2022megapose} in blue; ground-truth poses in red. 3rd column: results of CosyPose multi-view baseline. 4th column: results of our multi-view pose estimation method. CosyPose and our method both use MegaPose single-view pose candidates from four input views/images. In this challenging setting with multiple textureless objects our approach outputs significantly better poses than CosyPose (see rows 1-2) and is able to correctly obtain poses for more objects (see rows 1-4). }
    \label{fig:tless_qualitative}
    \vspace{-3mm}
\end{figure*}

\subsection{Unseen object pose estimation task}
Our work introduces a generalizable multi-view pose estimation method, capable of operating on novel objects without object-specific training. To validate this generalization capability, we evaluate our approach in the unseen object pose estimation setting defined by \citep{bop23}. In this setting, methods are allowed to onboard new objects within a short time budget (5 minutes on 1 GPU) using their 3D models.

\paragraph{Single-view estimates.}
To obtain input single-view pose candidates for YCB-V and T-LESS, we use state-of-the-art single-view pose estimation methods that operate on unseen objects: FoundPose~\citep{ornek2024foundpose}, GigaPose~\citep{nguyen2024gigapose}, MegaPose~\citep{labbe2022megapose} and Co-Op~\citep{moon2025co}. Specifically, we download their official results from
the BOP leaderboard.
For HouseCat6D and BOP-Industrial datasets, we do not have access to pre-computed single-view pose estimates: HouseCat6D is not part of the BOP challenge, and for the BOP-Industrial datasets, single-view public estimates are only available for one of the cameras used for multi-view pose estimation. We therefore generate single-view pose estimates using FoundPose~\cite{ornek2024foundpose} followed by Megapose refinement~\cite{labbe2022megapose}. Please refer to Appendix~\ref{sec:appendix_single-view} for implementation details and information about object 2D detection.

\paragraph{Multi-view baseline.}
We compare our multi-view pose estimation method against the multi-view aggregation-and-refinement strategy proposed in CosyPose \citep{labbe2020cosypose}. To our knowledge, this is the only published and open-source method capable of multi-view pose refinement of unseen objects from RGB images. Unlike our method, CosyPose jointly refines object and camera poses and does not require known camera extrinsics. To ensure a fair comparison, we evaluate CosyPose in a setting where camera extrinsics are known and the refinement is applied only to object poses. Additionally, CosyPose relies on annotated object symmetries, while our method is symmetry-agnostic.

\paragraph{Evaluation results.}
The results for the BOP-Classic, HouseCat6D, and BOP-Industrial datasets are shown in \cref{table:unseen_results_bop_classic}, \ref{table:unseen_results_housecat}, and \ref{table:unseen_results_bop_industrial}, respectively. The tables show that our multi-view pose estimation method significantly outperforms CosyPose multi-view on all datasets. For more detailed results and discussion of individual metrics, please refer to Appendix \ref{sec:appendix_detailed_results}.
Notably, we demonstrate significant improvements on the particularly challenging BOP-Industrial datasets and the metallic object classes of HouseCat6D. This is because our approach aggregates information from multiple viewpoints in an accurate and robust manner. 

\paragraph{Qualitative results.}
Qualitative results for the YCB-V and HouseCat6D datasets are shown in \cref{fig:ycbv_qualitative} and for T-LESS in \cref{fig:tless_qualitative}.
Our qualitative results demonstrate the superior performance of our multi-view method over both the initial single-view estimates and the CosyPose refinement. For each scene, we compare the initial single-view pose candidates against the refined multi-view outputs from both CosyPose and our approach.
A key advantage of our method is its robustness, \eg, CosyPose often fails to refine pose estimates when its RANSAC-based aggregation step does not find corresponding estimates from multiple views. In contrast, our method successfully refines any object pose detected in at least one valid view, showcasing a significant improvement in reliability. \textbf{More qualitative results are provided in Appendix~\ref{sec:appendix_qualitative_results}.}

\begin{table}[tbp]
\renewcommand{\arraystretch}{0.85} 
\centering
\caption{\textbf{Seen object pose estimation on the T-LESS dataset.} Comparing our refinement to multi-view methods CenDerNet\citep{10161514}, DPODv2\citep{shugurov2021dpodv2} and CosyPose\citep{labbe2020cosypose}, we outperform baselines across all metrics, often by a large margin. For baselines, we rely on publicly available results where some metrics are not reported.}
\label{table:tless_seen}
\vspace*{-2mm}
\small
\begin{tabular}{l l c | c | c }
\toprule
\textbf{Data} & \textbf{Method} & \#views &  AR & AP \\
\midrule
\multirow{6}{*}{ \rotatebox{90}{synt}}
&CenDerNet & 5 & 71.3 & \textit{not available} \\
&DPODv2 & 1 & 63.6 & \textit{not available}\\
&\;+ DPODv2 MV & 4 & 68.9 & \textit{not available} \\
&CosyPose & 1 &64.0 & 63.0\\
&\;+ CosyPose MV & 4 & 72.8 & 70.6 \\
&\;+ Our refinement & 4 & \textbf{85.5} & \textbf{89.2} \\
\midrule
\multirow{5}{*}{ \rotatebox{90}{synt+real}}
&DPODv2 & 1 & 65.5 & \textit{not available} \\
&\;+  DPODv2 MV & 4 & 72.0 & \textit{not available}\\
&CosyPose  & 1 & 72.8 & 74.1 \\
&\;+ CosyPose MV & 4 & 81.5 & 81.7 \\
&\;+ Our refinement &  4 & \textbf{86.8} & \textbf{91.0}  \\
\bottomrule
\end{tabular}
\vspace{-2mm}
\end{table}

\subsection{Seen object pose estimation task}
Although our method is designed for the primary challenge of generalizable pose estimation for unseen objects, we also benchmark its performance in the seen object setting. This allows for a direct comparison against a wider range of established multi-view baselines that require object-specific training on the exact test objects. For this evaluation, we focus on the industry-relevant T-LESS dataset for which existing multi-view methods provide results publicly.

We show the results for seen object pose estimation in \cref{table:tless_seen}.
We compare with methods trained on synthetic data (synt) as well as those trained on a mix of synthetic and real data (synt+real), namely CenDerNet~\citep{10161514}, DPODv2~\citep{shugurov2021dpodv2}, and CosyPose~\citep{labbe2020cosypose}. DPODv2 and CosyPose both have a single-view version and a multi-view (MV) version that combines and refines these single-view inputs. Results for CosyPose MV were obtained by running their original pipeline with known camera extrinsics. As shown in \cref{table:tless_seen}, our method outperforms all baselines across all metrics. This indicates that despite requiring no training, our approach is competitive even against methods explicitly trained on the test objects. For fairness, we refine pose candidates generated by CosyPose’s single-view estimator, which was trained on T-LESS, ensuring the inputs to our refinement are comparable to those used by the baselines.

\begin{table}[tbp]
\renewcommand{\arraystretch}{0.85} 
\centering
\caption{\textbf{Ablation of the key components of our method on YCB-V and T-LESS datasets.} Performance of the single-view input to our method (1 view), followed by successive stages of our approach: multi-view aggregation (4 views aggreg.),  3D non-maximum suppression (4 views aggreg. + NMS), and multi-view refinement (4 views aggreg. + NMS + refine).}
\label{table:pipeline_stages}
\vspace*{-2mm}
\small
\begin{tabular}{lcc | cc}
\toprule
\multirow{2}{*}{\textbf{Method}} & \multicolumn{2}{c}{YCBV} & \multicolumn{2}{c}{TLESS} \\
 & AR & AP & AR & AP \\
\midrule
1 view (Co-op~\citep{moon2025co}) & 69.7 & 69.5 & 68.2 & 68.9 \\
4 views aggreg. & 63.7 & 33.9 & 69.5 & 48.6 \\
4 views aggreg. + refine & \textbf{84.5} & 38.0 & 77.7 & 39.9 \\
4 views aggreg. + NMS & 63.0 & 62.3 & 73.1 & 80.2 \\
4 views aggreg. + NMS + refine & 83.8 & \textbf{83.3} & \textbf{89.6} & \textbf{92.4} \\
\bottomrule
\end{tabular}
\vspace{-3mm}
\end{table}

\subsection{Ablations}
\paragraph{Method components.}
To understand and quantify the contribution of each component of our approach, we performed an ablation study. The results are shown in \cref{table:pipeline_stages}. Aggregating single-view candidates from multiple views substantially reduces AP due to redundant or inconsistent pose estimates. Applying NMS restores precision by keeping only the best candidates. Finally, the full pipeline, including refinement, demonstrates that visual-feature-based refinement is crucial for achieving high-quality estimates.

\vspace{-0.75em}
\paragraph{Feature descriptors.} We analyze the influence of the features used during our refinement. We investigate different models of DINOv2 and compare the performance across several hidden layers. We also include more recent ViT foundation models~\cite{heinrich2025radiov2,simeoni2025dinov3}. As shown in \cref{table:ablation_features}, most results vary by no more than a few AP/AR points between models and layers. For completeness, we also evaluate the performance of dense SIFT features~\cite{lowe1999object}. Although using SIFT leads to some refinement, performance remains more than $7\%$ below the performance of the foundation models. In general, we found our approach to be remarkably robust to the choice of foundation features that are semantically rich, rather than low-level gradient- or keypoint-based cues. For the experiments in the rest of the paper we use DINOv2-L layer 18 to enable direct comparison with~\cite{ornek2024foundpose}.

\vspace{-2mm}
\paragraph{Template retrieval.}
Our approach generates templates by rendering at runtime ("online") from each view and the initial pose estimate. In contrast, for single-view refinement, \cite{ornek2024foundpose} retrieves templates pre-generated offline. 
As shown in \cref{table:template_retrieval}, online template retrieval achieves higher accuracy than offline retrieval of the closest template. The online templates can be rendered with the same camera intrinsics and initial viewpoints as the query image, which we hypothesize improves the quality of extracted template features and explains the increase of accuracy. 
\vspace{-1mm}

\begin{table}[t]
\renewcommand{\arraystretch}{0.85} 
\centering
\caption{\textbf{Ablation of image feature descriptors.} 
Performance of our approach with different foundation features and dense SIFT features. The model used in the rest of the experiments is in italics.
}
\label{table:ablation_features}
\vspace*{-2mm}
\small
\begin{tabular}{lcc | cc}
\toprule
\multirow{2}{*}{\textbf{Feature descriptor}} & \multicolumn{2}{c}{YCBV} & \multicolumn{2}{c}{TLESS} \\
 & AR & AP & AR & AP \\
\midrule
DINOv2-S - layer 9  & 80.8 & 80.7 & 87.6 & 90.7 \\
DINOv2-S - layer 11 & 81.4 & 80.8 & 88.5 & 91.0 \\
DINOv2-B - layer 9  & 83.6 & 83.8 & 89.0 & 91.8 \\
DINOv2-B - layer 11 & 81.6 & 81.3 & 88.7 & 91.5 \\
\textit{DINOv2-L - layer 18} & 83.8 & 83.3 & 89.6 & 92.4 \\
DINOv2-L - layer 23 & 82.5 & 81.4 & 89.0 & 92.0 \\
\midrule
DINOv3-S - layer 9  & 84.3 & 84.0 & 88.2 & 91.0 \\
DINOv3-S - layer 11 & 81.1 & 80.8 & 88.1 & 90.1 \\
DINOv3-B - layer 9  & 83.7 & 83.6 & 88.6 & 91.7 \\
DINOv3-B - layer 11 & 82.7 & 82.2 & 88.7 & 91.3 \\
DINOv3-L - layer 18 & 84.5 & 84.1 & 89.0 & 92.0 \\
DINOv3-L - layer 23 & 84.1 & 83.9 & 89.6 & 91.6 \\
\midrule
RADIO2.5-B - layer 11 & 80.7 & 80.4 & 88.7 & 91.7 \\
RADIO2.5-H - layer 31 & 80.5 & 79.6 & 89.0 & 92.2 \\
\midrule
DenseSIFT & 73.0 & 70.7  & 80.1 & 84.9 \\
\bottomrule
\end{tabular}
\vspace{-1.5mm}
\end{table}

\begin{table}[t]
\renewcommand{\arraystretch}{0.85} 
\centering
\caption{\textbf{Ablation of the template retrieval.} We compare online (rendering) with offline method used in FoundPose~\cite{ornek2024foundpose}.
\label{table:template_retrieval}}
\vspace*{-2mm}
\small
\begin{tabular}{lcc|cc}
\toprule
\multirow{2}{*}{\textbf{Method}} & \multicolumn{2}{c}{YCB-V} & \multicolumn{2}{c}{T-LESS} \\
& AR & AP & AR & AP \\
\midrule
Offline \cite{ornek2024foundpose} & 82.3 & 81.7 & 86.7 & 89.2 \\
Online (Ours)  & \textbf{83.8} & \textbf{83.3} & \textbf{89.6} & \textbf{92.4} \\
\bottomrule
\end{tabular}
\vspace{-4mm}
\end{table}
\section{Conclusion}
In this work, we introduce AlignPose, a novel and effective method for multi-view generalizable 6D object pose estimation. The method efficiently aggregates information from multiple views and generates high-quality pose estimates. We show that a frozen vision foundation model can be leveraged for this task as a powerful feature extractor. This allows AlignPose to achieve remarkable generalization to new, unseen objects without any additional training. Our strategy sets a new state-of-the-art on key BOP benchmark and provides a powerful solution that is well-suited for industrial setups with challenging conditions, where multi-view information is most useful.

\subsection*{Acknowledgments}
This work was supported by the European Union’s Horizon Europe projects AGIMUS (No. 101070165), euROBIN (No. 101070596), ERC FRONTIER (No. 101097822), and ELLIOT (No. 101214398). It was further supported by the Grant Agency of the Czech Technical University in Prague (SGS25/152/OHK3/3T/13).
Compute resources and infrastructure were supported by the Ministry of Education, Youth and Sports of the Czech Republic through the e-INFRA CZ (ID:90254) and by the European Union’s Horizon Europe project CLARA (No. 101136607).
{
    \small
    \bibliographystyle{ieeenat_fullname}
    \bibliography{main}

@inproceedings{alexa2022super,
  title={{Super-Fibonacci Spirals: Fast, Low-Discrepancy Sampling of {SO}(3)}},
  author={Alexa, Marc},
  booktitle={CVPR},
  pages={8291--8300},
  year={2022}
}

@article{baker2004lucas,
  title={{Lucas-Kanade 20 Years On: A Unifying Framework}},
  author={Baker, Simon and Matthews, Iain},
  journal={International Journal of Computer Vision (IJCV)},
  volume={56},
  number={3},
  pages={221--255},
  year={2004},
}

@inproceedings{barron2019general,
  title={{A General and Adaptive Robust Loss Function}},
  author={Barron, Jonathan T},
  booktitle={CVPR},
  pages={4331--4339},
  year={2019}
}

@inproceedings{caraffa2024freeze,
  title={{Freeze: Training-Free Zero-Shot 6D Pose Estimation with Geometric and Vision Foundation Models}},
  author={Caraffa, Andrea and Boscaini, Davide and Hamza, Amir and Poiesi, Fabio},
  booktitle={ECCV},
  pages={414--431},
  year={2024},
}

@article{caraffa2025freezev2,
      title={{Accurate and Efficient Zero-Shot 6D Pose Estimation with Frozen Foundation Models}},
      author={Caraffa, Andrea and Boscaini, Davide and Poiesi, Fabio},
      journal={arXiv preprint arXiv:2506.09784},
      year={2025}, 
}

@inproceedings{chen2022clearpose,
  title={{ClearPose: Large-Scale Transparent Object Dataset and Benchmark}},
  author={Chen, Xiaotong and Zhang, Huijie and Yu, Zeren and Opipari, Anthony and Chadwicke Jenkins, Odest},
  booktitle={ECCV},
  pages={381--396},
  year={2022},
}

@inproceedings{itodd,
  author={Drost, Bertram and Ulrich, Markus and Bergmann, Paul and Härtinger, Philipp and Steger, Carsten},
  booktitle={ICCVW}, 
  title={{Introducing {MVTec} {ITODD} — A Dataset for 3D Object Recognition in Industry}}, 
  year={2017},
  pages={2200--2208},
  doi={10.1109/ICCVW.2017.257}
}

@inproceedings{ipd,
    author    = {Kalra, Agastya and Stoppi, Guy and Marin, Dmitrii and Taamazyan, Vage and Shandilya, Aarrushi and Agarwal, Rishav and Boykov, Anton and Chong, Tze Hao and Stark, Michael},
    title     = {{Towards Co-Evaluation of Cameras HDR and Algorithms for Industrial-Grade 6DoF Pose Estimation}},
    booktitle = {CVPR},
    year      = {2024},
    pages     = {22691--22701}
}

@article{xyzibd,
  title={{XYZ-IBD: A High-precision Bin-picking Dataset for Object 6D Pose Estimation Capturing Real-world Industrial Complexity}},
  author={Huang, Junwen and Liang, Jizhong and Hu, Jiaqi and Sundermeyer, Martin and Yu, Peter KT and Navab, Nassir and Busam, Benjamin},
  journal={arXiv preprint arXiv:2506.00599},
  year={2025}
}

@inproceedings{engel2014lsd,
  title={{LSD-SLAM: Large-Scale Direct Monocular SLAM}},
  author={Engel, Jakob and Sch{\"o}ps, Thomas and Cremers, Daniel},
  booktitle={ECCV},
  pages={834--849},
  year={2014},
}

@inproceedings{erkent2016integration,
  title={{Integration of Probabilistic Pose Estimates from Multiple Views}},
  author={Erkent, {\"O}zg{\"u}r and Shukla, Dadhichi and Piater, Justus},
  booktitle={ECCV},
  pages={154--170},
  year={2016},
}

@article{fischler1981random,
  title={{Random Sample Consensus: A Paradigm for Model Fitting with Applications to Image Analysis and Automated Cartography}},
  author={Fischler, Martin A and Bolles, Robert C},
  journal={Communications of the ACM},
  volume={24},
  year={1981},
  number={6},
  pages={381--395},
}

@inproceedings{3pt,
    title = {{3D-Object Perception Transformer (3PT)}},
    author = {Kalra, Agastya and Salzmann, Tim and Stoppi, Guy and Marin, Dmitrii and Agarwal, Rishav and Taamazyan, Vage and Bokeloh, Martin and Hinterstoisser, Stefan and Boykov, Anton and Dall'Olio, Alberto and Dangol, Pravin and Venkataraman, Kartik and Chen, Huaijin}, 
    booktitle = {CVPR},
    month = {June},
    year = {2026},
    note = {To appear.}
    }

@inproceedings{fu2021multi,
  title={{A Multi-Hypothesis Approach to Pose Ambiguity in Object-Based SLAM}},
  author={Fu, Jiahui and Huang, Qiangqiang and Doherty, Kevin and Wang, Yue and Leonard, John J},
  booktitle={IROS},
  pages={7639--7646},
  year={2021},
}

@inproceedings{10161514,
  author={Haugaard, Rasmus Laurvig and Iversen, Thorbjorn Mosekjaer},
  booktitle={ICRA}, 
  title={{Multi-View Object Pose Estimation from Correspondence Distributions and Epipolar Geometry}}, 
  year={2023},
  pages={1786--1792},
}

@inproceedings{heinrich2025radiov2,
  title={{RadioV2.5: Improved Baselines for Agglomerative Vision Foundation Models}},
  author={Heinrich, Greg and Ranzinger, Mike and Yin, Hongxu and Lu, Yao and Kautz, Jan and Tao, Andrew and Catanzaro, Bryan and Molchanov, Pavlo},
  booktitle={CVPR},
  pages={22487--22497},
  year={2025}
}

@inproceedings{hodavn2016evaluation,
  title={{On Evaluation of 6D Object Pose Estimation}},
  author={Hoda{\v{n}}, Tom{\'a}{\v{s}} and Matas, Ji{\v{r}}{\'\i} and Obdr{\v{z}}{\'a}lek, {\v{S}}t{\v{e}}p{\'a}n},
  booktitle={ECCV},
  pages={606--619},
  year={2016},
}

@article{hodan2017tless,
  title={{T-LESS: An RGB-D Dataset for 6D Pose Estimation of Texture-Less Objects}},
  author={Hoda{\v{n}}, Tom{\'a}{\v{s}} and Haluza, Pavel and Obdr{\v{z}}{\'a}lek, {\v{S}}t{\v{e}}p{\'a}n and Matas, Ji{\v{r}}{\'\i} and Lourakis, Manolis and Zabulis, Xenophon},
  journal={WACV},
  year={2017}
}

@inproceedings{hodan2018bop,
  title={{BOP: Benchmark for 6D Object Pose Estimation}},
  author={Hodan, Tomas and Michel, Frank and Brachmann, Eric and Kehl, Wadim and GlentBuch, Anders and Kraft, Dirk and Drost, Bertram and Vidal, Joel and Ihrke, Stephan and Zabulis, Xenophon and others},
  booktitle={ECCV},
  pages={19--34},
  year={2018}
}

@inproceedings{hodavn2020bop,
  title={{BOP Challenge 2020 on 6D Object Localization}},
  author={Hoda{\v{n}}, Tom{\'a}{\v{s}} and Sundermeyer, Martin and Drost, Bertram and Labb{\'e}, Yann and Brachmann, Eric and Michel, Frank and Rother, Carsten and Matas, Ji{\v{r}}{\'\i}},
  booktitle={ECCVW},
  pages={577--594},
  year={2020},
}

@inproceedings{bop23,
  author={Hodan, Tomas and Sundermeyer, Martin and Labbe, Yann and Nguyen, Van Nguyen and Wang, Gu and Brachmann, Eric and Drost, Bertram and Lepetit, Vincent and Rother, Carsten and Matas, Jiri},
  title={{BOP Challenge 2023 on Detection, Segmentation and Pose Estimation of Seen and Unseen Rigid Objects}},
  booktitle={CVPRW},
  month={6},
  year={2024},
  pages={5610--5619}
}

@inproceedings{irani1999direct,
  title={{About Direct Methods}},
  author={Irani, Michal and Anandan, Prabu},
  booktitle={International Workshop on Vision Algorithms},
  pages={267--277},
  year={1999},
}

@inproceedings{iwase2021repose,
  author={Iwase, Shun and Liu, Xingyu and Khirodkar, Rawal and Yokota, Rio and Kitani, Kris M.},
  title={{RePOSE: Fast 6D Object Pose Refinement via Deep Texture Rendering}},
  booktitle={CVPR},
  month={October},
  year={2021},
  pages={3303--3312}
}

@inproceedings{jung2024housecat,
  author={Jung, HyunJun and Wu, Shun-Cheng and Ruhkamp, Patrick and Zhai, Guangyao and Schieber, Hannah and Rizzoli, Giulia and Wang, Pengyuan and Zhao, Hongcheng and Garattoni, Lorenzo and Meier, Sven and Roth, Daniel and Navab, Nassir and Busam, Benjamin},
  title={{HouseCat6D - A Large-Scale Multi-Modal Category Level 6D Object Perception Dataset with Household Objects in Realistic Scenarios}},
  booktitle={CVPR},
  month={June},
  year={2024},
  pages={22498--22508}
}

@inproceedings{kaskman20206,
  title={{6 DOF Pose Estimation of Textureless Objects from Multiple RGB Frames}},
  author={Kaskman, Roman and Shugurov, Ivan and Zakharov, Sergey and Ilic, Slobodan},
  booktitle={ECCVW},
  pages={612--630},
  year={2020},
}

@inproceedings{labbe2020cosypose,
  title={{CosyPose: Consistent Multi-View Multi-Object 6D Pose Estimation}},
  author={Labb{\'e}, Yann and Carpentier, Justin and Aubry, Mathieu and Sivic, Josef},
  booktitle={ECCV},
  pages={574--591},
  year={2020},
}

@inproceedings{labbe2022megapose,
  title={{MegaPose: 6D Pose Estimation of Novel Objects via Render \& Compare}},
  author={Labb\'e, Yann and Manuelli, Lucas and Mousavian, Arsalan and Tyree, Stephen and Birchfield, Stan and Tremblay, Jonathan and Carpentier, Justin and Aubry, Mathieu and Fox, Dieter and Sivic, Josef},
  booktitle={CoRL},
  year={2022}
}

@article{levenberg1944method,
  title={{A Method for the Solution of Certain Non-Linear Problems in Least Squares}},
  author={Levenberg, Kenneth},
  journal={Quarterly of Applied Mathematics},
  volume={2},
  number={2},
  pages={164--168},
  year={1944}
}

@inproceedings{li2018unified,
  title={{A Unified Framework for Multi-View Multi-Class Object Pose Estimation}},
  author={Li, Chi and Bai, Jin and Hager, Gregory D},
  booktitle={ECCV},
  pages={254--269},
  year={2018}
}

@inproceedings{li2018deepim,
  title={{DeepIM: Deep Iterative Matching for 6D Pose Estimation}},
  author={Li, Yi and Wang, Gu and Ji, Xiangyang and Xiang, Yu and Fox, Dieter},
  booktitle={ECCV},
  pages={683--698},
  year={2018}
}

@inproceedings{lin2014microsoft,
  title={{Microsoft COCO: Common Objects in Context}},
  author={Lin, Tsung-Yi and Maire, Michael and Belongie, Serge and Hays, James and Perona, Pietro and Ramanan, Deva and Doll{\'a}r, Piotr and Zitnick, C Lawrence},
  booktitle={ECCV},
  pages={740--755},
  year={2014}
}

@inproceedings{lindenberger2021pixelperfect,
  author={Lindenberger, Philipp and Sarlin, Paul-Edouard and Larsson, Viktor and Pollefeys, Marc},
  title={{Pixel-Perfect Structure-From-Motion with Featuremetric Refinement}},
  booktitle={ICCV},
  month={October},
  year={2021},
  pages={5987--5997}
}

@inproceedings{liu2024grounding,
  title={{Grounding DINO: Marrying DINO with Grounded Pre-Training for Open-Set Object Detection}},
  author={Liu, Shilong and Zeng, Zhaoyang and Ren, Tianhe and Li, Feng and Zhang, Hao and Yang, Jie and Jiang, Qing and Li, Chunyuan and Yang, Jianwei and Su, Hang and others},
  booktitle={ECCV},
  pages={38--55},
  year={2024}
}

@article{liu2025gdrnpp,
  title={{GDRNPP: A Geometry-Guided and Fully Learning-Based Object Pose Estimator}},
  author={Liu, Xingyu and Zhang, Ruida and Zhang, Chenyangguang and Wang, Gu and Tang, Jiwen and Li, Zhigang and Ji, Xiangyang},
  journal={IEEE Transactions on Pattern Analysis and Machine Intelligence (TPAMI)},
  year={2025}
}

@inproceedings{lowe1999object,
  title={{Object Recognition from Local Scale-Invariant Features}},
  author={Lowe, David G},
  booktitle={ICCV},
  volume={2},
  pages={1150--1157},
  year={1999},
}

@inproceedings{lu2024adapting,
  title={{Adapting Pre-Trained Vision Models for Novel Instance Detection and Segmentation}},
  author={Lu, Yangxiao and Guo, Yunhui and Ruozzi, Nicholas and Xiang, Yu and others},
  booktitle={IROS},
  pages={13341--13348},
  year={2025},
}

@article{marquardt1963algorithm,
  title={{An Algorithm for Least-Squares Estimation of Nonlinear Parameters}},
  author={Marquardt, Donald W},
  journal={Journal of the Society for Industrial and Applied Mathematics},
  volume={11},
  number={2},
  pages={431--441},
  year={1963}
}

@inproceedings{moon2025co,
  title={{Co-Op: Correspondence-Based Novel Object Pose Estimation}},
  author={Moon, Sungphill and Son, Hyeontae and Hur, Dongcheol and Kim, Sangwook},
  booktitle={CVPR},
  pages={11622--11632},
  year={2025}
}

@inproceedings{nguyen2023cnos,
  title={{CNOS: A Strong Baseline for CAD-Based Novel Object Segmentation}},
  author={Nguyen, Van Nguyen and Groueix, Thibault and Ponimatkin, Georgy and Lepetit, Vincent and Hodan, Tomas},
  booktitle={ICCV},
  pages={2134--2140},
  year={2023}
}

@inproceedings{nguyen2024gigapose,
  title={{GigaPose: Fast and Robust Novel Object Pose Estimation via One Correspondence}},
  author={Nguyen, Van Nguyen and Groueix, Thibault and Salzmann, Mathieu and Lepetit, Vincent},
  booktitle={CVPR},
  pages={9903--9913},
  year={2024}
}

@article{oquab2023dinov2,
  title={{DINOv2: Learning Robust Visual Features without Supervision}},
  author={Oquab, Maxime and Darcet, Timoth{\'e}e and Moutakanni, Th{\'e}o and Vo, Huy and Szafraniec, Marc and Khalidov, Vasil and Fernandez, Pierre and Haziza, Daniel and Massa, Francisco and El-Nouby, Alaaeldin and others},
  journal={Transactions on Machine Learning Research Journal},
  pages={1--31},
  year={2024}
}

@inproceedings{ornek2024foundpose,
  title={{FoundPose: Unseen Object Pose Estimation with Foundation Features}},
  author={{\"O}rnek, Evin P{\i}nar and Labb{\'e}, Yann and Tekin, Bugra and Ma, Lingni and Keskin, Cem and Forster, Christian and Hodan, Tomas},
  booktitle={ECCV},
  pages={163--182},
  year={2024}
}

@article{poiesi2022learning,
  title={{Learning General and Distinctive 3D Local Deep Descriptors for Point Cloud Registration}},
  author={Poiesi, Fabio and Boscaini, Davide},
  journal={IEEE Transactions on Pattern Analysis and Machine Intelligence (TPAMI)},
  volume={45},
  number={3},
  pages={3979--3985},
  year={2022}
}

@inproceedings{ravi2024sam,
  title={{SAM 2: Segment Anything in Images and Videos}},
  author={Ravi, Nikhila and Gabeur, Valentin and Hu, Yuan-Ting and Hu, Ronghang and Ryali, Chaitanya and Ma, Tengyu and Khedr, Haitham and R{\"a}dle, Roman and Rolland, Chloe and Gustafson, Laura and others},
  booktitle={ICLR},
  pages={28085--28128},
  year={2025}
}

@inproceedings{salas2013slam,
  title={{SLAM++: Simultaneous Localisation and Mapping at the Level of Objects}},
  author={Salas-Moreno, Renato F and Newcombe, Richard A and Strasdat, Hauke and Kelly, Paul HJ and Davison, Andrew J},
  booktitle={CVPR},
  pages={1352--1359},
  year={2013}
}

@inproceedings{sarlin2021back,
  title={{Back to the Feature: Learning Robust Camera Localization from Pixels to Pose}},
  author={Sarlin, Paul-Edouard and Unagar, Ajaykumar and Larsson, Mans and Germain, Hugo and Toft, Carl and Larsson, Viktor and Pollefeys, Marc and Lepetit, Vincent and Hammarstrand, Lars and Kahl, Fredrik and others},
  booktitle={CVPR},
  pages={3247--3257},
  year={2021}
}

@inproceedings{shu2020feature,
  title={{Feature-Metric Loss for Self-Supervised Learning of Depth and EgoMotion}},
  author={Shu, Chang and Yu, Kun and Duan, Zhixiang and Yang, Kuiyuan},
  booktitle={ECCV},
  pages={572--588},
  year={2020}
}

@article{shugurov2021multi,
  title={{Multi-View Object Pose Refinement with Differentiable Renderer}},
  author={Shugurov, Ivan and Pavlov, Ivan and Zakharov, Sergey and Ilic, Slobodan},
  journal={IEEE Robotics and Automation Letters},
  volume={6},
  number={2},
  pages={2579--2586},
  year={2021}
}

@article{shugurov2021dpodv2,
  title={{DPODv2: Dense Correspondence-Based 6 DOF Pose Estimation}},
  author={Shugurov, Ivan and Zakharov, Sergey and Ilic, Slobodan},
  journal={IEEE Transactions on Pattern Analysis and Machine Intelligence (TPAMI)},
  volume={44},
  number={11},
  pages={7417--7435},
  year={2021}
}

@article{simeoni2025dinov3,
  title={{DINOv3}},
  author={Sim{\'e}oni, Oriane and Vo, Huy V and Seitzer, Maximilian and Baldassarre, Federico and Oquab, Maxime and Jose, Cijo and Khalidov, Vasil and Szafraniec, Marc and Yi, Seungeun and Ramamonjisoa, Micha{\"e}l and others},
  journal={arXiv preprint arXiv:2508.10104},
  year={2025}
}

@inproceedings{sock2017multi,
  title={{Multi-View 6D Object Pose Estimation and Camera Motion Planning Using RGBD Images}},
  author={Sock, Juil and Hamidreza Kasaei, S and Seabra Lopes, Luis and Kim, Tae-Kyun},
  booktitle={ICCVW},
  pages={2228--2235},
  year={2017}
}

@inproceedings{taher2024fit,
  title={{Fit-NGP: Fitting Object Models to Neural Graphics Primitives}},
  author={Taher, Marwan and Alzugaray, Ignacio and Davison, Andrew J},
  booktitle={ICRA},
  pages={18186--18192},
  year={2024}
}

@inproceedings{trivigno2024unreasonable,
  title={{The Unreasonable Effectiveness of Pre-Trained Features for Camera Pose Refinement}},
  author={Trivigno, Gabriele and Masone, Carlo and Caputo, Barbara and Sattler, Torsten},
  booktitle={CVPR},
  pages={12786--12798},
  year={2024}
}

@article{van2025bop,
  title={{BOP Challenge 2024 on Model-Based and Model-Free 6D Object Pose Estimation}},
  author={Van Nguyen, Nguyen and Tyree, Stephen and Guo, Andrew and Fourmy, Mederic and Gouda, Anas and Lee, Taeyeop and Moon, Sungphill and Son, Hyeontae and Ranftl, Lukas and Tremblay, Jonathan and Brachmann, Eric and others},
  journal={CoRR},
  year={2025}
}

@inproceedings{von2020lm,
  title={{LM-Reloc: Levenberg-Marquardt Based Direct Visual Relocalization}},
  author={von Stumberg, Lukas and Wenzel, Patrick and Yang, Nan and Cremers, Daniel},
  booktitle={2020 International Conference on 3D Vision (3DV)},
  pages={968--977},
  year={2020},
}

@inproceedings{wada2020morefusion,
  title={{MoreFusion: Multi-Object Reasoning for 6D Pose Estimation from Volumetric Fusion}},
  author={Wada, Kentaro and Sucar, Edgar and James, Stephen and Lenton, Daniel and Davison, Andrew J},
  booktitle={CVPR},
  pages={14540--14549},
  year={2020}
}

@inproceedings{wang2019normalized,
  title={{Normalized Object Coordinate Space for Category-Level 6D Object Pose and Size Estimation}},
  author={Wang, He and Sridhar, Srinath and Huang, Jingwei and Valentin, Julien and Song, Shuran and Guibas, Leonidas J},
  booktitle={CVPR},
  pages={2642--2651},
  year={2019}
}

@inproceedings{wen2024foundationpose,
  title={{FoundationPose: Unified 6D Pose Estimation and Tracking of Novel Objects}},
  author={Wen, Bowen and Yang, Wei and Kautz, Jan and Birchfield, Stan},
  booktitle={CVPR},
  pages={17868--17879},
  year={2024}
}

@inproceedings{xiang2018posecnn,
  title={{PoseCNN: A Convolutional Neural Network for 6D Object Pose Estimation in Cluttered Scenes}},
  author={Xiang, Yu and Schmidt, Tanner and Narayanan, Venkatraman and Fox, Dieter},
  booktitle={RSS},
  year={2018},
}

@inproceedings{yang20236d,
  title={{6D Pose Estimation for Textureless Objects on RGB Frames Using Multi-View Optimization}},
  author={Yang, Jun and Xue, Wenjie and Ghavidel, Sahar and Waslander, Steven L},
  booktitle={ICRA},
  pages={2905--2912},
  year={2023},
}

@article{zorina2024temporally,
  title={{Temporally Consistent Object 6D Pose Estimation for Robot Control}},
  author={Zorina, Kateryna and Priban, Vojtech and Fourmy, Mederic and Sivic, Josef and Petrik, Vladimir},
  journal={IEEE Robotics and Automation Letters},
  year={2024},
}

@inproceedings{riba2020kornia,
  title={{Kornia: An Open Source Differentiable Computer Vision Library for PyTorch}},
  author={Riba, Edgar and Mishkin, Dmytro and Ponsa, Daniel and Rublee, Ethan and Bradski, Gary},
  booktitle={WACV},
  pages={3674--3683},
  year={2020}
}
}

\clearpage

\appendix
\section*{\large{Appendix}}
\label{sec:supplementary}
This appendix contains additional implementation details for our method (\cref{sec:appendix_implementation_details}) and supplementary experimental results supporting our design choices (\cref{sec:appendix_additional_experiments}). Additional qualitative results on the HouseCat6D and ITODD-MV datasets are presented in \cref{sec:appendix_qualitative_results}. Failure cases are summarized in \cref{sec:appendix_limitations}.

\section{Implementation Details}\label{sec:appendix_implementation_details}
This section is organized according to the stages of our pipeline. We first describe how single-view object detections and pose candidates are obtained in \cref{sec:appendix_single-view}. Details on aggregating these candidates are given in \cref{sec:appendix_aggregation}. The multi-view refinement procedure is split into Sections \ref{sec:appendix_preprocessing}, \ref{sec:appendix_feature_extraction} and \ref{sec:appendix_optimization}, that cover the preprocessing, feature extraction, and optimization steps. Finally, we report the hardware setup and timing measurements in \cref{sec:appendix_hardware}.

\subsection{Single-view pose estimation}\label{sec:appendix_single-view}

\subsubsection{T-LESS, YCB-V}
To evaluate our method on T-LESS and YCB-V dataset, we use single-view pose candidates downloaded from BOP Leaderboard\footnote{\url{https://bop.felk.cvut.cz/leaderboards/}}, specifically we use the following versions:
FoundPose+FeatRef+Megapose-5hyp, GigaPose+GenFlow (5 hypotheses), 
MegaPose-CNOS\_fastSAM+MultiHyp, Co-op (F3DT2D, 5 Hypo). For all ablations on these datasets, we use downloaded single-view pose candidates by Co-op (F3DT2D, 5 Hypo).

\vspace{1em}
\subsubsection{HouseCat6D, BOP-Industrial}
For HouseCat6D and BOP-Industrial datasets (ITODD-MV, IPD, XYZ-IBD), we do not have access either to 2D detections or single-view 6D pose estimates\footnote{BOP's default detections and poses are insufficient as they cover only a single view of a scene. For HouseCat6D, we do not have access to model-based RGB pose estimates using a realistic RGB 2D detector.}. Following our goal of building a pose estimation method generalizable to novel objects, we designed a single-view 6D object detection pipeline based on existing generalizable and open-source methods consisting of 2D detection, segmentation and single view pose estimation.

\paragraph{2D detection and segmentation.} 
Inspired by~\citep{lu2024adapting}, we implemented a three step training-free 2D detection method, called FOSOD (Foundation model based Open-Set Object Detection). First, we detect many object candidates using the open-set 2D detector GroundingDINO~\citep{liu2024grounding}, using a fairly low threshold of 0.1. We then obtain object masks using SAM2~\citep{ravi2024sam} using detection bounding boxes as box prompts. Finally, we use template matching to match detections to a set of available textured models. In an offline procedure (less than 5 minutes per object), 128 object templates are rendered offline with viewpoints obtained using Fibonnacci sampling of SO(3)~\citep{alexa2022super}. At inference time, 2D detections are used to crop parts of the query image, and a crop descriptor is used to detect an object label following the CNOS methodology~\citep{nguyen2023cnos},~\citep{liu2024grounding}. We found that using cosine similarity of DINOv3-B layer 11 CLS tokens as descriptors gave reasonable matching results. 

For BOP-Industrial datasets, we use 2D detections by 3PT~\citep{3pt}, a transformer-based model trained for CAD-prompted 2D detection, and the winner of the BOP Challenge 2025. Based on these detections as box prompts, we generate segmentations with SAM2~\citep{ravi2024sam}.

We evaluate the performance of detection methods and analyze how detection quality impacts 6D pose estimation in \cref{sec:ablation_detection_quality}.

\paragraph{Single-view 6D pose estimates.} We adopted a coarse-to-fine 6D pose estimation approach. We first compute a coarse pose estimate using FoundPose~\citep{ornek2024foundpose}, then use FoundPose feature-metric refinement, and finally refine the results using the MegaPose~\citep{labbe2022megapose} refiner with 5 iterations. For textureless objects, we adjusted the MegaPose rendering by adding directional lights to accentuate the 3D shape through shading, which would otherwise appear as a featureless gray silhouette. We use this method for all experiments on HouseCat6D and BOP-Industrial datasets.

\subsection{Aggregation}\label{sec:appendix_aggregation}
Our 3D Non-Maximum Suppression (NMS) is implemented using world-axis-aligned 3D bounding boxes of object pose candidates. We use 3D IoU as an overlap measure with a threshold of 0.4. We apply NMS globally across all object candidates in the scene.
We apply NMS at two stages in our pipeline: once during the aggregation stage to filter out redundant initial proposals, and again after the refinement stage to handle cases where object poses may have converged or become significantly overlapping during the refinement. We evaluate different NMS settings in \cref{sec:ablation_nms}. 

\subsection{Preprocessing}\label{sec:appendix_preprocessing}
As a preparation for multi-view alignment, we prepare two distinct, fixed representations for each view: a 2D query feature map and a set of 3D registered features. In this section, we describe the generation process inspired by FoundPose~\citep{ornek2024foundpose}. We now describe how the query features and registered features are created for a view associated with camera $C$ and for an object $O$ with a coarse pose estimate $\bm{T}_{WO}$.

\paragraph{2D query features.} Given a 2D bounding box obtained from the coarse pose $\bm{T}_{WO}$, we crop the image using the perspective cropping method implemented by \citep{ornek2024foundpose}. For each view, this process yields a cropped image of size $420\times420$ and a corresponding crop camera $C'$ with pose $\bm{T}_{WC'}$ in the world coordinates. We then extract a 2D feature map of this image. The cropped image is decomposed into a grid of non-overlapping $14 \times 14$ patches, and each of these patches is embedded with a feature descriptor. We use the hidden state of layer 18 of the DINOv2 backbone, which has been empirically found to provide a good balance between positional and semantic information~\citep{ornek2024foundpose}. The resulting feature map is then up-sampled to the crop resolution via bilinear interpolation.

\paragraph{3D registered features.} To generate the set of registered features, we render an RGB-D image of the object as it would appear from the crop camera $C'$ given the coarse pose $\bm{T}_{WO}$. After rendering, we extract a feature descriptor for each patch, analogously to the process used for the query features, but we keep only the descriptors of patches centered on the object rather than the background. Additionally, we lift the 2D descriptors into 3D object space. Each descriptor $\mathbf{p}_i$ corresponding to a patch centered at pixel $c_i$ is assigned the 3D point $\mathbf{x}_i$ in object coordinates that projects to $c_i$. This yields a set of {\em registered features} $\mathcal{F}_{C'O} = \{\mathbf{p}_i,\mathbf{x}_i\}$ where $\mathbf{p}_i$ is a patch descriptor and $\mathbf{x}_i$ its corresponding 3D location in object space.

\subsection{Feature extraction.}\label{sec:appendix_feature_extraction}
For all feature extraction (except for the feature descriptor ablation study), we use the following configuration: We produce cropped images and rendered images of size $420\times420$~px and use the DINOv2 ViT-L model with registers to extract $14\times14$ patch features as our image descriptors. We use layer 18 of the model and apply layer normalization on the output. To save memory and compute, we project the patch features into the top 256 PCA components.

For the feature descriptor ablation study, we use resolutions of $480\times480$~px for DINOv3 and $432\times432$~px for RADIO2.5, which are the recommended resolutions and otherwise follow the same process. Both of these models use $16\times16$ patch size. For the extraction of dense SIFT features, we use the implementation by Kornia \citep{riba2020kornia}, with resolution $420\times420$~px and $14\times14$ spatial bins.
 
\paragraph{Dimensionality reduction.}  Feature extractors such as DINOv2/v3 produce high-dimensional descriptors (e.g., 1024 dimensions). To make the optimization process tractable, we reduce the dimensionality of these descriptors using Principal Component Analysis (PCA). The principal components are pre-computed for each object type during an offline onboarding stage and then applied to all subsequent patch descriptors of that object. 

Specifically, the components are estimated from feature descriptors extracted from renders of the object observed under a diverse set of viewpoints. We render 788 templates per object and extract their feature descriptors, which takes less than 5 minutes per object. The PCA is computed only on the masked part of the renders that contain the object. Note that, contrary to~\citep{ornek2024foundpose}, these templates are only used to compute the PCA projection matrix and are not kept in memory at inference time.

\subsection{Optimization}\label{sec:appendix_optimization}
Our nonlinear solver is implemented as a Levenberg-Marquardt adapted from~\citep{sarlin2021back}. We use the Barron robust loss function with parameters $\alpha=-5$, $c=0.5$. Using these parameters does not penalize large residuals as aggressively as an $\ell_2$ loss, which helps to stabilize the refinement process. A study of the effects of these parameters is presented in \cref{sec:ablation_cost}.

\subsection{Hardware and timings}\label{sec:appendix_hardware}
We ran all of our experiments on nodes equipped with NVIDIA-L40S GPUs and Intel(R) Xeon(R) 6760P CPUs. Our method takes less than a second per detection to refine aggregated poses. Individual dataset timings are reported in~\cref{table:timings}, where we measure time for pose optimization per object candidate in YCB-B and T-LESS datasets. The time scales linearly with the number of object candidates.
For reference, CosyPose processes all candidates in a scene jointly (requires at least three objects) and takes on average 0.400$\pm$0.695\,s \emph{per scene} on YCB-V and 0.277$\pm$0.664\,s on T-LESS when refining single-view candidates predicted by Co-op.

\begin{table}[t]
\centering
\caption[]{\textbf{Our refinement timings.} In this table we report per-object runtime of our multi-view refinement (avg$\pm$std).}
\label{table:timings}
\vspace*{-2mm}
\small
\begin{tabular}{lc|c}
\toprule
\textbf{Method} & YCB-V & T-LESS \\
\midrule
Refinement time [s] & 0.560$\pm$0.064 & 0.556$\pm$0.067 \\
\bottomrule
\end{tabular}
\vspace{-5mm}
\end{table}

\section{Additional experimental results}\label{sec:appendix_additional_experiments}

\begin{table*}[t]
\caption[]{\textbf{Multi-view pose estimation of unseen objects on YCB-V and T-LESS.} This table is an extension of Tab.~\ref{table:unseen_results_bop_classic} in the main paper. Both our method and CosyPose~\citep{labbe2020cosypose} refine candidates from FoundPose~\citep{ornek2024foundpose}, GigaPose~\citep{nguyen2024gigapose}, MegaPose~\citep{labbe2022megapose}, and Co-Op~\citep{moon2025co}. Our approach achieves higher performance across datasets.}
\small
\label{table:unseen_results_full}
\vspace{-4mm}
\begin{center}
\begin{tabular}{l l c c c c | c c c}
\toprule
\textbf{Dataset}&
\textbf{Method} & 
AR & AR\textsubscript{VSD} & AR\textsubscript{MSSD} & AR\textsubscript{MSPD} & AP & AP\textsubscript{MSSD} & AP\textsubscript{MSPD} \\ \noalign{\vskip 0.2em}
\toprule
\multirow{12}{*}{YCB-V}
& FoundPose & 69.0 & 60.2 & 67.0 & 79.7 & 63.0 & 54.5 & 71.4 \\
&\quad + CosyPose MV & 79.2 & 73.6 & 81.4 & 82.7 & 76.1 & 74.3 & 77.8 \\
&\quad + Ours & \textbf{83.9} & \textbf{78.7} & \textbf{88.8} & \textbf{84.2} & \textbf{83.2} & \textbf{85.0} & \textbf{81.4}\\
\noalign{\vskip 0.2em} \cline{2-9}\noalign{\vskip 0.2em}
&GigaPose & 66.6 & 57.4 & 64.2 & 78.2 & 63.1 & 54.2 & 72.0 \\
&\quad + CosyPose MV & 76.5 & 70.3 & 77.8 & 81.2 & 70.9 & 68.3 & 73.4 \\
&\quad + Ours  & \textbf{81.9} & \textbf{77.0} & \textbf{86.7} & \textbf{82.0} & \textbf{79.7} & \textbf{81.3} & \textbf{78.1} \\
\noalign{\vskip 0.2em} \cline{2-9}\noalign{\vskip 0.2em}
&MegaPose & 62.0 & 53.5 & 59.7 & 72.8 & 56.1 & 47.8 & 64.5 \\
&\quad + CosyPose MV & 71.1 & 65.1 & 72.3 & 75.8 & 64.5 & 61.8 & 67.2 \\
&\quad + Ours  & \textbf{80.0} & \textbf{75.0} & \textbf{84.6} & \textbf{80.3} & \textbf{77.3} & \textbf{78.7} & \textbf{75.8}\\
\noalign{\vskip 0.2em} \cline{2-9}\noalign{\vskip 0.2em}
&Co-op & 69.7 & 58.3 & 66.3 & 84.6 & 69.5 & 57.8 & 81.2 \\
&\quad + CosyPose MV & 81.0 & 73.9 & 83.0 & \textbf{86.1} & 79.2 & 76.3 & \textbf{82.1} \\
&\quad + Ours  & \textbf{83.8} & \textbf{78.6} & \textbf{88.5} & 84.2 & \textbf{83.3} & \textbf{84.8} & 81.7\\
\midrule

\multirow{12}{*}{T-LESS} 
&FoundPose & 57.0 & 53.6 & 54.9 & 62.3 & 57.0 & 52.9 & 61.1 \\
&\quad + CosyPose MV & 66.4 & 62.4 & 63.9 & 67.0 & 63.0 & 62.1 & 64.0 \\
&\quad + Ours  & \textbf{84.1} & \textbf{80.9} & \textbf{85.3} & \textbf{85.9} & \textbf{88.6} & \textbf{88.5} & \textbf{88.6} \\
\noalign{\vskip 0.2em} \cline{2-9}\noalign{\vskip 0.2em}
&GigaPose & 58.2 & 54.9 & 56.0 & 63.7 & 54.3 & 50.8 & 57.8 \\
&\quad + CosyPose MV & 61.9 & 59.5 & 60.6 & 65.6 & 55.9 & 55.5 & 57.3 \\
&\quad + Ours  & \textbf{81.9} & \textbf{79.3} & \textbf{83.0} & \textbf{83.6} & \textbf{84.9} & \textbf{84.8} & \textbf{85.0} \\
\noalign{\vskip 0.2em} \cline{2-9}\noalign{\vskip 0.2em}
&Megapose & 50.8 & 48.1 & 48.5 & 55.9 & 50.5 & 46.6 & 54.4 \\
&\quad + CosyPose MV & 57.6 & 55.8 & 56.7 & 60.3 & 56.1 & 54.9 & 57.3 \\
&\quad + Ours  & \textbf{78.8} & \textbf{76.1} & \textbf{79.7} & \textbf{80.5} & \textbf{81.4} & \textbf{81.3} & \textbf{81.6} \\
\noalign{\vskip 0.2em} \cline{2-9}\noalign{\vskip 0.2em}
&Co-op & 68.2 & 64.0 & 65.8 & 74.8 & 68.9 & 64.3 & 73.4 \\
&\quad + CosyPose MV& 78.7 & 76.9 & 78.5 & 80.7 & 78.9 & 78.3 & 79.4 \\
&\quad + Ours & \textbf{89.6} & \textbf{86.3} & \textbf{90.9} & \textbf{91.5} & \textbf{92.4} & \textbf{92.3} & \textbf{92.5} \\
\bottomrule
\end{tabular}
\end{center}
\vspace{-3mm}
\end{table*}

\begin{table*}[t]
\caption[]{\textbf{HouseCat6D results for the full dataset and metallic and glass categories.} Extended version of Tab.~\ref{table:unseen_results_housecat} in the main paper. Both our method and CosyPose~\citep{labbe2020cosypose} refine candidates from FoundPose~\citep{ornek2024foundpose}. We demonstrate significant improvements of the pose estimation scores on the groups of glass and metallic objects compared to the single-view and multi-view baselines.}
\small
\label{table:housecat6d_metal_glass_supp}
\vspace{-4mm}
\begin{center}
\begin{tabular}{l l c c c c | c c c }
\toprule
\textbf{Subset} & \textbf{Method} & AR & AR\textsubscript{VSD} & AR\textsubscript{MSSD} & AR\textsubscript{MSPD} & AP & AP\textsubscript{MSSD} & AP\textsubscript{MSPD}\\
\midrule
\multirow{3}{*}{Full dataset} 
&FoundPose & 80.3 & 79.1 & 79.0 & 82.7 & 72.2 & 70.1 & 74.2\\
&\quad + CosyPose MV & 86.7 & 84.5 & 87.0 & 88.6 & 86.6 & 85.9 & 87.3\\
&\quad + Ours & \textbf{88.9} & \textbf{85.3} & \textbf{90.4} & \textbf{91.2} &\textbf{89.6} & \textbf{89.4} & \textbf{89.8}\\
\midrule
\multirow{3}{*}{Glass objects}
&FoundPose & 75.1 & 66.0 & 77.6 & 81.8 & 71.4 & 69.2 & 73.7 \\
&\quad + CosyPoseMV & 84.6 & 75.8 & 88.8 & 89.2 & 86.8 & 86.6 & 87.0\\
&\quad + Ours & \textbf{88.9} & \textbf{79.2} & \textbf{92.7} & \textbf{95.0} & \textbf{92.8} & \textbf{91.9} & \textbf{93.8}\\
\midrule
\multirow{3}{*}{Metallic objects}
&FoundPose & 19.3 & 11.2 & 15.3 & 31.5 & 19.0 & 10.3 & 27.7 \\
&\quad + CosyPose MV & 25.8 & 15.9 & 23.7 & 37.7& 27.1 & 21.0 & 33.2 \\
&\quad + Ours  & \textbf{43.5} & \textbf{27.0} & \textbf{49.9} & \textbf{53.5}& \textbf{46.5} & \textbf{45.2} & \textbf{47.8} \\
\bottomrule
\end{tabular}
\end{center}
\vspace{-3mm}
\end{table*}

\subsection{Complete set of results with all metrics}\label{sec:appendix_detailed_results}
In \cref{table:unseen_results_full}, \ref{table:housecat6d_metal_glass_supp} and \ref{tab:bop_industrial_full}, we present additional results of unseen pose estimation, detailing the AR/AP scores of each VSD/MSSD/MSPD metric. Our method is able to refine all single-view pose estimates with a higher accuracy than our baseline CosyPose, as shown by the consistently higher AP/AR scores.
An exception to the overall trend appears in \cref{table:unseen_results_full} for the MSPD error when refining YCB-V candidates from Co-Op~\citep{moon2025co}. In this case, our method does not yield an improvement in this single reprojection error metric. A~likely explanation is that our refinement adjusts the object pose to better align in the spatial/depth direction, thereby reducing MSSD errors (increasing AR\textsubscript{MSSD} and AP\textsubscript{MSSD}). However, this stricter alignment in 3D can sometimes lead to a slight increase in the 2D projection MSPD error (decrease of AR\textsubscript{MSPD} and AP\textsubscript{MSPD}), for instance when the refined pose corrects depth or orientation errors in ways that shift the projected silhouette. This highlights a trade-off: optimizing for geometric consistency in 3D may not always translate to lower reprojection error in 2D, especially when the input candidates have a very low reprojection error.

Detailed results for HouseCat6D in \cref{table:housecat6d_metal_glass_supp} show that metallic objects are very challenging for FoundPose, especially their depth alignment (low AR\textsubscript{MSSD}, AR\textsubscript{VSD} and AP\textsubscript{MSSD}). When refined by our method, these metrics improve significantly. Similar trends can be observed for industrial datasets in \cref{tab:bop_industrial_full}.

\begin{table}[t]
\centering
\caption[]{\textbf{Evaluation on BOP-Industrial datasets.} This table is an extension of Tab.~\ref{table:unseen_results_bop_industrial} in the main paper. We obtain 2D detections from 3PT~\citep{3pt}, generate segmentations by SAM2~\citep{ravi2024sam} and generate candidates using FoundPose~\citep{ornek2024foundpose} official code base.}
\label{tab:bop_industrial_full}
\vspace{-3mm}
\small
\begin{tabular}{llccc}
\toprule
&\textbf{Method} & AP & AP\textsubscript{MSPD} & AP\textsubscript{MSSD} \\
\midrule
\multirow{3}{*}{\small IPD}
& FoundPose & 31.4 & 47.3 & 15.5 \\
& + CosyPose MV & 36.7 & 48.5 & 24.8 \\
& + Ours & \textbf{79.8} & \textbf{85.2} & \textbf{74.3} \\
\midrule
\multirow{2}{*}{\small XYZ} & FoundPose & 32.5 & 41.4 & 23.6 \\
\multirow{2}{*}{\small -IBD} & + CosyPose MV & 52.4 & 56.0 & 48.8 \\
& + Ours & \textbf{66.5} & \textbf{67.4} & \textbf{65.7} \\
\midrule
\multirow{2}{*}{\small ITODD} & FoundPose & 41.2 & 49.0 & 33.4 \\
\multirow{2}{*}{\small -MV} & + CosyPose MV & 54.5 & 56.7 & 52.2 \\
& + Ours & \textbf{76.8} & \textbf{75.9} & \textbf{77.7} \\
\bottomrule
\end{tabular}
\end{table}

\subsection{Non Maximum Suppression ablation}\label{sec:ablation_nms}
During NMS, each pose candidate is converted to its simple geometric representation which is used to measure overlap. Overlapping detections are then filtered via threshold $\theta$. The ablation of these paremeters is shown in \cref{tab:nms_ablation}.
While low values of $\theta$ over-suppress proximal objects and high  values of $\theta$ retain duplicates, $\theta=0.4$ provides an optimal balance. At this operating point, the geometric representation for IoU (e.g., axis-aligned boxes vs. bounding spheres) has negligible impact. We employ axis-aligned boxes; bounding spheres are used to approximate the translation-distance NMS of FreezeV2~\citep{caraffa2025freezev2}. The lack of public implementation details for the latter precludes a more direct comparison.

\begin{table}[t]
\centering
\caption[]{NMS with different object representation types (AA: axis-aligned boxes; S: bounding spheres) and IoU thresholds for the T-LESS dataset.}
\label{tab:nms_ablation}
\vspace{-3mm}
\small
\begin{tabular}{lcccc}
\toprule
Type & $\theta=0.2$ & $\theta=0.4$ & $\theta=0.6$  & $\theta=0.8$  \\
\midrule
 AA & 90.9 & 91.6 & 91.4 & 91.2 \\
 S & 90.6 & 91.6 & 91.6 & 90.6 \\
\bottomrule
\end{tabular}
\end{table}

\subsection{Cost function ablation}\label{sec:ablation_cost}
\begin{figure}[t]
    \centering
    \begin{subfigure}[b]{0.5\linewidth}
        \centering
        \includegraphics[width=\linewidth,  trim=3mm 0mm 2mm 0mm, clip]{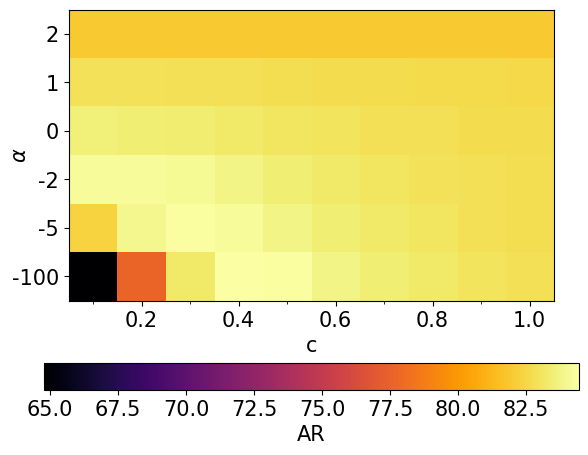}
        \caption{YCB-V}
        \label{fig:cost_abl_ycbv}
    \end{subfigure}%
    \begin{subfigure}[b]{0.5\linewidth}
        \centering
        \includegraphics[width=\linewidth,  trim=3mm 0mm 2mm 0mm, clip]{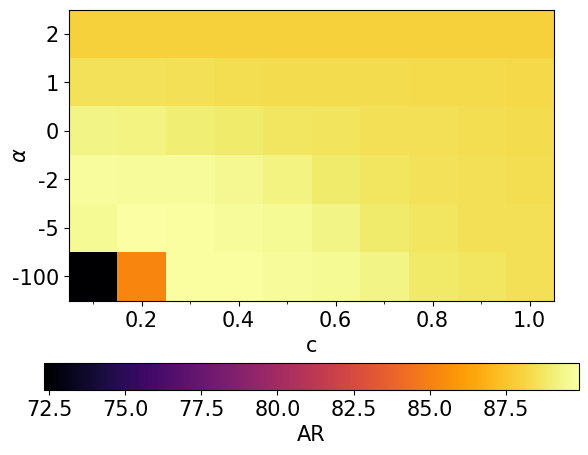}
        \caption{T-LESS}
        \label{fig:cost_abl_tless}
    \end{subfigure}
    \vspace{-6mm}
    \caption[]{\textbf{Ablation of the cost function parameters.} We ablate the value of parameters $\alpha$ (robustness) and $c$ (scale) of the Barron\citep{barron2019general} robust cost function. The maximum is reached at $\alpha = -5$ and $c \in [0.2,0.3]$. 
    }
    \label{fig:cost_ablation}
\end{figure}

We evaluate our approach under varying settings of the robust cost function. We ablate the Barron general loss~\citep{barron2019general} with respect to the robustness ($\alpha$) and scale ($c$) parameters in \cref{fig:cost_ablation}. This family spans several common losses, such as $\ell_2$ loss ($\alpha=2$). Decreasing $\alpha$, and in particular using negative $\alpha$, reduces the influence of outliers and increases robustness. We obtain the best performance for $\alpha=-5$. The difference in the final score between the $\ell_2$ loss and the optimal robust configuration is within $3\%$–$4\%$. In all experiments, we use $\alpha=-5$ and $c=0.5$.
Note that very negative $\alpha$ (e.g., $\alpha=-100$) combined with small $c$ (e.g., 0.1) flattens the loss, effectively treating all points as outliers and hindering optimization (see minima of \cref{fig:cost_ablation}). For further details on the Barron loss, please refer to the original work \citep{barron2019general}.

\subsection{Scoring function ablation}
We present an ablation study to assess the importance of the scoring function described in~\cref{sec:scoring} of the main paper. We replace the \textit{average} per-view feature-metric loss score by the \textit{minimal} or the \textit{maximal} score across views. The results are reported in~\cref{table:scoring_ablation}. The chosen average feature-metric loss scoring is better than the alternatives.

\begin{table}[t!]
\renewcommand{\arraystretch}{0.85} 
\centering
\caption[]{\textbf{Ablation of the scoring function.} We compare the proposed scoring based on the average feature-metric loss across views with a minimum or maximum score across views.
\label{table:scoring_ablation}}
\vspace*{-2mm}
\small
\begin{tabular}{lcc|cc}
\toprule
\multirow{2}{*}{\textbf{Scoring}} & \multicolumn{2}{c}{YCB-V} & \multicolumn{2}{c}{T-LESS} \\
& AR & AP & AR & AP \\
\midrule
Max  & 82.7 & 81.6 & 87.0 & 90.5 \\
Min  & 83.4 & 82.7 & 88.4 & 91.4 \\
Average & \textbf{83.8} & \textbf{83.3} & \textbf{89.6} & \textbf{92.4} \\
\bottomrule
\end{tabular}
\vspace{-4mm}
\end{table}

\begin{table*}[t]
\centering
\caption[]{\textbf{Impact of 2D object detection quality on 6D pose estimation.} We compare 3PT~\citep{3pt}, FOSOD (our implementation of \citep{lu2024adapting}), and CNOS (SAM)~\citep{nguyen2023cnos} detection methods.}
\label{tab:detection_quality}
\begin{tabular}{llccc|cc}
\toprule
\multirow{2}{*}{\textbf{Dataset}} & \multirow{2}{*}{\textbf{Detection method}} & \multicolumn{3}{c}{\textbf{2D Detection}} & \multicolumn{2}{c}{\textbf{6D Pose (AP)}}\\
\cmidrule(lr){3-5} \cmidrule(lr){6-7}
 &   &  AP & $\text{AR}_{10}$ & $\text{AR}_{100}$ &  FoundPose & AlignPose (ours) \\
\midrule
\multirow{3}{*}{\small IPD}
& 3PT       & 62.8 & 85.0 & 85.9 & 31.4 & 79.8 \\
& FOSOD     & 24.0 & 55.7 & 56.7 & 26.7 & 61.2 \\
& CNOS (SAM) & 20.7 & 26.5 & 26.5 & 17.1 & 37.0 \\
\midrule
\multirow{2}{*}{\small XYZ} & 3PT       & 54.2 & 35.3 & 72.4 & 32.5 & 66.5 \\
\multirow{2}{*}{\small -IBD} & FOSOD     & 25.8 & 22.0 & 46.4 & 24.4 & 50.8 \\
& CNOS (SAM) & 27.6 & 22.7 & 34.7 & 28.3 & 52.4 \\
\midrule
\multirow{2}{*}{\small ITODD} & 3PT       & 51.7 & 78.3 & 80.2 & 41.2 & 76.8 \\
\multirow{2}{*}{\small -MV} & FOSOD     & 40.2 & 61.1 & 62.8 & 31.7 & 69.3 \\
& CNOS (SAM) & 31.2 & 49.3 & 49.7 & 32.9 & 65.8 \\
\bottomrule
\end{tabular}
\end{table*}

\subsection{Impact of detection quality ablation}\label{sec:ablation_detection_quality}
To assess how 2D detection quality influences 6D pose estimation, we compare three detectors: 3PT~\citep{3pt}, FOSOD (our implementation of~\citep{lu2024adapting}), and CNOS (SAM)~\citep{nguyen2023cnos}. For 2D detection, we report standard BOP/COCO metrics: AP, $\text{AR}_{10}$, and $\text{AR}_{100}$~\citep{lin2014microsoft}.
\cref{tab:detection_quality} shows that high-quality 2D detections are essential for 6D pose estimation. Across all datasets, better detection performance consistently leads to higher pose estimation scores. The results confirm that our multi-view method AlignPose significantly outperforms single-view FoundPose across all detection qualities, as it can partially compensate for imperfect detections through multi-view aggregation. The results also illustrate that industrial datasets remain challenging for general-purpose detectors like CNOS and FOSOD.

\section{Qualitative results}\label{sec:appendix_qualitative_results}
\begin{figure*}[p]
    \centering
    \begin{tabular}{@{}c@{}c@{}c@{}c@{}}
        \small
        \textbf{Input image} & \textbf{Single-view} & \textbf{CosyPose MV} & \textbf{Ours} \\
        
        \includegraphics[width=0.24\linewidth, frame]{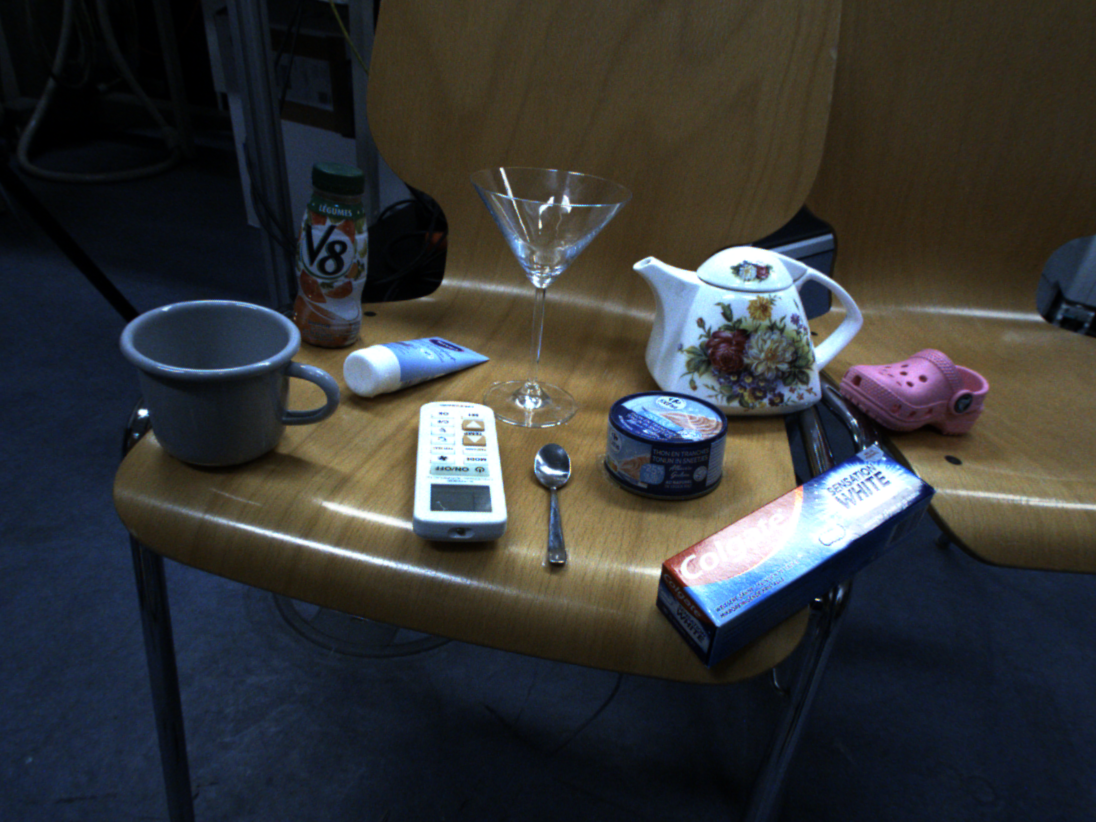} & 
        \colorbox{gray!80}{\includegraphics[width=0.24\linewidth, frame]{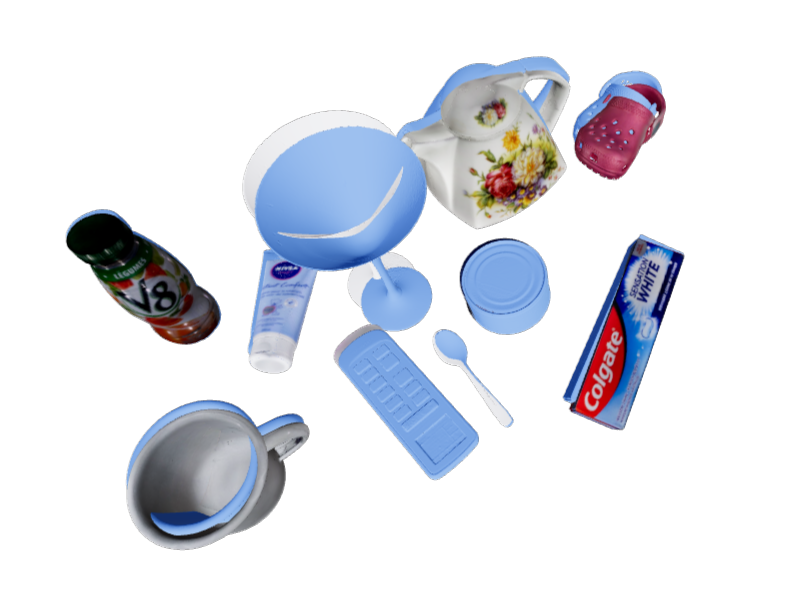}} & 
        \colorbox{gray!80}{\includegraphics[width=0.24\linewidth, frame]{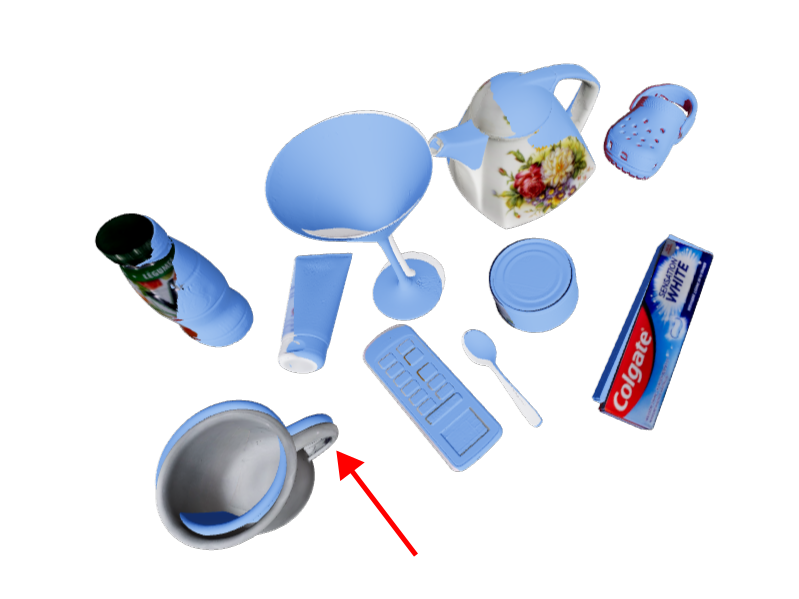}} & 
        \colorbox{gray!80}{\includegraphics[width=0.24\linewidth, frame]{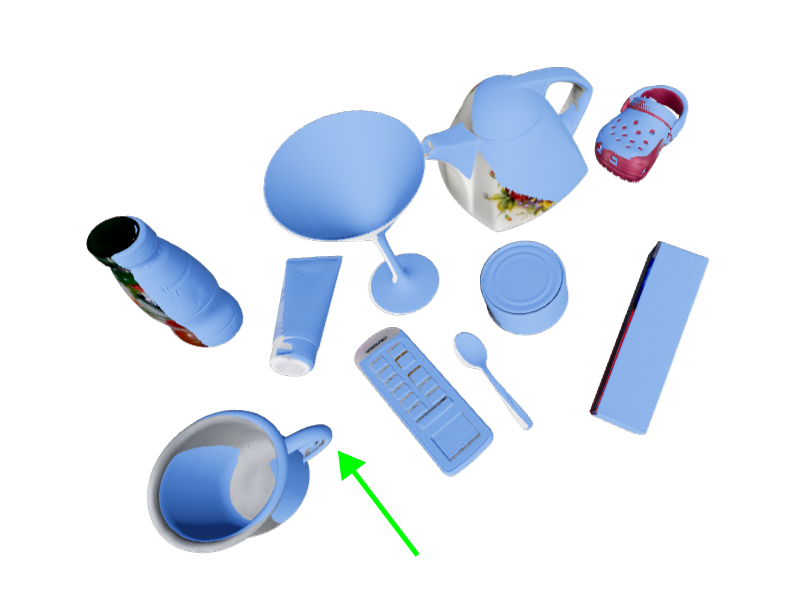}} \\ 
        
        \includegraphics[width=0.24\linewidth, frame]{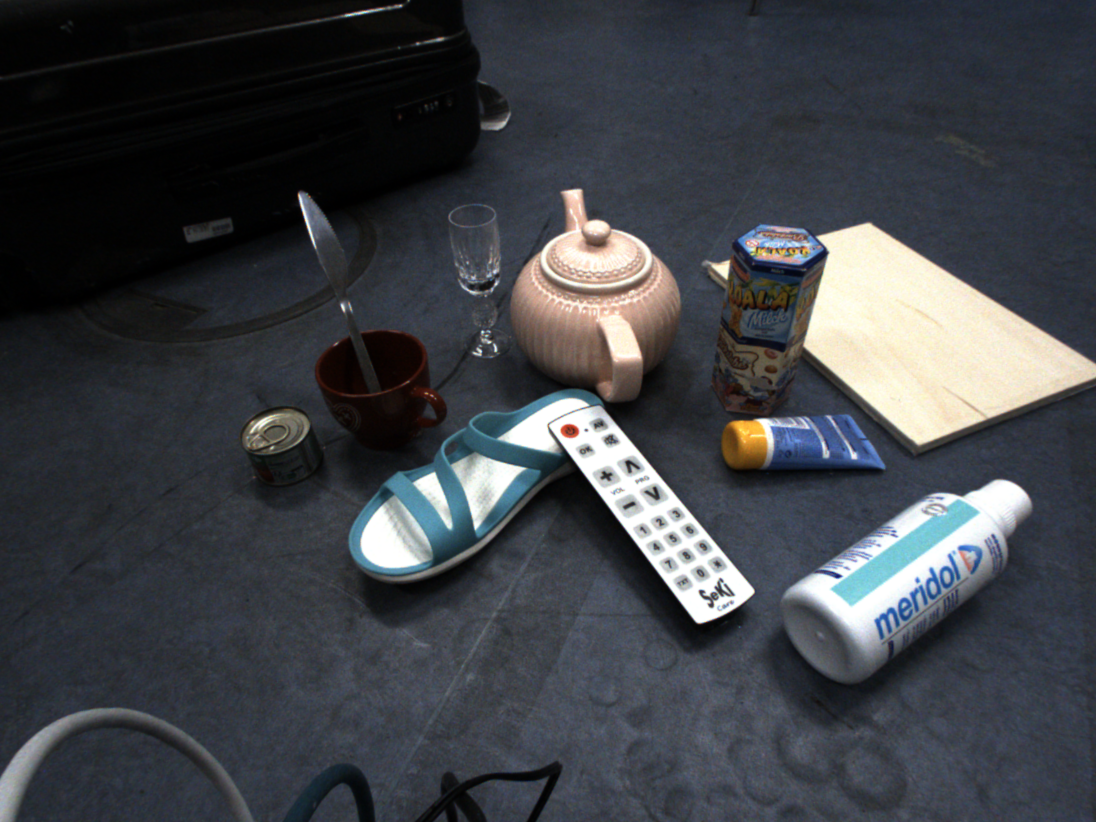} & 
        \colorbox{gray!80}{\includegraphics[width=0.24\linewidth, frame]{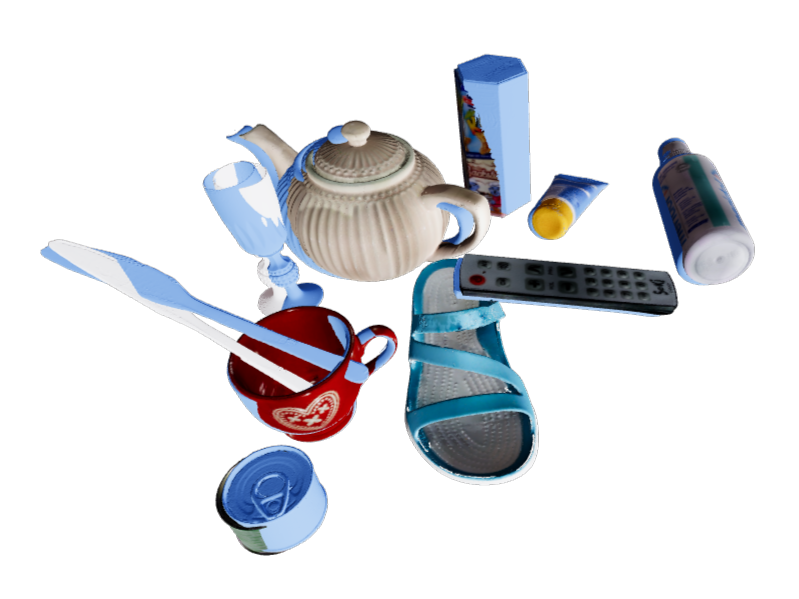}} & 
        \colorbox{gray!80}{\includegraphics[width=0.24\linewidth, frame]{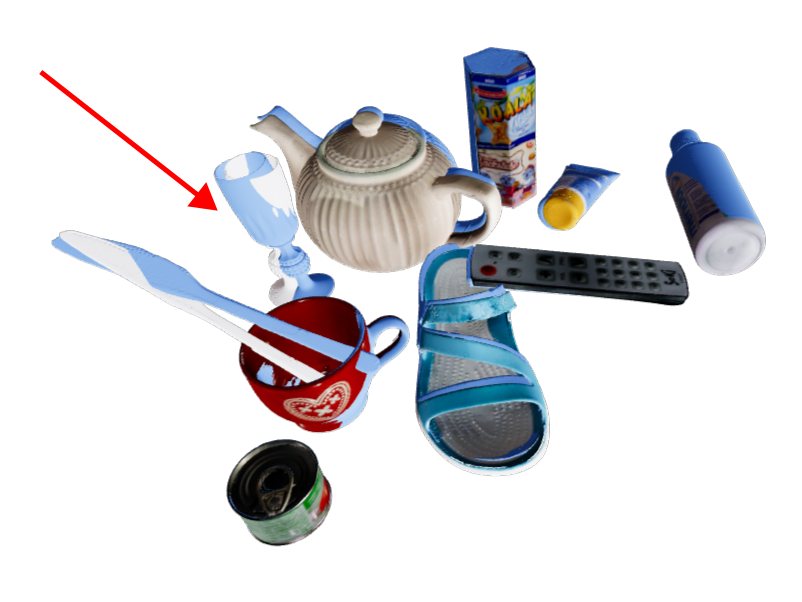}} & 
        \colorbox{gray!80}{\includegraphics[width=0.24\linewidth, frame]{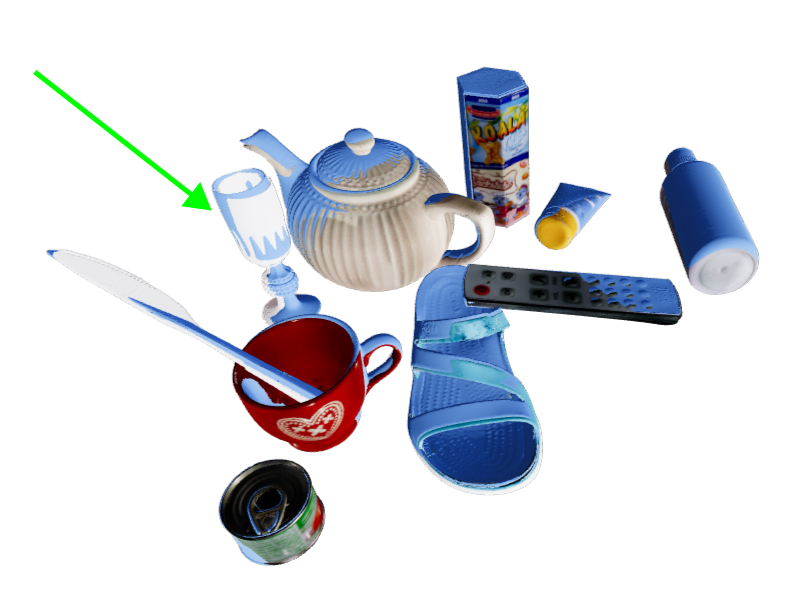}} \\
        
        \includegraphics[width=0.24\linewidth, frame]{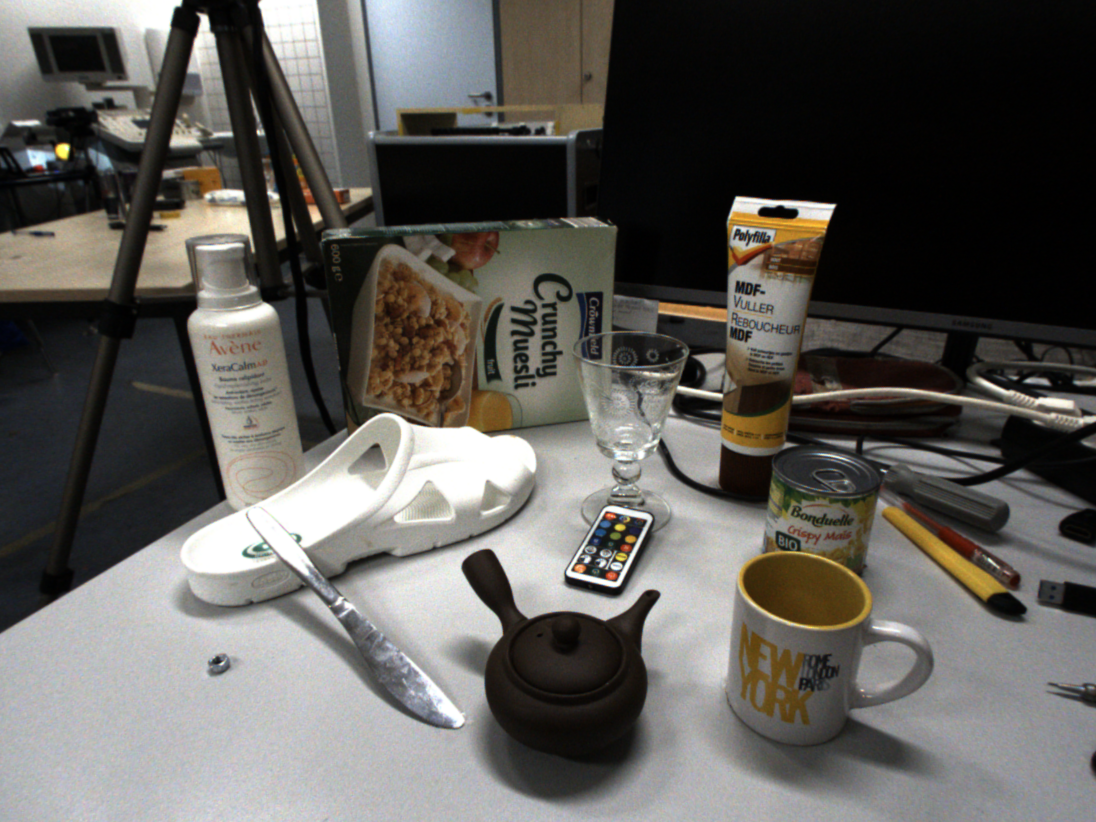} & 
        \colorbox{gray!80}{\includegraphics[width=0.24\linewidth, frame]{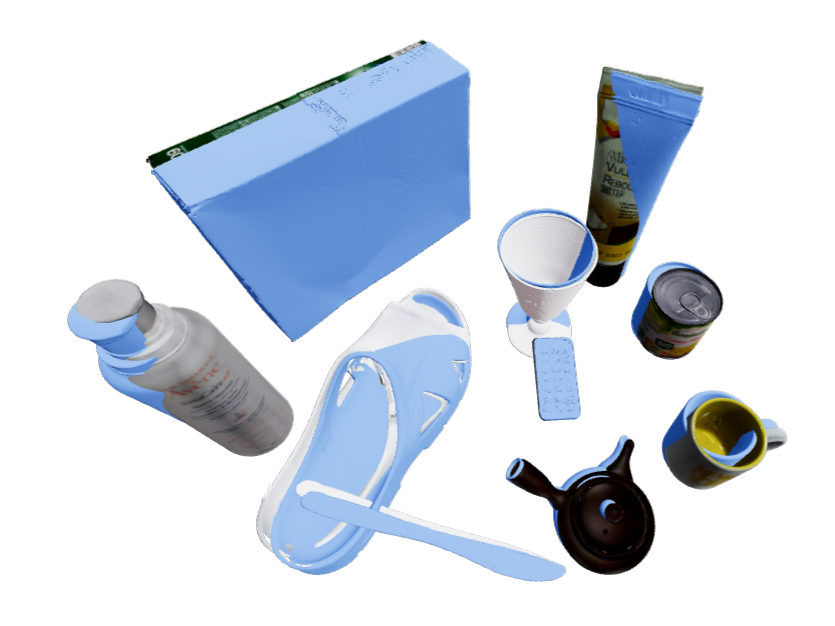}} & 
        \colorbox{gray!80}{\includegraphics[width=0.24\linewidth, frame]{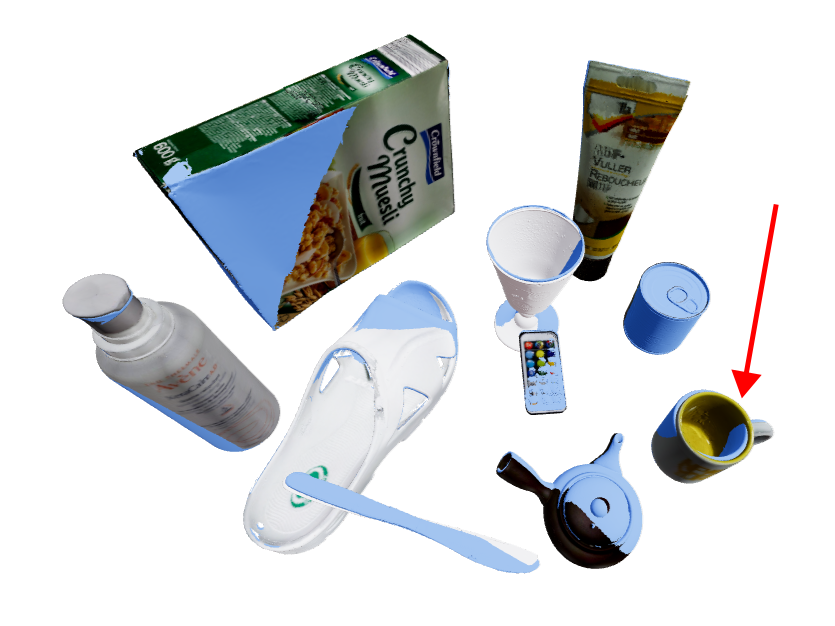}} & 
        \colorbox{gray!80}{\includegraphics[width=0.24\linewidth, frame]{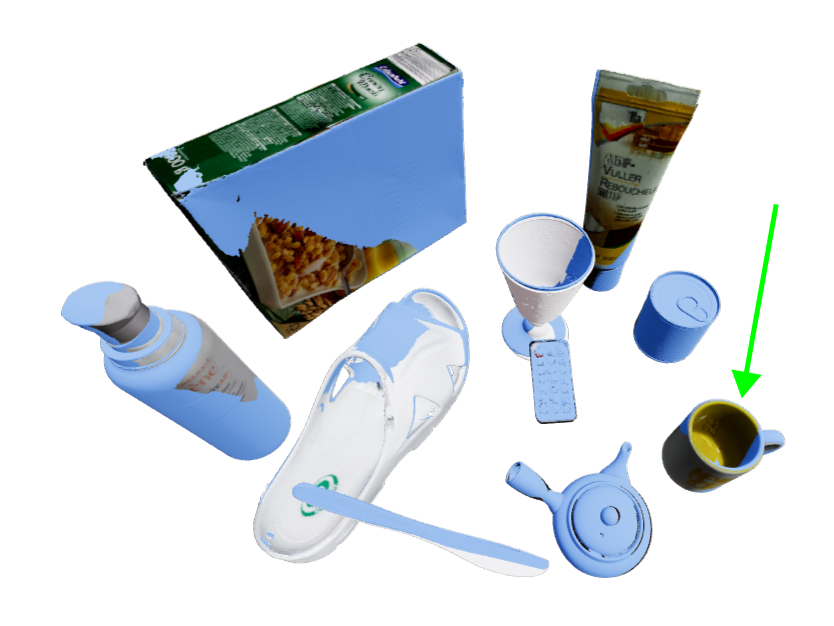}} \\
        
        \includegraphics[width=0.24\linewidth, frame]{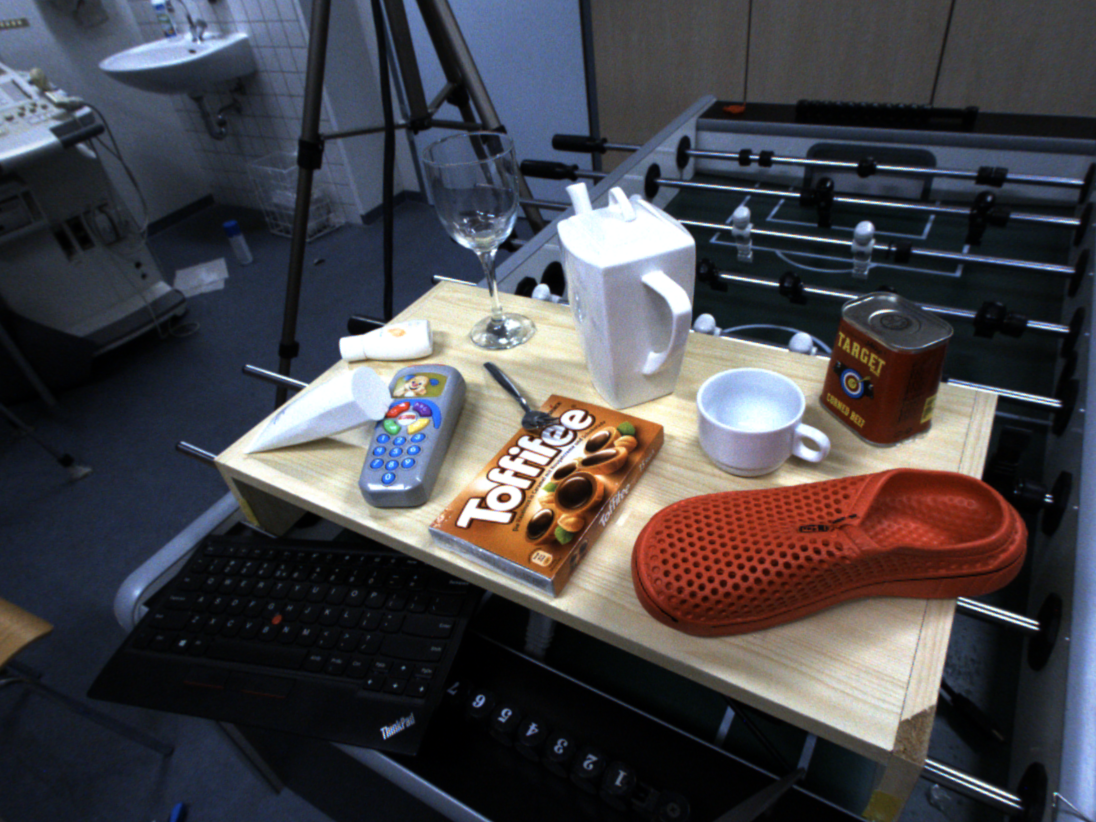} & 
        \colorbox{gray!80}{\includegraphics[width=0.24\linewidth, frame]{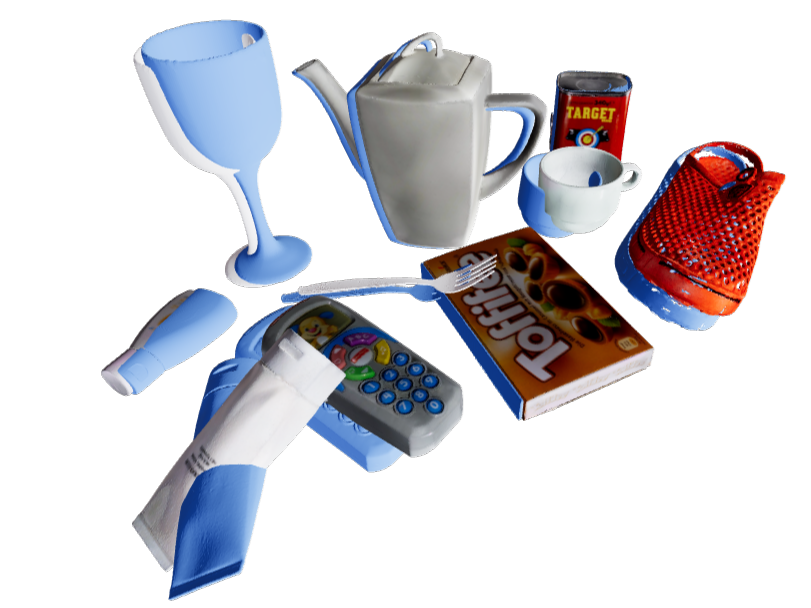}} & 
        \colorbox{gray!80}{\includegraphics[width=0.24\linewidth, frame]{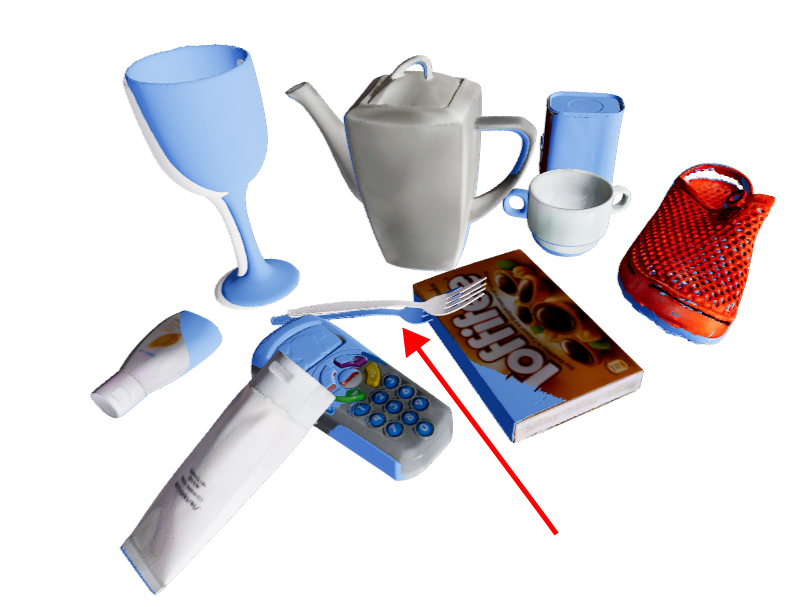}} & 
        \colorbox{gray!80}{\includegraphics[width=0.24\linewidth, frame]{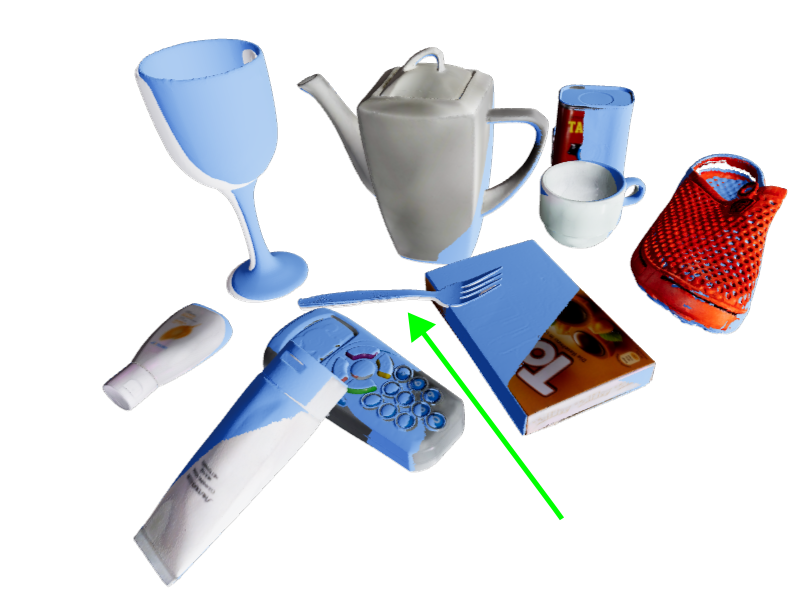}} \\ 
        
        \includegraphics[width=0.24\linewidth, frame]{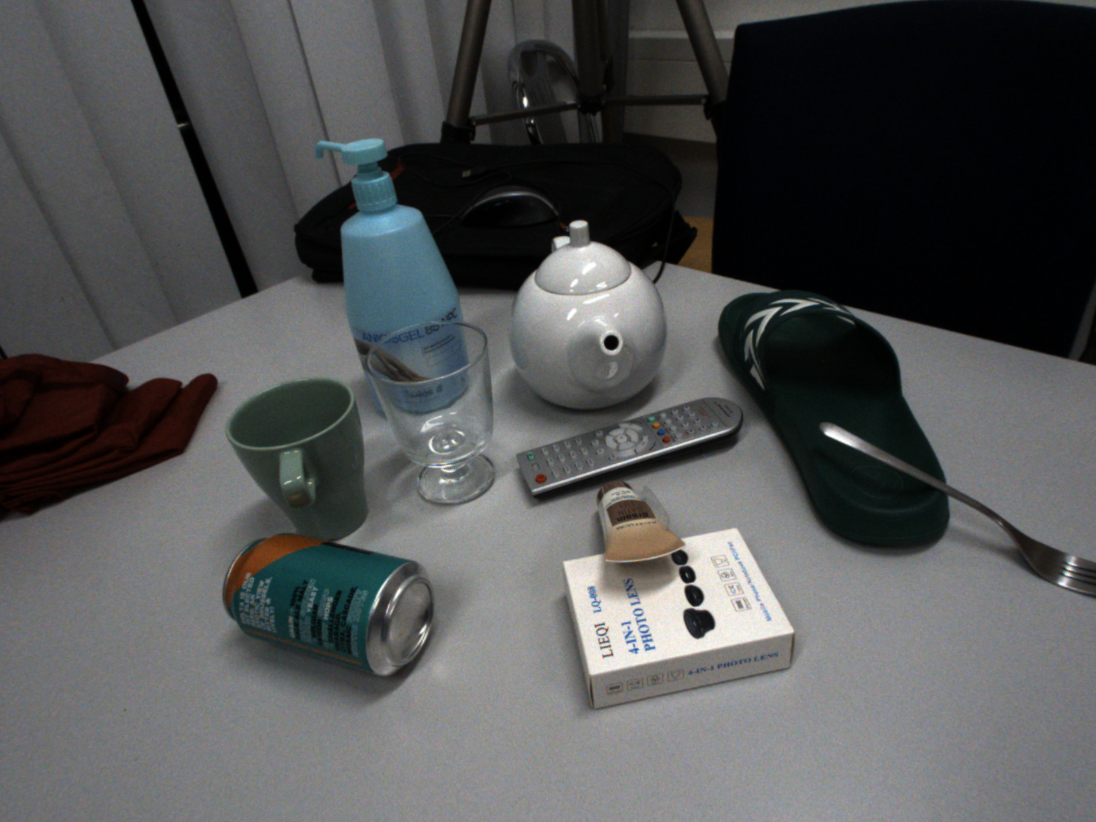} & 
        \colorbox{gray!80}{\includegraphics[width=0.24\linewidth, frame]{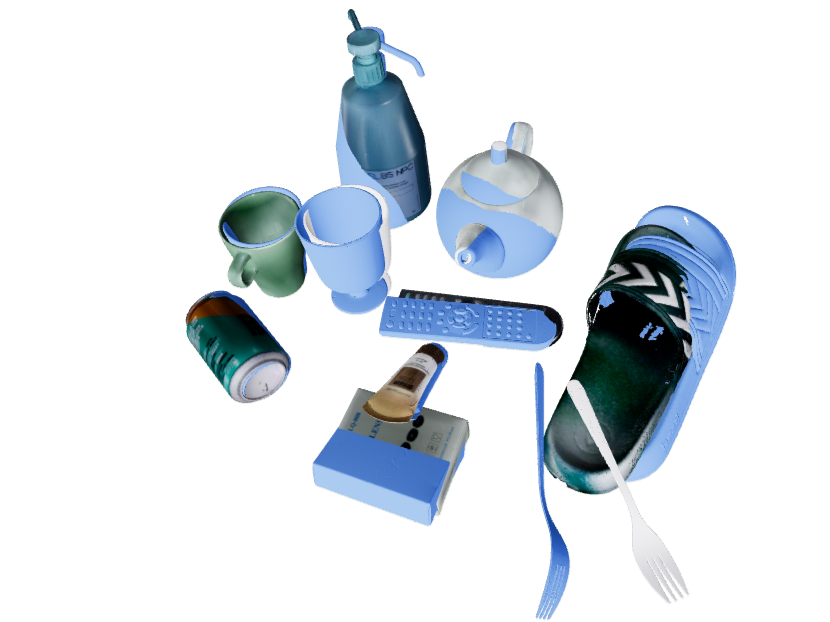}} & 
        \colorbox{gray!80}{\includegraphics[width=0.24\linewidth, frame]{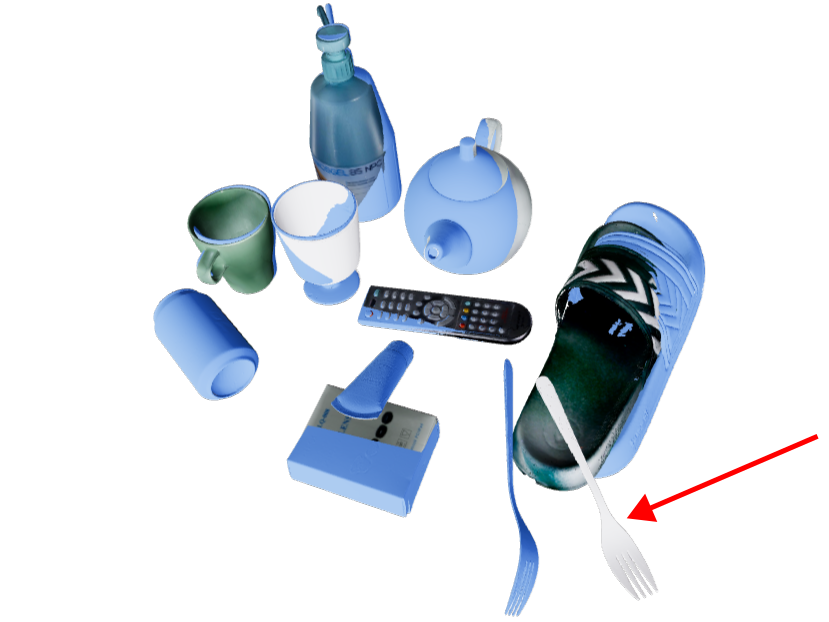}} & 
        \colorbox{gray!80}{\includegraphics[width=0.24\linewidth, frame]{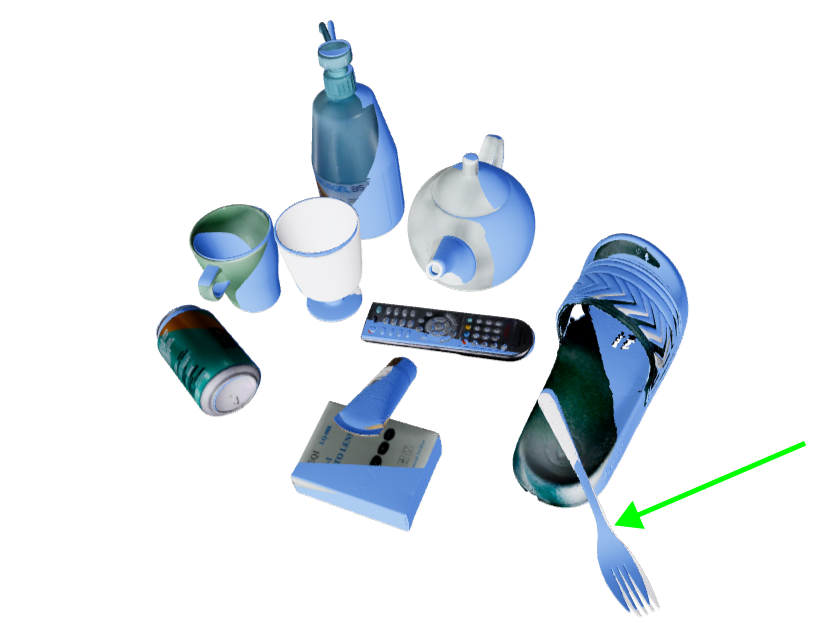}} \\
        \\

    \end{tabular}
    \vspace*{-2mm}
    \caption[]{\textbf{Qualitative results of multi-view refinement on the HouseCat6D dataset.} Column 1: Representative input image from the four-view set. Column 2: Single-view estimates (blue) overlaid on ground-truth (textured models). Column 3: Multi-view baseline (CosyPose). Column 4: Our proposed refinement. Both CosyPose and our method utilize single-view estimates from four input views. We highlight the improved pose accuracy of our method over CosyPose using \textcolor{MoreReadableGreen}{green} and \textcolor{red}{red} arrows respectively across several examples: "mug" (rows 1, 3), "glass" (row 2), and "cutlery" (rows 4, 5). 
    Note that while single-view estimates for cutlery are often significantly misaligned and left uncorrected by CosyPose, our method successfully recovers the poses by leveraging cross-view visual information. With the exception of occasional axial rotations for cutlery, our method achieves highly accurate alignment across all object categories.}
    \label{fig:supp_housecat_qualitative}
    \vspace{-3mm}
\end{figure*}

\begin{figure*}[p]
    \centering
    \small
    \begin{tabular}{@{}c@{}c@{}} 
        \textbf{} & \small 
        \makebox[0.23\linewidth]{\textbf{Main View}} 
        \makebox[0.23\linewidth]{\textbf{View 2}} 
        \makebox[0.23\linewidth]{\textbf{View 3}} 
        \makebox[0.23\linewidth]{\textbf{View 4}}\\
        
        \raisebox{1\height}{\textbf{1}} & 
        \includegraphics[width=0.92\textwidth, frame]{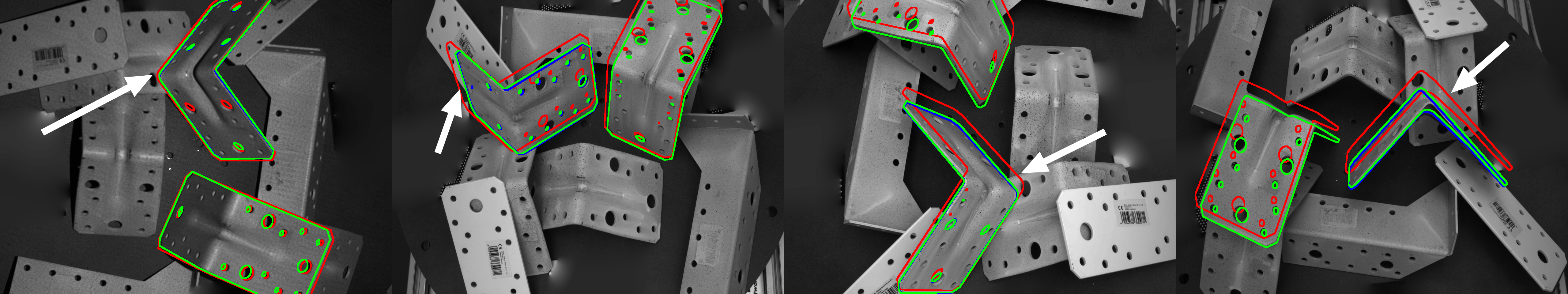} \\[10pt]
        
        \raisebox{0.5\height}{\textbf{2}} & 
        \includegraphics[width=0.92\textwidth, frame]{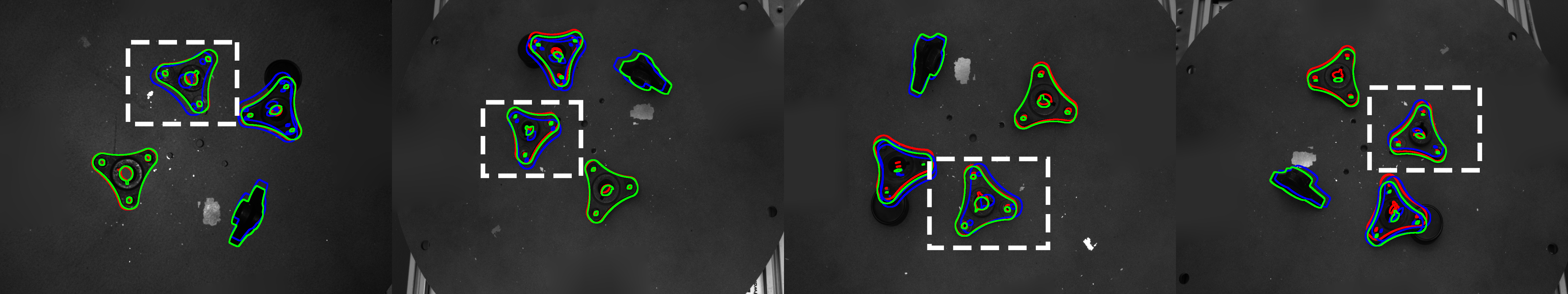} \\
        
        & 
        \raisebox{0pt}{\includegraphics[width=0.92\textwidth, frame]{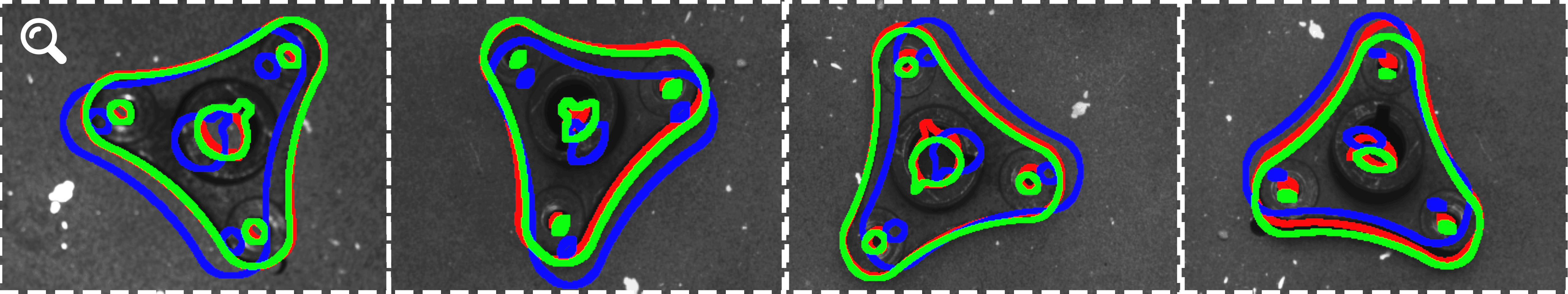}} \\[10pt]
        
        \raisebox{0.5\height}{\textbf{3}} & 
        \includegraphics[width=0.92\textwidth, frame]{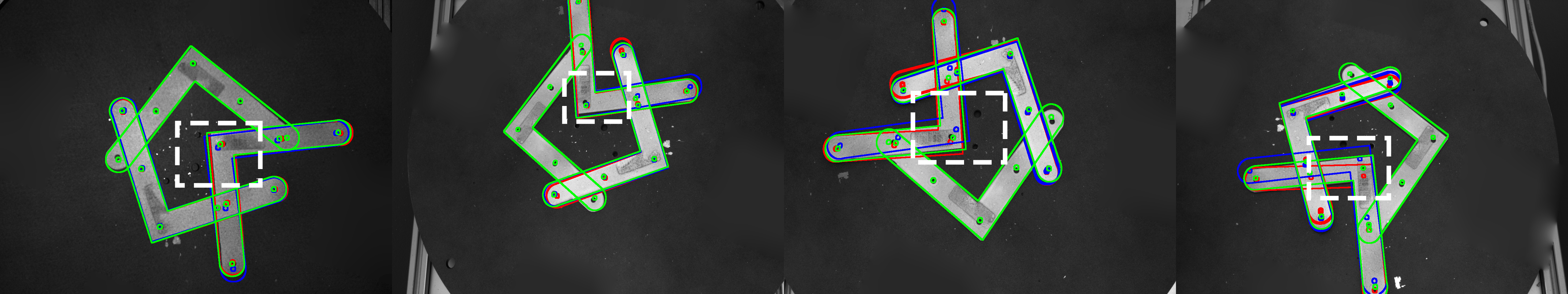} \\
        
        & 
        \includegraphics[width=0.92\textwidth, frame]{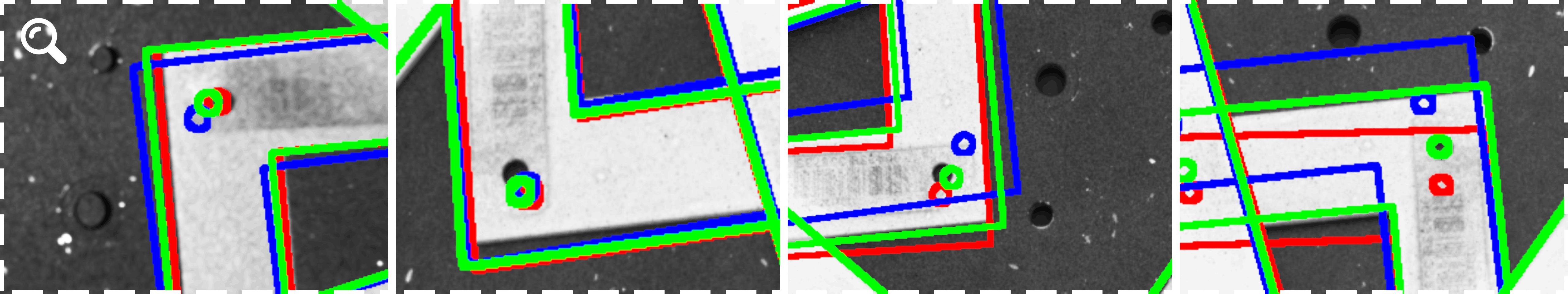} \\
    \end{tabular}

    \caption[]{\textbf{Qualitative results of multi-view refinement on the ITODD dataset.} Each row shows one scene observed from four viewpoints. Contours indicate predicted object poses. \textcolor{red}{Single-view} pose estimates (red) are computed using only the Main View; \textcolor{blue}{CosyPose multi-view} estimates are shown in blue; and \textcolor{MoreReadableGreen}{our multi-view} estimates are shown in green. Both our method and CosyPose use all four views. \textbf{Example 1:} The single-view method produces a reasonable pose in the Main View, but it becomes inaccurate when projected to the other views, as highlighted by the white arrows. In contrast, both multi-view methods yield consistent poses across all four views. \textbf{Examples 2 and 3:} Our method achieves higher accuracy than CosyPose: please pay attention to the slight misalignment of the blue contours, which is visible in the zoomed-in images.}
    \label{fig:supp_itoddmv_qualitative}
    \vspace{-3mm}
\end{figure*}

We provide additional qualitative results for HouseCat6D in \cref{fig:supp_housecat_qualitative} and for ITODD-MV in \cref{fig:supp_itoddmv_qualitative}.

The main trend in HouseCat6D is that single-view predictions are often slightly misaligned with the ground truth due to inaccurate estimation of the distance from the camera. Such errors are especially prominent for the transparent glass objects. The cutlery category presents an even greater challenge because the objects are thin, metallic, and texture-less. Single-view predictions are often inconsistent in both rotation and translation, and in some cases objects are not detected in all viewpoints. In these conditions, CosyPose may fail to perform its geometric multi-view refinement. Our method overcomes these issues by refining pose estimates with respect to visual evidence. We recover globally consistent poses even when individual per-view pose estimates are sparse, noisy, or missing entirely.  

The ITODD-MV dataset does not provide publicly available ground-truth object poses, therefore, we cannot show a full 3D scene visualization. Instead, in~\cref{fig:supp_itoddmv_qualitative}, we show the object contours corresponding to the predicted poses, as seen from four available views. The single-view predictions appear accurate in the "Main View" (the source view used for single-view pose estimation), but become highly misaligned when re-projected to the other views. In contrast, multi-view methods produce poses that are consistent across all views and our method demonstrates better pixel-wise alignment compared to both single-view and CosyPose baselines.

\section{Failure Cases and Limitations}\label{sec:appendix_limitations}
We identified the following three failure cases. First, we observe that when all available views come from a similar angle, our method cannot leverage multi-view information fully. This issue becomes more noticeable in challenging situations involving occlusions (\cref{fig:failure_case_occluded}). The second failure case appears when the input single-view pose candidate has a very large depth error, causing major cross-view projection misalignment. We provide an example in \cref{fig:failure_case_depth} where thin cutlery appears correctly placed from one viewpoint, while a second view reveals a large translation error. Our method fails to correct this estimate because the projected pose does not overlap with pixels corresponding to the cutlery.
Finally, nearly symmetric objects introduce ambiguity that can cause the optimization to become trapped in local minima. We illustrate this with an example of a cup from the HouseCat6D dataset, which features identical text on both sides (see \cref{fig:failure_case_symmetric}). As a result, the optimization may converge to an incorrect orientation of the cup.

In addition to these failure cases, our approach shares a common limitation with other methods in this category: it assumes accurately calibrated camera extrinsics. This limitation can be relaxed in future work by extending the objective to jointly optimize both object and camera poses.

\FloatBarrier

\noindent                       
\begin{minipage}{\columnwidth}  
    \centering                  
    \captionsetup{type=figure}  
    
    \begin{subfigure}{0.48\linewidth}
        \centering
        \includegraphics[width=\linewidth]{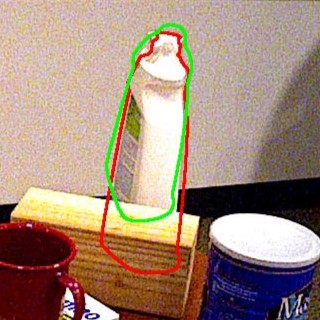}
    \end{subfigure}
    \begin{subfigure}{0.48\linewidth}
        \centering
        \includegraphics[width=\linewidth]{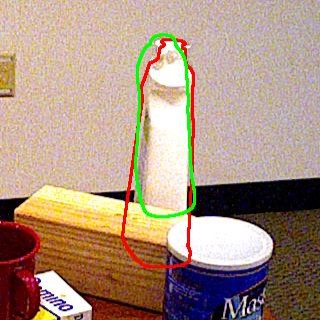}
    \end{subfigure}
     \vspace{-2mm}
     \captionof{figure}         
     {\textbf{Failure case I.: Similar viewpoints.} All cameras have similar viewpoints, causing the bleach bottle to be occluded by the wooden block in all views. The predicted pose (\textcolor{MoreReadableGreen}{green}) does not align with the ground truth pose (\textcolor{red}{red}) in the occluded regions.}
    \label{fig:failure_case_occluded}
    \vspace{7mm}

    \begin{subfigure}{0.48\linewidth}
        \centering
        \includegraphics[width=\linewidth]{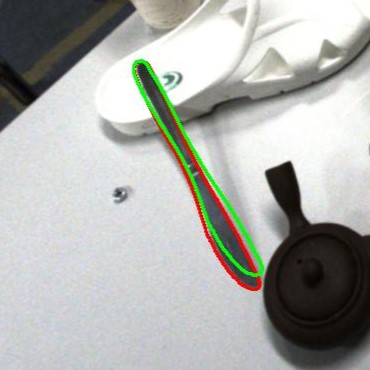}
    \end{subfigure}
    \begin{subfigure}{0.48\linewidth}
        \centering
        \includegraphics[width=\linewidth]{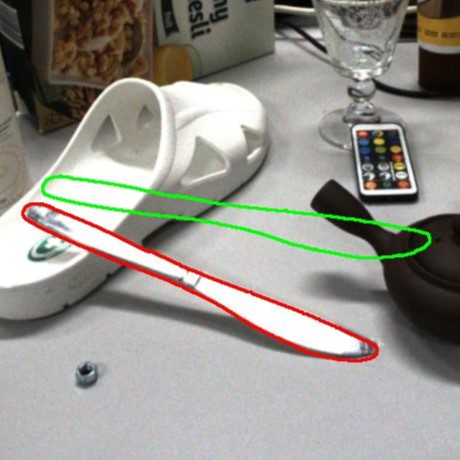}
    \end{subfigure}
     \vspace{-2mm}
     \captionof{figure}         
     {\textbf{Failure case II.: Large depth errors.} Pose prediction (\textcolor{MoreReadableGreen}{green}) appears correct from the first view (left), but the second view (right) reveals a translation error too large for optimization.}
    \label{fig:failure_case_depth}
    
    \vspace{5mm}
    \begin{subfigure}{0.48\linewidth}
        \centering
        \includegraphics[width=\linewidth, trim=0 50 0 0, clip]{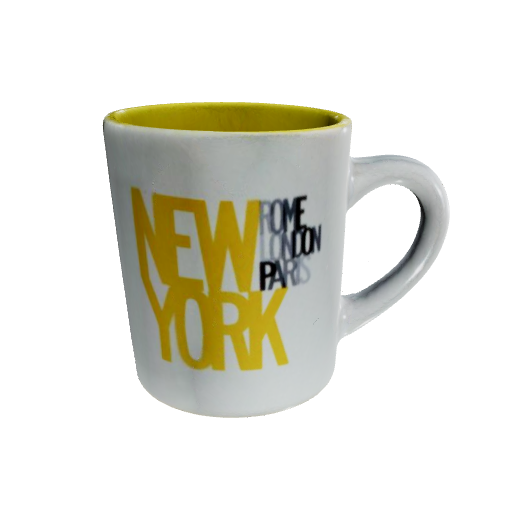}
        \caption{Front view}
    \end{subfigure}
    \begin{subfigure}{0.48\linewidth}
        \centering
        \includegraphics[width=\linewidth, trim=0 50 0 0, clip]{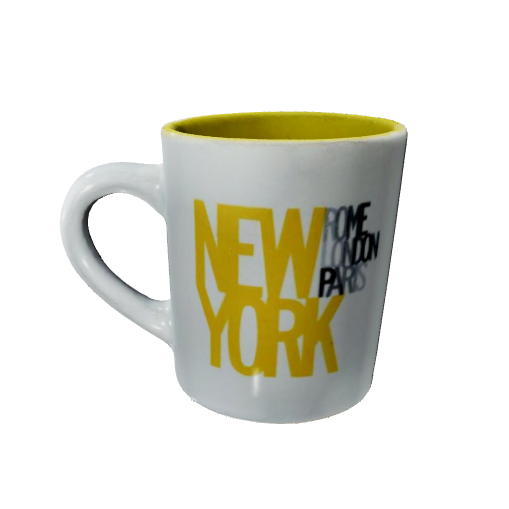}
        \caption{Back view}
    \end{subfigure}
    \begin{subfigure}{0.48\linewidth}
        \centering
        \includegraphics[width=\linewidth]{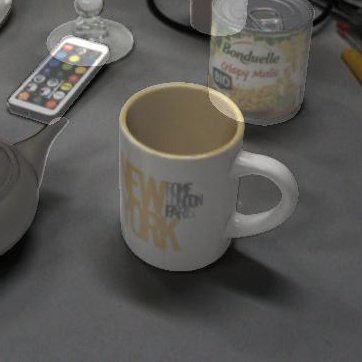}
        \caption{Ground truth pose}
    \end{subfigure}
    \begin{subfigure}{0.48\linewidth}
        \centering
        \includegraphics[width=\linewidth]{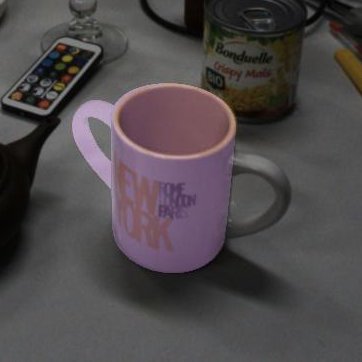}
        \caption{AlignPose prediction}
    \end{subfigure}
     \vspace{-2mm}
     \captionof{figure}         
     {\textbf{Failure case III.: Nearly symmetric objects.} Object "cup-new\_york" has the same texture on both sides, optimization gets stuck in local minimum.}
    \label{fig:failure_case_symmetric}
\end{minipage}                  

\end{document}